\documentclass{article}

\PassOptionsToPackage{font=small}{caption}
\PassOptionsToPackage{numbers}{natbib}
\usepackage{iclr2022_conference,times}

\usepackage{macros}
\usepackage{todonotes}

\usepackage[utf8]{inputenc} 
\usepackage[T1]{fontenc}    
\usepackage{hyperref}       
\usepackage{url}            
\usepackage{booktabs}       
\usepackage{amsfonts}       
\usepackage{nicefrac}       
\usepackage{microtype}      
\usepackage{xcolor}         
\usepackage{xparse}

\DeclareDocumentCommand\new{ m g }{%
    \IfNoValueF{#2}{\marginpar{~\\\textcolor{myorange}{#2}}}%
    {\color{myorange}#1}%
}

\theoremstyle{thmstyle}
\declaretheorem[style=thmstyle,name=Theorem,numberwithin=section]{thm}
\declaretheorem[style=thmstyle,name=Lemma,sharenumber=thm]{lemma}
\declaretheorem[style=thmstyle,name=Corollary,sharenumber=thm]{cor}
\declaretheorem[style=thmstyle,name=Proposition,sharenumber=thm]{prop}
\declaretheorem[style=thmstyle,name=Fact,sharenumber=thm]{fact}
\declaretheorem[style=thmstyle,name=Remark,sharenumber=thm]{rmk}

\newtheoremstyle{thmstyle}
  {\topsep} 
  {0.3pt} 
  {} 
  {} 
  {\bfseries} 
  {.} 
  {.5em} 
  {\thmname{#1}\thmnumber{ #2}\thmnote{ (#3)}} 

\theoremstyle{definition}
\declaretheorem[style=thmstyle,name=Definition,sharenumber=thm]{defn}
\declaretheorem[style=thmstyle,name=Notation,sharenumber=thm]{notation}

\setlist[itemize]{itemsep=-1pt, topsep=-1pt, leftmargin=18pt}

\title{
\resizebox{0.92\width}{!}{On the Learning and Learnability of \Qmets}
}

\author{%
  Tongzhou Wang \\
  MIT CSAIL\\
  \And
  Phillip Isola \\
  MIT CSAIL\\
}
\iclrfinalcopy
\begin{document}

\maketitle

\begin{abstract}

Our world is full of asymmetries. Gravity and wind can make reaching a place easier than coming back. Social artifacts such as genealogy charts and citation graphs are inherently directed. In reinforcement learning and control, optimal goal-reaching strategies are rarely reversible (symmetrical). Distance functions supported on these asymmetrical structures are called \emph{\qmets}. Despite their common appearance, little research has been done on the learning of quasimetrics.
Our theoretical analysis reveals that a common class of learning algorithms, including \uncon multilayer perceptrons (MLPs), provably fails to learn a \qmet consistent with training data. In contrast, our proposed \PQE (PQE) is the first \qmet learning formulation that both is learnable with gradient-based optimization and enjoys strong performance guarantees. Experiments on random graphs, social graphs, and offline Q-learning
demonstrate its effectiveness over
many
common baselines.

\vspace{-2pt}\noindent\begin{tabular*}{\textwidth}{@{}lr@{}}
Project Page: & \href{https://ssnl.github.io/quasimetric}{\small\texttt{ssnl.github.io/quasimetric}}.\\
Code: & \hspace{2.05cm}\href{https://github.com/SsnL/poisson_quasimetric_embedding}{\small\texttt{github.com/SsnL/poisson\_quasimetric\_embedding}}.
\end{tabular*}\vspace{-2.5pt}




\end{abstract}

\section{Introduction}

Learned \emph{symmetrical} metrics
have been proven useful for innumerable tasks including
dimensionality reduction \citep{tenenbaum2000global},
clustering \citep{xing2002distance}, classification \citep{weinberger2006distance,hoffer2015deep},
and
information retrieval \citep{wang2014learning}. 
However, the real world is largely \emph{asymmetrical}, and \emph{symmetrical} metrics can only capture a small fraction of it.

\begin{figure}
    \vspace{-26pt}
    \centering
    \includegraphics[scale=0.283, trim=385 105 80 80, clip]{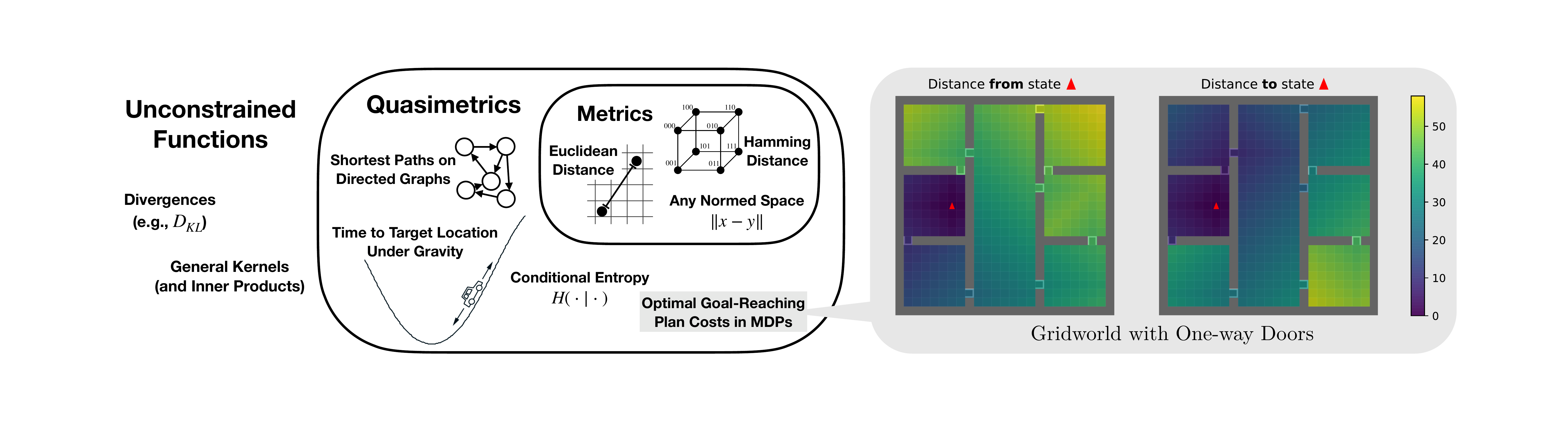}\vspace{-8pt}
    \caption{Examples of \qmet spaces. The car drawing is borrowed from \citet{sutton2018reinforcement}.}\label{fig:intro-qmet}%
    \vspace{-6pt}
\end{figure}

Generalizing metrics, \emph{\qmets} (\Cref{defn:qmet-space}) allow for \emph{asymmetrical} distances and can be found in a wide range of domains (see \Cref{fig:intro-qmet}).
Ubiquitous physical forces, such as gravity and wind, as well as human-defined rules, such as one-way roads, make the traveling time between places a \qmet. Furthermore, many of our social artifacts are directed graphs--- genealogy charts, follow-relation on Twitter \citep{snapnets}, citation graphs \citep{price2011networks}, hyperlinks over the Internet, \etc. Shortest paths on these graphs naturally induce \qmet spaces. In fact, we can generalize to Markov Decision Processes (MDPs) and observe that optimal goal-reaching plan costs (\ie, universal value/Q-functions \citep{schaul2015universal,sutton2011horde}) always form a \qmet \citep{bertsekas1991analysis,tian2020model}. Moving onto more abstract structures, \qmets can also be found as expected hitting times in Markov chains, and as conditional Shannon entropy $H(\cdot\given\cdot)$ in information theory. (See the appendix for proofs and discussions of these \qmets.)

In this work, we study the task of \emph{\qmet learning}. Given a sampled training set of pairs and their \qmet distances, we ask: \ul{how well can we learn a quasimetric that fits the training data}?
We define \emph{\qmet learning} in analogy to metric learning: whereas metric learning is the problem of learning a metric function, \qmet learning is the problem of learning a \qmet function. This may involve searching over a hypothesis space constrained to only include \qmet functions (which is what our method does) or it could involve searching for approximately \qmet functions (we compare to and analyze such approaches).
Successful formulations have many potential applications, such as structural priors in reinforcement learning \citep{schaul2015universal,tian2020model}, graph learning \citep{rizi2018shortest} and causal relation learning \citep{balashankar2021learning}.

Towards this goal, our contributions are  \begin{itemize}[topsep=-3.3pt, itemsep=-1.2pt]
    \item We study the \qmet learning task with two goals: (1) fitting training data well and (2) respecting \qmet constraints (\Cref{sec:task-intuition});
    \item We prove that a large family of algorithms, including \uncon networks trained in the Neural Tangent Kernel (NTK) regime  \citep{jacot2018neural}, fail at this task, while a learned embedding into a latent \qmet space can potentially succeed (\Cref{sec:theory});
    \item We propose \PQEs (PQEs), the first \qmet embedding formulation learnable with gradient-based optimization that also enjoys strong theoretical guarantees on approximating arbitrary \qmets (\Cref{sec:emb});
    \item Our experiments complement the theory and demonstrate the benefits of PQEs on random graphs, social graphs and offline Q-learning (\Cref{sec:expr}).
\end{itemize}

\section{Preliminaries on \Qmets and Poisson Processes}

\label{sec:preliminaries}


{\textbf{\Qmet space}} is a generalization of metric space where all requirements of metrics are satisfied, except that the distances can be asymmetrical.
\begin{repthm}[\Qmet Space]{defn}{defn:qmet-space}{defnQmetSpace}
    A \emph{\qmet space} is a pair $(\mathcal{X}, d)$, where $\mathcal{X}$ is a set of points and $d \colon \mathcal{X} \times \mathcal{X} \rightarrow [0, \infty]$ is the \qmet, satisfying the following conditions: \begin{align}
        \forall x, y \in \mathcal{X}, \qquad & x = y \iff d(x, y) = 0,\hspace{-2.2cm}  \tag{\idtindis} \\[-0.2ex]
        \forall x, y, z \in \mathcal{X}, \qquad & d(x, y) + d(y, z) \geq d(x, z).\hspace{-2.2cm} \tag{Triangle Inequality}
    \end{align}
\end{repthm}

Being asymmetric, \qmets are often thought of as (shortest-path) distances of some (possibly infinite) weighted directed graph. A natural way to quantify the complexity of a \qmet is to consider that of its underlying graph. \emph{\Qmet \tw} is an instantiation of this idea.




\begin{repthm}[Treewidth of \Qmet Spaces \citep{memoli2018quasimetric}]{defn}{defn:qmet-tw}{defnQmetTw}
\ifdefined\appdefn%
    Consider representations of a \qmet space $M$ as shortest-path distances on a positively-weighted directed graph.  \emph{\Tw} of $M$ is the minimum over all such graphs' \tws. (Recall that the \tw of a graph (after replacing directed edges with undirected ones) is a measure of its complexity.)
\else%
    Consider a \qmet space $M$ as shortest-path distances on a positively-weighted directed graph.  \emph{\Tw} of $M$ is the minimum over all such graphs' \tws.
\fi
\end{repthm}



{\textbf{Poisson processes}} are commonly used to model events (or points) randomly occurring across a set $A$ \citep{kingman2005p} , \eg, raindrops hitting a windshield,
photons captured by a camera. The number of such events within a subset of $A$ is modeled as a Poisson distribution, whose mean is given by a measure $\mu$ of $A$ that determines how ``frequently the events happen at each location''.

\begin{repthm}[Poisson Process]{defn}{defn:poisson-process}{defnPoissonProcess}
    For nonatomic measure $\mu$ on set $A$, a \emph{Poisson process} on $A$ with \emph{mean measure} $\mu$ is a random countable subset $P \subset A$ (\ie, the  random events / points) such that \begin{itemize}[topsep=-3pt, itemsep=-2.5pt]
        \item for any disjoint measurable subsets $A_1, \dots, A_n$ of $A$, the random variables $N(A_1), \dots, N(A_n)$ are independent, where $\smash{N(B) \trieq \#\{P \cap B\}}$ is 
        the number of points of $P$ in $B$, and
        \item $N(B)$ has the Poisson distribution with mean $\mu(B)$, denoted as $\pois(\mu(B))$.
    \end{itemize}
\end{repthm}

\begin{fact}[Differentiability of $\prob{N(A_1) \leq N(A_2)}$]\label{fact:diff-poi-race}

For two measurable subsets $A_1, A_2$, \begin{equation}
    \prob{N(A_1) \leq N(A_2)}
    = \mathbb{P}\big[\underbrace{\pois(\mu(A_1 \setminus A_2)) \leq \pois(\mu(A_2 \setminus A_1))}_{\text{two \emph{independent} Poissons}}\big].
\end{equation}
Furthermore, for independent $X \sim \pois(\mu_1)$, $Y \sim \pois(\mu_2)$, the probability $\prob{X \leq Y}$ is \emph{differentiable \wrt $\mu_1$ and $\mu_2$}. In the special case where $\mu_1$ or $\mu_2$ is zero, we can simply compute \begin{align}
    \prob{X \leq Y}
    & = \begin{cases}
        \prob{0 \leq Y}=1 & \text{if $\mu_1 = 0$} \\[-0.2ex]
        \prob{X \leq 0}=\prob{X = 0} = e^{-\mu_1} & \text{if $\mu_2 = 0$}
    \end{cases} \tag{$\pois(0)$ is always $0$} \\[-0.4ex]
    & = \exp\left(-(\mu_1 - \mu_2)^+\right), \label{eq:pois-race-prob-special-case}
\end{align}
where $x^+ \trieq \max(0, x)$.
For general $\mu_1, \mu_2$, this probability and its gradients can be obtained via a connection to noncentral $\chi^2$ distribution \citep{johnson1959poissonchisq}. We derive the formulas in the appendix.

Therefore, if $A_1$ and $A_2$ are parametrized by some $\theta$ such that $\mu(A_1 \setminus A_2)$ and $\mu(A_2 \setminus A_1)$ are differentiable \wrt $\theta$, so is $\prob{N(A_1) \leq N(A_2)}$.
\end{fact}

\section{\Qmet Learning}
\label{sec:task-intuition}

Consider a \qmet space $(\mathcal{X}, d)$. The \emph{\qmet learning} task aims to infer a \qmet from observing
a training set $\{(x_i, y_i, d(x_i, y_i))\}_i \subset \mathcal{X} \times \mathcal{X} \times [0, \infty]$.
Naturally, our goals for a learned predictor $\smash{\hat{d} \colon \mathcal{X} \times \mathcal{X} \rightarrow \R}$ are: \ul{respecting the quasimetric constraints} and \ul{fitting training distances}.

Crucially, we are not simply aiming for the usual sense of \emph{generalization}, \ie, low population error. Knowing that true distances have a \qmet structure, we can better evaluate predictors and desire ones that fit the training data and are (approximately) \qmets.
These objectives also indirectly capture generalization because a predictor failing either requirement must have large error on some pairs, whose true distances follow \qmet constraints. We formalize this relation in \Cref{thm:dis-vio-gen}.

\subsection{Learning Algorithms and Hypothesis Spaces}
Ideally, \qmet learning should scale well with data, potentially generalize to unseen samples, and support integration with other deep learning systems (\eg, via differentiation). 

\textbf{Relaxed hypothesis spaces.} One can simply learn a generic function approximator that maps the (concatenated) input pair to a scalar as the prediction of the pair's distance, or its transformed version (\eg, log distance). This approach has been adopted in learning
graph distances \citep{rizi2018shortest} and
plan costs in MDPs \citep{tian2020model}. When the function approximator is a deep neural network, we refer to such methods as \emph{\uncon networks}. While they are known to fit training data well \citep{jacot2018neural}, in this paper we also investigate whether they learn to be (approximately) \qmets.

\textbf{Restricted hypothesis spaces.} %
Alternatively, we can encode each input to a latent space $\smash{\mathcal{Z}}$, where a latent \qmet $\smash{d_z}$ gives the distance prediction. This guarantees learning a quasimetric over data space $\smash{\mathcal{X}}$. Often $d_z$ is restricted to a subset unable to approximate all \qmets, \ie, an \textbf{overly restricted hypothesis space}, such as metric embeddings and the recently proposed DeepNorm and WideNorm  \citep{pitis2020inductive}. While our proposed \PQE (PQE) (specified in \Cref{sec:emb}) is also a latent \qmet, it  can approximate arbitrary \qmets (and is differentiable). PQE thus searches in \textbf{a space that approximates all \qmets and only \qmets}.





\subsection{A Toy Example}\label{sec:failure-uncon-metric-emb}

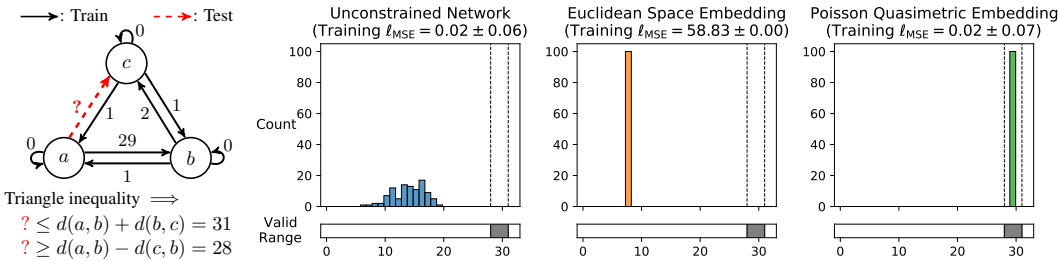
\begin{figure}
\vspace{-22pt}%
\vspace{-11pt}
    \begin{subfigure}[t]{0.21\linewidth}
        \centering%
        \scalebox{0.75}{%
           \begin{tikzpicture}[bayes_net, node distance = 1.5cm, every node/.style={inner sep=0}]
                \node[main_node, minimum size=0.72cm] (a) {$a$};
                \node[main_node, minimum size=0.72cm] (b) [right = 1.5cm of a] {$b$};
                \node[main_node, minimum size=0.72cm] (c) [above right = 1.15cm and 0.58cm of a] {$c$};

                \node[] (legend) [above = 0.36cm of c, inner sep=0pt] {\small

                \raisebox{2pt}{\protect\tikz{\protect\draw[line width=1.5pt, ->, thick, >=stealth'] (0, 0) -- (0.7, 0);}}\hspace{-0.5pt}: Train\hspace{5ex}
                \raisebox{2pt}{\protect\tikz{\protect\draw[line width=1.5pt, ->, thick, >=stealth', dashed, red] (0, 0) -- (0.7, 0);}}\hspace{-0.5pt}: Test
                };

                \node[] (constraints) [below = 1.9cm of c, inner sep=0pt, align=left] {\shortstack{\small
                Triangle inequality $\implies$\hphantom{wqewcm}\\[0.1ex]
                $
                    \begin{aligned}
                        {\color{red} ? } & \leq d(a, b) + d(b, c) = 31 \\[-1.1ex]
                        {\color{red} ? } & \geq d(a, b) - d(c, b) = 28
                    \end{aligned}
                $}};
                \node[] (filler) [below = 0.05cm of constraints, inner sep=0pt] {};

                \tikzset{myptr/.style={decoration={markings,mark=at position 1 with %
                    {\arrow[scale=0.825,>=stealth']{>}}},postaction={decorate}}}
                \path[]
                (a) edge[out=165,in=195,-, myptr, line width=1pt,looseness=4] [right] node [above=4pt] {\small$0$} (a)
                (c) edge[out=75,in=105,-, myptr, line width=1pt,looseness=4] [right] node [right=4pt] {\small$0$} (c)
                (b) edge[out=345,in=15,-, myptr, line width=1pt,looseness=4] [right] node [above right=3pt and -0.5pt] {\small$0$} (b)
                (a.15) edge[-, myptr, line width=1pt] [right] node [midway, above=3pt] {\small$29$} (b.165)
                (b.135) edge[-, myptr, line width=1pt] [right] node [midway, below left=-3pt and 3pt] {\small$2$} (c.302)
                (b.195) edge[-, myptr, line width=1pt] [right] node [midway, below=3pt] {\small$1$} (a.-15)
                (c.332) edge[-, myptr, line width=1pt] [right] node [midway, right=3pt] {\small$1$} (b.105)
                (c.247) edge[-, myptr, line width=1pt] [right] node [midway, right=3pt] {\small$1$} (a.45)
                (a.75) edge[-, myptr, dashed, line width=1.3pt, red] [right] node [left=3pt] {\color{red} \small$\boldsymbol{?}$} (c.217)
                ;
            \end{tikzpicture}%
        }
    \end{subfigure}\hfill%
    \begin{subfigure}[t]{0.273\linewidth}
        \centering
        \includegraphics[scale=0.464]{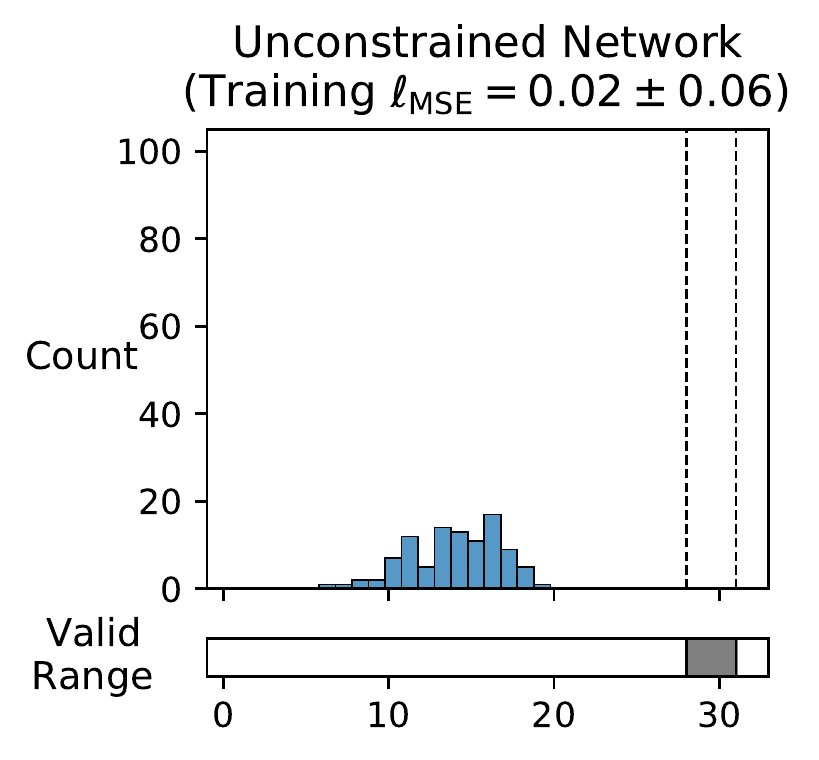}
    \end{subfigure}\hspace{0.165ex}%
    \begin{subfigure}[t]{0.243\linewidth}
        \centering
        \includegraphics[scale=0.464]{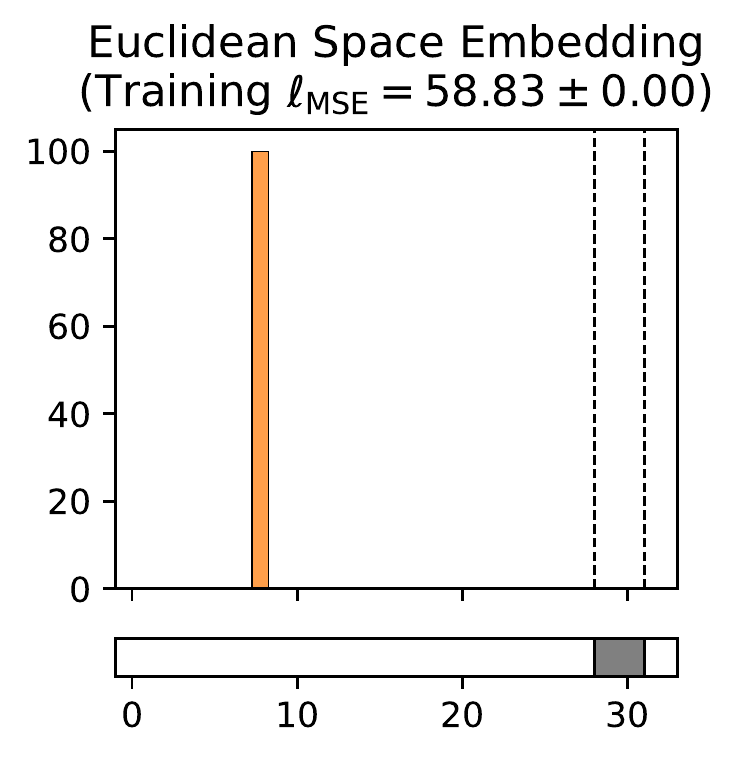}
    \end{subfigure}\hspace{0.165ex}%
    \begin{subfigure}[t]{0.243\linewidth}
        \centering
        \includegraphics[scale=0.464]{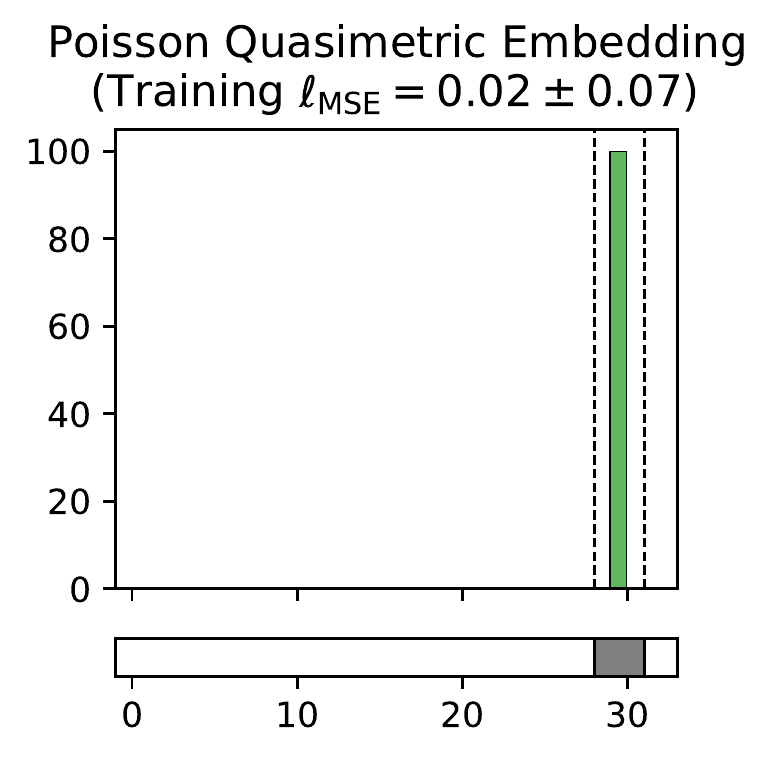}
    \end{subfigure}%
    \vspace{-7pt}
    \vspace{-2pt}
    \caption{\Qmet learning on a $3$-element space. \textbf{Leftmost:} Training set contains all pairs except for $(a, c)$. Arrow labels show \qmet distances (rather than edge weights). A \qmet $\smash{\hat{d}}$ should predict $\smash{\hat{d}(a, c) \in [28, 30]}$. \textbf{Right three: }Different formulations are trained to fit training pairs distances, and then predict on the test pair. Plots show distribution of the prediction over $100$ runs.
    }    \label{fig:3-node} %
    \vspace{-5pt}
\end{figure}

To build up intuition on how various algorithms perform according to our two goals, we consider a toy \qmet space with only $3$ elements in \Cref{fig:3-node}. The space has a total of $9$ pairs, $8$ of which form the training set. Due to \qmet requirements (\esp triangle inequality), knowing distances of these $8$ pairs restricts valid values for the heldout pair to a particular range (which is $[28, 31]$ in this case). If a model approximates $8$ training pairs well \emph{and} respects \qmet constraints well, its prediction on that heldout pair should fall into this range.

We train three models \wrt mean squared error (MSE) over the training set using gradient descent: \begin{itemize}[itemsep=-0.6pt, topsep=-4.5pt]
    \item \Uncon deep network that predicts distance,
    \item Metric embedding into a latent Euclidean space with a deep encoder,
    \item \Qmet embedding into a latent PQE space with a deep encoder (our method from \Cref{sec:emb}).
\end{itemize}

The three approaches exhibit interesting qualitative differences.
Euclidean embedding, unable to model asymmetries in training data, fails to attain a low training error. While both other methods approximate training distances well, \uncon networks greatly violate \qmet constraints; only PQEs respect the constraints and consistently predicts within the valid range.

Here, the structural prior of embedding into a \qmet latent space appears important to successful learning. Without any such prior, \uncon networks fail badly. In the next section, we present a rigorous theoretical study of the \qmet learning task, which confirms this intuition.


\section{Theoretical Analysis of Various Learning Algorithms}

\label{sec:theory}

In this section, we define concrete metrics for the two \qmet learning objectives stated above, and present positive and negative theoretical findings for various learning algorithms.

\myparagraph{Overview.} Our analysis focuses on data-agnostic bounds, which are of great interests in machine learning (\eg, VC-dimension \citep{vapnik2015uniform}).  We prove a strong negative result for a general family of learning algorithms (including \uncon MLPs trained
in NTK regime, $k$-nearest neighbor, and min-norm linear regression): they may arbitrarily badly fail to fit training data or respect \qmet constraints  (\Cref{thm:orth-equi-bad}). Our informative construction reveals the core reason of their failure. \Qmet embeddings, however, enjoy nice properties as long as they can approximate arbitrary \qmets, which motivates searching for ``universal \qmets''. The next section presents PQEs as such universal approximators and states their theoretical guarantees.

\myparagraph{Assumptions.} We consider \qmet spaces $(\mathcal{X}, d)$ with $\mathcal{X}\subset \R^d$, finite size $n = \size{X} < \infty$, and finite distances (\ie,  $d$ has range $[0, \infty)$). It allows discussing deep networks which can’t handle infinities well. This mild assumption can be satisfied by simply capping max distances in \qmets. For training, $\smash{m < n^2}$ pairs are uniformly sampled as training pairs $\smash{S \subset \mathcal{X} \times  \mathcal{X}}$ without replacement.

In the appendix, we provide all full proofs, further discussions of our assumptions and presented results, as well as additional results concerning specific learning algorithms and settings.

\subsection{\Dis and \Vio Metrics for \Qmet Learning}


We use \emph{\dis} as a measure of how well the distance is preserved, as is standard in embedding analyses (\eg, \citet{bourgain1985lipschitz}). In this work, we especially consider \emph{\dis over a subset of pairs}, to quantify how well a predictor $\smash{\hat{d}}$ approximates distances over the training subset $S$.

\begin{defn}[\Dis]\label{defn:dis}
    \Dis of $\hat{d}$ over a subset of pairs
    $S \subset \mathcal{X} \times \mathcal{X}$ is $\dis_S(\hat{d}) \trieq \big( \max_{(x, y) \in S, x \neq y} \frac{ \hat{d}(x, y)}{d(x, y)} \big) \big( \max_{(x, y) \in S, x \neq y} \frac{d(x, y)}{\hat{d}(x, y)} \big)$,
    and its overall distortion is $\dis(\hat{d}) \trieq \dis_{\mathcal{X} \times \mathcal{X}}(\hat{d})$.\vspace{-2pt}
\end{defn}


For measuring consistency \wrt \qmet constraints, we define the \emph{(\qmet) violation} metric. \Vio focuses on \emph{triangle inequality}, which can often be more complex (\eg, in \Cref{fig:3-node}), compared to the relatively simple \emph{non-negativity} and \emph{\idtindis}.



\begin{repthm}[\Qmet \Vio]{defn}{defn:qmet-vio}{defnQmetVio}
\emph{\Qmet violation} (\emph{\vio} for short)  of $\hat{d}$ is $\vio(\hat{d}) \trieq \max_{
    \substack{
    A_1, A_2, A_3 \in \mathcal{X}
    }} \frac{\hat{d}(A_1, A_3)
    }{\hat{d}(A_1, A_2) + \hat{d}(A_2, A_3)}$, where we define $\frac{0}{0}=1$ for notation simplicity.%
 \vspace{-2pt}
\end{repthm}

Both \dis and \vio are nicely agnostic to scaling. Furthermore, assuming \emph{non-negativity} and \emph{\idtindis}, $\smash{\vio(\hat{d}) \geq 1}$ always, with equality iff $\hat{d}$ is a \qmet.

\Dis and \vio also capture generalization. Because the true distance $d$ has optimal training \dis (on $S$) and \vio, a predictor $\smash{\hat{d}}$ that does badly on either must also be far from truth.
\begin{repthm}[\Dis and \Vio Lower-Bound Generalization Error]{thm}{thm:dis-vio-gen}{thmDisVioGen}
For non-negative $\hat{d}$, $\dis(\hat{d}) \geq \max(\dis_S(\hat{d}), \sqrt{\smash{\vio(\hat{d})}\vphantom{d^{2^2}}})$, where $\dis(\hat{d})$ captures generalization over the entire $\mathcal{X}$ space.%
 \vspace{-2pt}%
\end{repthm}\vspace{-1pt}

\subsection{Learning Algorithms Equivariant to Orthogonal Transforms}\label{sec:theory-qmet-vio-orth-equi}

For \qmet space $(\mathcal{X}, d)$, $\mathcal{X} \subset \R^d$,
we consider applying general learning algorithms by concatenating pairs to form inputs $\in \R^{2d}$ (\eg, \uncon networks). While straightforward, this approach means that algorithms are generally \emph{unable to relate the same element appearing as 1st or 2nd input}. As we will show, this is sufficient for a wide family of learning algorithms to fail badly-- ones \textbf{equivariant to orthogonal transforms} (\orthequi algorithms; \Cref{defn:equi-alg}).

For an \orthequi algorithm, training on orthogonally transformed data does not affect its prediction, as long as test data is identically transformed. In fact, many standard learning algorithms are \orthequi, including unconstrained MLP trained in NTK regime (\Cref{lemma:orth-equi-ex}).
%
\begin{repthm}[Equivariant Learning Algorithms]{defn}{defn:equi-alg}{defnEquiAlg}
    Given training set $\mathcal{D} = \{(z_i, y_i)\}_i \subset \mathcal{Z} \times \mathcal{Y}$, where $z_i$ are inputs and $y_i $ are targets, a learning algorithm $\mathsf{Alg}$ produces a function $\mathsf{Alg}(\mathcal{D}) \colon \mathcal{Z} \rightarrow Y$ such that $\mathsf{Alg}(\mathcal{D})(z')$ is the function's prediction on sample $z'$.
    Consider $\mathcal{T}$ a set of transformations $\mathcal{Z} \rightarrow \mathcal{Z}$. $\mathsf{Alg}$ is equivariant to $\mathcal{T}$ iff for all transform $T \in \mathcal{T}$, training set $\mathcal{D}$, $\mathsf{Alg}(\mathcal{D}) = \mathsf{Alg}(T \mathcal{D})\circ T$,
    where $T \mathcal{D} = \{(T z, y) \colon (z, y) \in \mathcal{D}\}$ is the training set with transformed inputs.
\end{repthm}
%
%
\begin{repthm}[Examples of \orthequi Algorithms]{lemma}{lemma:orth-equi-ex}{lemmaOrthEquiEx}
$k$-nearest-neighbor with Euclidean distance, dot-product kernel ridge regression (including min-norm linear regression and MLP trained with squared loss in NTK regime) are \orthequi.%
 \vspace{-1pt}
\end{repthm}

\begin{figure}
    \centering%
    \vspace{-6.5pt}%
\ifdefined\appdefn%
\vspace{-14pt}%
\else
\vspace{-29pt}%
\fi
\scalebox{0.89}{
\begin{tikzpicture}[bayes_net, node distance = 0.5cm, every node/.style={inner sep=0}]
    \node[main_node, minimum size=0.82cm] (1y) {$y$};
    \node[main_node, minimum size=0.82cm] (1x) [above left = -0.02cm and 2.4cm of 1y] {$x$};
    \node[main_node, minimum size=0.82cm] (1z) [below left = -0.02cm and 2.4cm of 1y] {$z$};
    \node[main_node, minimum size=0.82cm] (1yp) [above right = -0.02cm and 2.4cm of 1x] {$y'$};
    \node[main_node, minimum size=0.82cm] (1wp) [above right = -0.02cm and 2.4cm  of 1y] {$w'$};
    \node[main_node, minimum size=0.82cm] (1w) [below right = -0.02cm and 2.4cm of 1y] {$w$};
    \node[] (1ineq) [below = 0.8cm of 1y, inner sep=4pt, rounded corners=0.3cm, fill=black, opacity=0.08, text opacity=1, align=center,
    minimum width=7.07cm
    ]
    {\small
    $\displaystyle
        \vio(\hat{d})
        \geq \frac{\hat{d}(x, z)}{\hat{d}(x, y) + \hat{d}(y, z)}
        \geq \frac{c\vphantom{\hat{d}(y, z)}}{\dis_S(\hat{d})(\dis_S(\hat{d})+\hat{d}(y, z))}
        $%
    };

    \node[] (1tr) [below left = 0cm and -7.5cm of 1ineq, inner sep=4pt, rounded corners=0.3cm] {$\displaystyle\begin{aligned}
        \text{Training (\raisebox{2pt}{\protect\tikz{\protect\draw[line width=1.25pt, ->, thick, >=stealth'] (0, 0) -- (0.65, 0);}}\hspace{-1pt}) : }\hphantom{d}
        d(x, z) &= c,\hphantom{1l}
        d(w, z) = 1, \\[-0.6ex]
        d(x, {\color{blue} y}) &= 1,\hphantom{cl}
        d(y, {\color{blue} w'}) = 1. \\[-0.3ex]
        \text{Test (\raisebox{2pt}{\protect\tikz{\protect\draw[line width=1.25pt, ->, thick, >=stealth', dashed] (0, 0) -- (0.65, 0);}}\hspace{-1pt}) : }\hphantom{d} \hat{d}(y, z) & = {\color{red} ?}
    \end{aligned}$};

    \tikzset{myptr/.style={decoration={markings,mark=at position 1 with %
        {\arrow[scale=0.825,>=stealth']{>}}},postaction={decorate}}}
    \path[]
    (1x) edge[-, myptr, line width=1.4pt] [right] node [left=3pt] {$c$} (1z)
    (1w) edge[-, myptr, line width=1.4pt] [right] node [below=3pt] {$1$} (1z)
    (1x) edge[-, myptr, line width=1.4pt] [right] node [midway, above=3pt] {$1$} (1y)
    (1y) edge[-, myptr, line width=1.4pt] [right] node [pos=0.45, above=3pt] {$1$} (1wp)
    (1y) edge[-, myptr, dashed, line width=1.4pt] [right] node [above=3pt] {\color{red} $?$} (1z)
    ;

    \node[main_node, minimum size=0.82cm] (2y) [right = 7.25cm of 1y] {$y$};
    \node[main_node, minimum size=0.82cm] (2x) [above left = -0.02cm and 2.4cm of 2y] {$x$};
    \node[main_node, minimum size=0.82cm] (2z) [below left = -0.02cm and 2.4cm of 2y] {$z$};
    \node[main_node, minimum size=0.82cm] (2yp) [above right = -0.02cm and 2.4cm of 2x] {$y'$};
    \node[main_node, minimum size=0.82cm] (2wp) [above right = -0.02cm and 2.4cm  of 2y] {$w'$};
    \node[main_node, minimum size=0.82cm] (2w) [below right = -0.02cm and 2.4cm of 2y] {$w$};

    \node[] (2ineq) [below = 0.8cm of 2y, inner sep=4pt, rounded corners=0.3cm, fill=black, opacity=0.08, text opacity=1, align=center,
    minimum width=7.07cm
    ]
    {
    \small
    $\displaystyle
     \vio(\hat{d})
        \geq \frac{\hat{d}(y, z)}{\hat{d}(y, w) + \hat{d}(w, z)}
        \geq \frac{\hat{d}(y, z)}{2\cdot\dis_S(\hat{d})}
    $
    };

    \node[] (2tr) [below left = 0cm and -7cm of 2ineq, inner sep=4pt, rounded corners=0.3cm] {$\displaystyle\begin{aligned}
        \text{Training (\raisebox{2pt}{\protect\tikz{\protect\draw[line width=1.25pt, ->, thick, >=stealth'] (0, 0) -- (0.65, 0);}}\hspace{-1pt}) : }\hphantom{d}
        d(x, z) &= c,\hphantom{1l}
        d(w, z) = 1, \\[-0.6ex]
        d(x, {\color{blue} y'}) &= 1,\hphantom{cl}
        d(y, {\color{blue} w}) = 1. \\[-0.3ex]
        \text{Test (\raisebox{2pt}{\protect\tikz{\protect\draw[line width=1.25pt, ->, thick, >=stealth', dashed] (0, 0) -- (0.65, 0);}}\hspace{-1pt}) : }\hphantom{d} \hat{d}(y, z) & = {\color{red} ?}
    \end{aligned}$};

    \path[]
    (2x) edge[-, myptr, line width=1.4pt] [right] node [left=3pt] {$c$} (2z)
    (2w) edge[-, myptr, line width=1.4pt] [right] node [below=3pt] {$1$} (2z)
    (2x) edge[-, myptr, line width=1.4pt] [right] node [midway, above=3pt] {$1$} (2yp)
    (2y) edge[-, myptr, line width=1.4pt] [right] node [pos=0.45, above=3pt] {$1$} (2w)
    (2y) edge[-, myptr, dashed, line width=1.4pt] [right] node [above=3pt] {\color{red} $?$} (2z)
    ;


    \begin{scope}[on background layer]
        \fill[fill=black, opacity=0.08] \convexpath{1z,1x,1y}{0.53cm};
        \fill[fill=black, opacity=0.08] \convexpath{2y,2w,2z}{0.53cm};
    \end{scope}
\end{tikzpicture}
}
\vspace{-22pt}
\caption{
Two training sets
pose incompatible constraints (\raisebox{-2pt}{\protect\tikz{\fill[fill=black, opacity=0.15]  circle(1ex);}})
for the test pair distance $d(y, z)$.
%
With one-hot features, an orthogonal transform can exchange $(*, {\color{blue}y})\leftrightarrow (*, {\color{blue}y'})$ and $(*, {\color{blue}w}) \leftrightarrow (*, {\color{blue}w'})$, leaving the test pair $(y, z)$ unchanged, but transforming the training set from one scenario to the other. Given either set, an \orthequi algorithm must attain same training \dis and predict identically on $(y, z)$. For appropriate $c$, this implies large \dis (not fitting training set) or \vio (not approximately a quasimetric) in one of these cases.
}
\ifdefined\appdefn%
\else
\vspace{-1pt}%
\fi%
    \label{fig:orth-equi-fail-simple}\vspace{-5.5pt}
\end{figure}
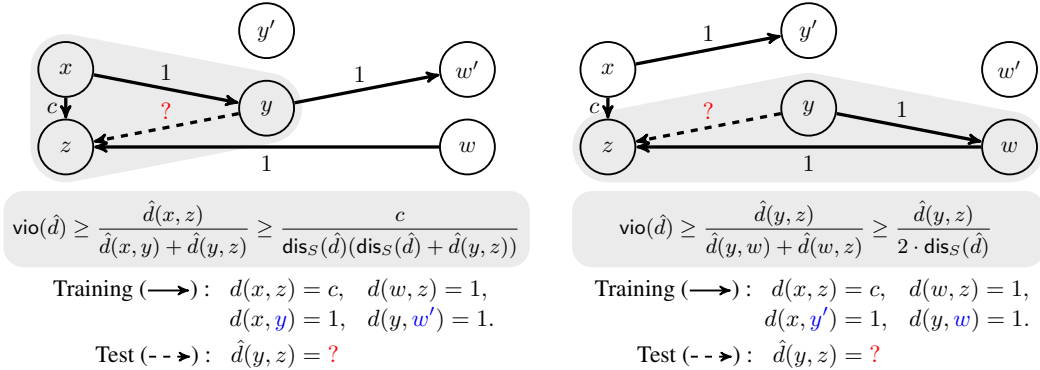

\myparagraph{Failure case.} 
These algorithms treat the concatenated inputs as generic vectors. If a transform fundamentally changes the \qmet structure but is not fully reflected in the learned function (\eg, due to equivariance), learning must fail.
The two training sets  in \Cref{fig:orth-equi-fail-simple} are sampled from two different \qmets over the same $6$ elements An orthogonal transform links both training sets \emph{without affecting the test pair}, which is constrained differently in two \qmets. An \orthequi algorithm, necessarily predicting the test pair identically seeing either training set, must thus fail on one.
In the appendix, we empirically verify that \uncon MLPs indeed \emph{do fail} on this construction.

Extending to larger \qmet spaces, we consider graphs containing many copies of \emph{both} patterns in \Cref{fig:orth-equi-fail-simple}. With high probability, our sampled training set fails in the same way---the learning algorithm can not distinguish it from another training set with different \qmet constraints.

\begin{repthm}[Failure of \orthequi Algorithms]{thm}{thm:orth-equi-bad}{thmQmetGenOrthEqui}
Let $(f_n)_n$ be an \emph{arbitrary} sequence of large values. There is an infinite sequence of \qmet spaces $((\mathcal{X}_n, d_n))_n$ with $\size{\mathcal{X}_n} = n$, $\mathcal{X}_n \subset \R^n$ such that, over a random training set $S$ of size $m$, any \orthequi algorithm outputs a predictor $\smash{\hat{d}}$ that
\begin{itemize}[topsep=-4pt, itemsep=-4pt]
    \item $\smash{\hat{d}}$ fails \emph{non-negativity}, or
    \item $\smash{\max(\dis_S(\hat{d}),\vio(\hat{d})) \geq f_n}$~~(\ie, $\smash{\hat{d}}$ approximates training $S$ badly or  is far from a \qmet),
\end{itemize}
with probability $1/2 - o(1)$,
as long as $S$ does not contain almost all pairs $\smash{1 - m / n^2 = \omega(n^{-1/3})}$, and does not only include few pairs $\smash{m / n^2 = \omega(n^{-1/2})}$.%
 \vspace{-1pt}
\end{repthm}


Furthermore, standard NTK results show that \uncon MLPs trained in NTK regime converge to a function with zero training loss. By the above theorem, the limiting function is not a \qmet with nontrivial probability. In the appendix, we formally state this result. Despite their empirical usages, these results suggest that \uncon networks are likely not suited for \qmet learning.


\subsection{\Qmet Embeddings}\label{sec:theory-qmet-emb}

A \qmet embedding consists of a mapping $f$ from data space $\mathcal{X}$ to a latent \qmet space $(\mathcal{Z}, d_z)$, and predicts $\hat{d}(x, y) \trieq d_z(f(x), f(y))$. Therefore, they always respect all \qmet constraints and attain optimal \vio of value $1$, \emph{regardless of training data}.


However, unlike deep networks, their \dis (approximation) properties depend on the specific latent \qmets. If the latent \qmet is not overly restrictive and can approximate \emph{any} \qmet (with flexible learned encoders), we have nice guarantees for both \dis and \vio.

In the section below, we present \PQE (PQE) as such a latent \qmet, along with its theoretical \dis and \vio guarantees.

\section{\PQEs (PQEs)}

\label{sec:emb}

Motivated by above theoretical findings, we aim to find a latent \qmet space $(\R^d, d_z)$ with a deep network encoder $f\colon \mathcal{X}\rightarrow \R^d$, and a \qmet $d_z$ that is both \emph{universal} and \emph{differentiable}: \begin{itemize}[topsep=-3pt, itemsep=-1.4pt]
    \item for any data \qmet $\smash{(\mathcal{X}, d)}$, there exists an encoder $f$ such that $\smash{d_z(f(x), f(y)) \approx d(x, y)}$;
    \item $d_z$ is differentiable (for optimizing $f$ and possible integration with other gradient-based systems).
\end{itemize}

\begin{notation}
    We use $x, y$ for elements of the data space $\mathcal{X}$, $u, v$ for elements of the latent space $\R^d$, upper-case letters for random variables, and $(\cdot)_z$ for indicating functions in latent space (\eg, $d_z$).
\end{notation}


An existing line of machine learning research learns \emph{\qparts}, or \emph{partial orders}, via Order Embeddings \citep{vendrov2015order}. \Qparts are in fact special cases of \qmets whose distances are restricted to be binary, denoted as $\pi$. An Order Embedding is a representation of a \qpart, where $\pi^\mathsf{OE}(x, y) = 0$ (\ie, $x$ is related to $y$) iff $f(x) \leq f(y)$ coordinate-wise: \begin{equation}
    \pi^\mathsf{OE}(x, y) \trieq \pi_z^\mathsf{OE}(f(x), f(y)) \trieq 1 - \prod_{j} \indic{f(x)_j - f(y)_j \leq 0}.\label{eq:order-emb}
\end{equation}
Order Embedding is \emph{universal} and can model \emph{any \qpart} (see appendix and \citet{hiraguchi1951dimension}).

Can we extend this discrete idea to general continuous \qmets? Quite na\"{i}vely, one may attempt a straightforward soft modification of Order Embedding: \begin{equation}
    \pi_z^\mathsf{SoftOE}(u, v) \trieq 1 - \prod_j\exp\big(-(u_j - v_j)^+\big) = 1 - \exp\Big(-\sum_{j} (u_j - v_j)^+\Big), \label{eq:soft-order-emb}
\end{equation}
which equals $0$ if $u \leq v$ coordinate-wise, and increases to $1$ as some coordinates violate this condition more.
However, it is unclear whether this gives a \qmet.

A more principled way is to parametrize a (scaled) \emph{distribution of latent \qparts} $\Pi_z$, whose expectation naturally gives a continuous-valued \qmet: \begin{equation}
    d_z(u, v; \Pi_z, \alpha) \trieq \alpha \cdot \expect[\pi_z \sim \Pi_z]{\pi_z(u, v)},\mathrlap{\qquad\quad\alpha\geq0.}
\end{equation}

\PQE (PQE) gives a general recipe for constructing such $\Pi_z$ distributions so that $d_z$ is  \emph{universal} and \emph{differentiable}. Within this framework, we will see that $\pi_z^\mathsf{SoftOE}$ is actually a \qmet based on such a distribution and is (almost) sufficient for our needs.

\subsection{Distributions of Latent \Qparts}\label{sec:pqelh}

A random latent \qpart  $\pi_z \colon \R^d \times \R^d \rightarrow \{0, 1\}$ is a difficult object to model, due to complicated \qpart constraints. Fortunately, the Order Embedding representation (\Cref{eq:order-emb}) is without such constraints. If, instead of fixed latents $u, v$, we have \emph{random latents} $R(u), R(v)$, we can compute: \begin{equation}
    \expectnear[\pi_z]{\pi_z(u, v)}
    = \expectnear[R(u), R(v)]{\pi_z^\mathsf{OE}(R(u), R(v))}  = 1 - \prob{R(u) \leq R(v) \text{ coordinate-wise}}.
\end{equation}

In this view, we represent a random $\pi_z$ via a joint distribution of  random vectors\footnote{In general, these random vectors $R(u)$ do not have to be of the same dimension as $u \in \R^d$, although the dimensions do match in the PQE variants we experiment with. }
$\{R(u)\}_{u \in \R^d}$, \ie, a \emph{stochastic process}. To easily compute the probability of this coordinate-wise event, we assume that each dimension of random vectors is from an independent process, and obtain \begin{equation}
    \expect[\pi_z]{\pi_z(u, v)} = 1 - \prod_j \prob{R_j(u) \leq R_j(v)}.
\end{equation}

The choice of stochastic process is flexible. Using \emph{Poisson processes} (with Lebesgue mean measure; \Cref{defn:poisson-process}) that count random points on half-lines\footnote{Half-lines has Lebesgue measure $\infty$. More rigorously, consider using a small value as the lower bounds of these intervals, which leads to same result.} $(-\infty, a]$, we can have $R_j(u) = N_j((\infty, u_j])$,  the (random) count of events in $(\infty, u_j]$ from $j$-th Poisson process: \begin{align}
    \expect[\pi_z \sim \Pi_z]{\pi_z(u, v)}
    & = 1 - \prod_j \mathbb{P}\big[N_j((-\infty, u_j]) \leq N_j((-\infty, v_j])\big]  \\[-0.7ex]
    & = 1 - \prod_j \exp \big(-(u_j - v_j)^+\big)   = \pi_z^\mathsf{SoftOE}(u, v),
\end{align}%
where we used  \Cref{fact:diff-poi-race} and the observation that one half-line is either subset or superset of another. 
Indeed, $\pi_z^\mathsf{SoftOE}$ is an expected \qpart (and thus a \qmet), and is \emph{differentiable}.


Considering a mixture of such distributions for expressiveness, the full latent \qmet formula is \begin{equation}
    d_z^\PQELH(u, v; \alpha) \trieq \sum_i \alpha_i \cdot \Big(1 - \exp \big(-\sum_j(u_{i,j} - v_{i,j})^+\big)\Big),\label{eq:pqe-lebesgue-halfline}\vspace{-1pt}
\end{equation}%
where we slightly abuse notation and consider latents $u$ and $v$ as (reshaped to) 2-dimensional. We will see that this is a special PQE case with \ul{\textbf{L}}ebesgue measure and \ul{\textbf{h}}alf-lines, and thus denoted \PQELH.

\subsection{General PQE Formulation}\label{sec:general-pqe}

We can easily generalize the above idea to independent Poisson processes of general mean measures $\mu_j$ and (sub)set parametrizations $u \rightarrow A_j(u)$, and obtain an expected \qpart as: \begin{align}
    \hspace{-1.2ex}\mathbb{E}_{\pi_z \sim \Pi_z^{\mathsf{PQE}}(\mu, A)}\hspace*{-1pt}[\pi_z(u, v)]
    & \hspace{-2pt}\trieq\hspace{-2pt} 1 - \prod_j \prob{N_j(A_j(u)) \leq N_j(A_j(v))} \\[-0.7ex]
    & \hspace{-2pt}=\hspace{-2pt} 1 - \prod_j \mathbb{P}\Big[
    \pois(\underbrace{\mu_j(A_j(u) \setminus A_j(v))}_{\mathclap{\text{Poisson rate of points landing only in $A_j(u)$}}}) \leq
    \pois(\mu_j(A_j(v) \setminus A_j(u)))
    \Big],\label{eq:pqe-qpart-expect}
\end{align}
which is \emph{differentiable} as long as the measures and set parametrizations are (after set differences). Similarly, considering a mixture gives us an expressive latent \qmet.

\emph{A general PQE latent \qmet} is defined with $\{(\mu_{i,j}, A_{i,j})\}_{i,j}$ and weights $\alpha_i \geq 0$ as:
\begin{align}
    \hspace{-1ex}d_z^{\mathsf{PQE}}(u, v; \mu, A, \alpha)
    & \trieq \sum_i \alpha_i\cdot  \expect[\pi_z \sim \Pi_z^{\mathsf{PQE}}(\mu_i, A_i)]{\pi_z(u, v)} \label{eq:pqe}\\[-0.7ex]
    & = \sum_i \alpha_i \Big(1 - \prod_j \mathbb{P}\Big[
    \pois(\mu_{i,j}(A_{i,j}(u) \setminus A_{i,j}(v))) \leq
    \pois(\mu_{i,j}(A_{i,j}(v) \setminus A_{i,j}(u)))
    \Big]\Big),\notag\vspace{-1pt}
\end{align}%
whose optimizable parameters include $\{\alpha_i\}_i$, possible ones from $\{(\mu_{i,j}, A_{i,j})\}_{i,j}$ (and encoder $f$).

This general recipe can be instantiated in many ways. Setting $A_{i,j}(u) \rightarrow (-\infty, u_{i,j}]$ and Lebesgue $\mu_{i,j}$, recovers \PQELH. In the appendix, we consider a form with \textbf{\ul{G}}aussian-based measures and \textbf{\ul{G}}aussian-shapes, denoted as \PQEGG. Unlike \PQELH, \PQEGG always gives nonzero gradients.

The appendix also includes several implementation techniques that empirically improve stability, including learning $\alpha_i$'s with deep linear networks, a formulation that outputs discounted distance, \etc.


\subsection{Continuous-valued Stochastic Processes}\label{sec:continuous-nondiff}
But why Poisson processes over more common choices such as Gaussian processes?  It turns out that common continuous-value processes fail to give a \emph{differentiable} formula.

Consider a non-degenerate process $\{R(u)\}_{u}$, where $(R(u), R(v))$ has bounded density if $u \neq v$. Perturbing $u \rightarrow {\color{red} u +\delta}$ leaves $\prob{R(u) = R({\color{red}u + \delta})} = 0$. Then one of $\mathbb{P}\big[ R(u) \leq R({\color{red}u + \delta}) \big]$ and $\mathbb{P}\big[ R({\color{red}u + \delta}) \leq R(u) \big]$ must be far away from $1$ (as they sum to $1$), breaking differentiability at $\prob{R(u) \leq R(u)} = 1$. (This argument is 
formalized in the appendix.)
%
Discrete-valued processes, however, can leave most probability mass on $R(u) = R({\color{red}u+\delta})$ and thus remain differentiable.

\subsection{Theoretical Guarantees}\label{sec:pqe-guarantees}

Our PQEs bear similarity with the algorithmic \qmet embedding construction in \citet{memoli2018quasimetric}. Extending their analysis to PQEs, we obtain the following \dis and \vio guarantees.

\begin{repthm}[\Dis and \vio of PQEs]{thm}{thm:pqe-lhgg-low-dis-vio}{thmPQELHGGLowDisVio}
Under the assumptions of \Cref{sec:theory}, \emph{any} \qmet space with size $n$ and treewidth $t$ admits a \PQELH and a \PQEGG with \dis $\smash{\bigO(t \log^2 n)}$ and \vio $1$, with an expressive encoder (\eg, a ReLU network with $\geq3$ layers and polynomial width).
\end{repthm}

In fact, these guarantees apply to any PQE formulation that satisfies a mild condition. Informally, any PQE with $h \times k$ Poisson processes (\ie, $h$ mixtures) enjoys the above guarantees if it can approximate the discrete counterpart: mixtures of $h$ Order Embeddings, each specified with $k$ dimensions. In the appendix, we make this condition precise and provide a full proof of the above theorem.

\section{Experiments}
\label{sec:expr}

Our experiments are designed to (1) confirm our theoretical findings and (2) compare PQEs against a wider range of baselines, across different types of tasks.  In all experiments, we optimize $\gamma$-discounted distances (with $\gamma \in \{0.9, 0.95\}$), and compare the following five families of methods: \begin{itemize}[topsep=-2pt, itemsep=-1pt]
    \item \textbf{PQEs (2 formulations):} \PQELH and \PQEGG with techniques mentioned in \Cref{sec:general-pqe}.
    \item \textbf{\Uncon networks (20 formulations):} Predict raw distance (directly, with $\exp$ transform, and with $\smash{(\cdot)^2}$ transform) or $\gamma$-discounted distance  (directly, and with a sigmoid-transform). Each variant is run with a possible triangle inequality regularizer ${\mathbb{E}_{x,y,z}\big[\max(0, \gamma^{\hat{d}(x, y) + \hat{d}(y, z)} - \gamma^{\hat{d}(x, z)})^2\big]}$ for each of $4$ weights $\in \{0, 0.3, 1, 3\}$.
    \item \textbf{Asymmetrical dot products (20 formulations):} On input pair $(x, y)$, encode each into a feature vector with a \emph{different} network, and take the dot product. Identical to \uncon networks, the output is used in the same $5$ ways, with the same $4$ triangle inequality regularizer options.
    \item \textbf{Metric encoders (4 formulations):} Embed into Euclidean space, $\ell_1$ space, hypersphere with (scaled) spherical distance, or a mixture of all three.
    \item \textbf{DeepNorm (2 formulations) and WideNorm (3 formulations):} \Qmet embedding methods that often require significantly more parameters than PQEs (often on the order of $10^6\sim10^7$ more effective parameters; see the appendix for detailed  comparisons) but can only approximate a subset of all possible \qmets \citep{pitis2020inductive}.
\end{itemize}

We show average results from $5$ runs.
The appendix provides experimental details, full results (including standard deviations), additional experiments, and ablation studies.

\begin{figure}
\vspace{-25pt}
    \begin{subfigure}[t]{0.3\linewidth}
        \centering
        \includegraphics[scale=0.4175, trim=45 18 0 0]{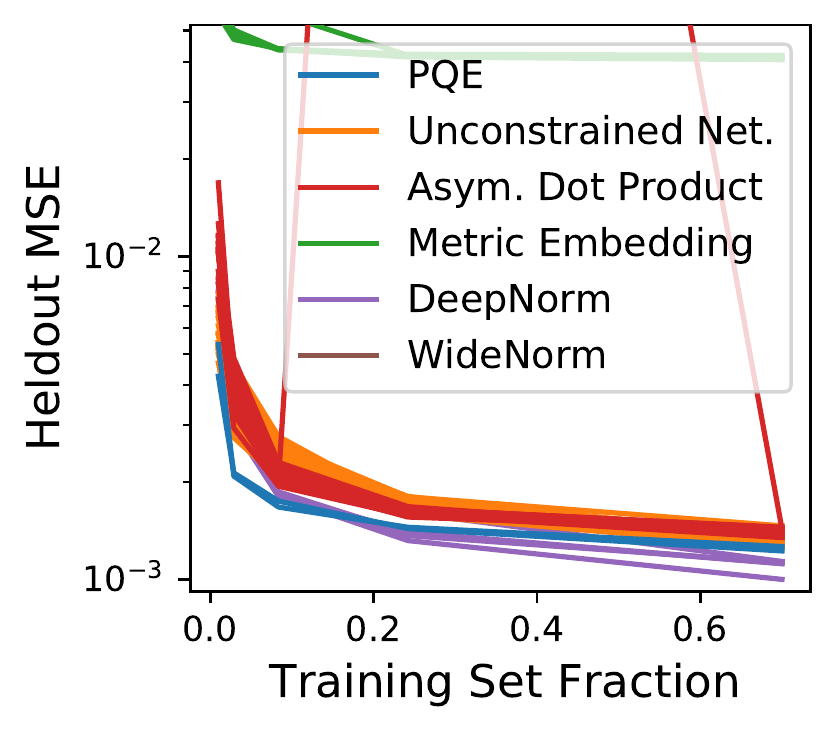}%
        \includegraphics[scale=0.325, trim=15 -15 10 0]{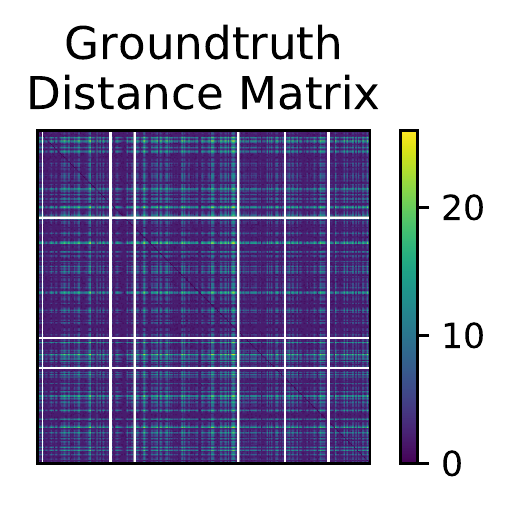}%
        \caption{A dense graph.}
    \end{subfigure}\hfill%
    \begin{subfigure}[t]{0.335\linewidth}
        \centering
        \includegraphics[scale=0.4175, trim=-5 18 0 0]{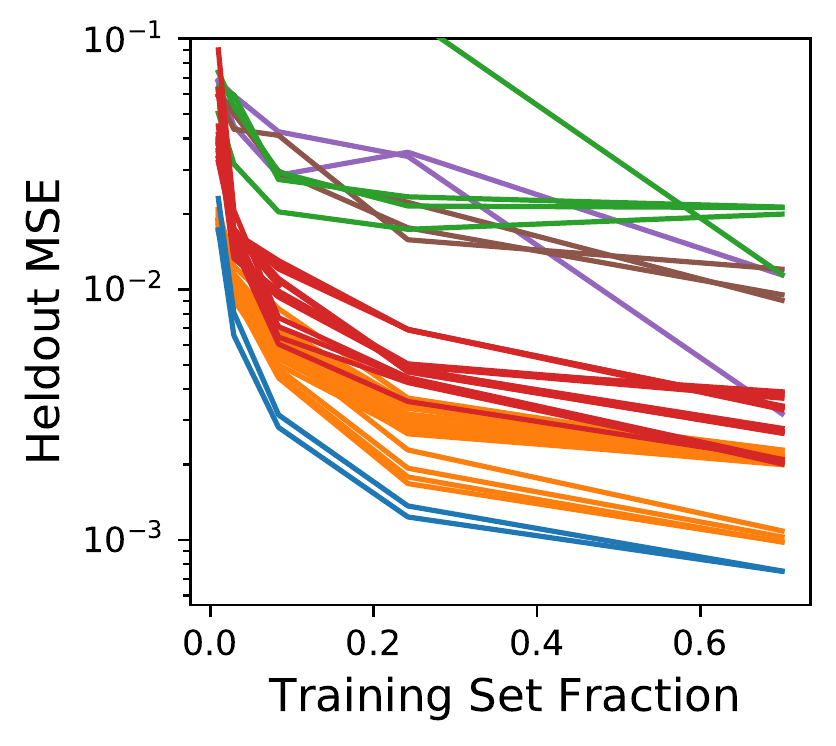}%
        \includegraphics[scale=0.325, trim=15 -15 10 0]{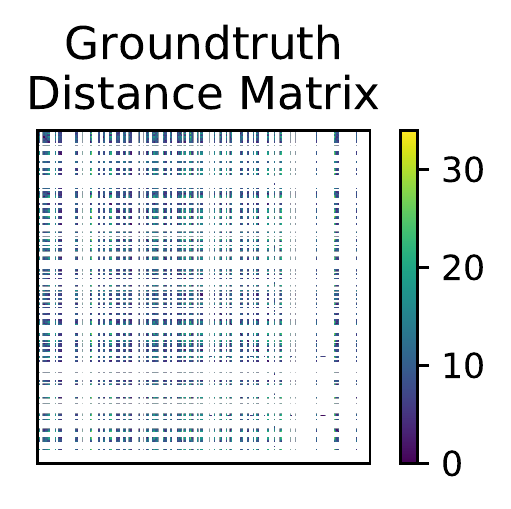}%
        \caption{A sparse graph.}
    \end{subfigure}\hfill%
    \begin{subfigure}[t]{0.36\linewidth}
        \centering
        \includegraphics[scale=0.4175, trim=-20 18 0 0]{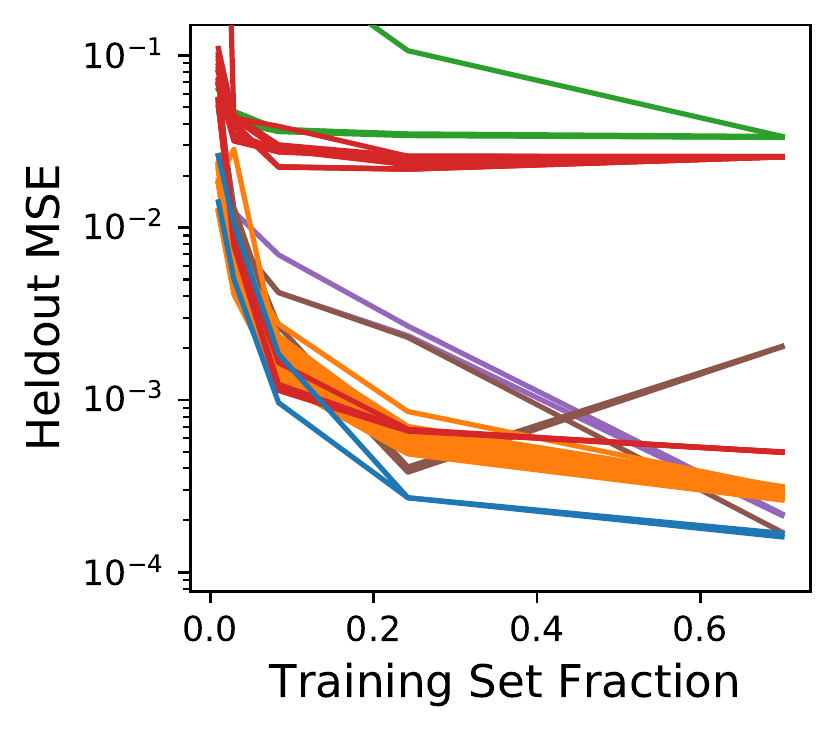}%
        \includegraphics[scale=0.325, trim=15 -15 10 0]{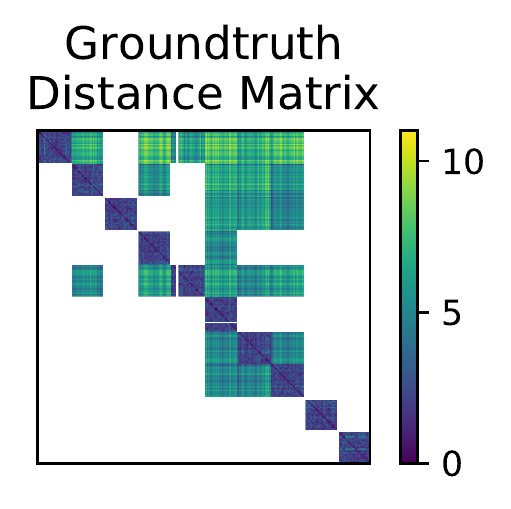}%
        \caption{A sparse graph with block structure.}
    \end{subfigure}\vspace{-7.5pt}
    \caption{Comparison of PQE and baselines on \qmet learning in random directed graphs.}\label{fig:expr-random-graph}\vspace{-8pt}
\end{figure}

\myparagraph{Random directed graphs.}
We start with randomly generated directed graphs of $300$ nodes, with $64$-dimensional node features given by randomly initialized neural networks. After training with MSE on discounted distances, we test the models' prediction error on the unseen pairs (\ie, generalization), measured also by MSE on discounted distances. On three graphs with distinct structures, PQEs significantly outperform baselines across almost all training set sizes (see \Cref{fig:expr-random-graph}). Notably, while DeepNorm and WideNorm do well on the dense graph \qmet, they struggle on the other two, attaining both high test MSE (\Cref{fig:expr-random-graph}) and train MSE (not shown). This is consistent with the fact  that they can only approximate a subset of all \qmets, while PQEs can approximate all \qmets.

\begin{figure}
\vspace{-16.5pt}
\vspace{-3pt}
  \begin{minipage}[b][][b]{.585\linewidth}
    \centering
    \resizebox{
      1\linewidth
    }{!}{%
    \renewcommand\normalsize{\small}%
    \setlength{\tabcolsep}{2.8pt}
    \normalsize
    \centering
    \newcommand{\bftab}{\fontseries{b}\selectfont}%
    \renewcommand{\arraystretch}{1.1}%
    \begin{tabular}[b]{lc|ccc}
        \toprule
        &  \multirow{3}{*}{\vspace{0pt}\shortstack{Triangle\\inequality\\regularizer}}
        &  \multirow{3}{*}{\vspace{0pt}\shortstack{MSE \wrt\\$\gamma$-discounted\\[0.1ex]distances $\smash{(\times 10^{-3})}$ $\boldsymbol\downarrow$}}
        &  \multirow{3}{*}{\vspace{0pt}\shortstack{L1 Error\\when true\\$d < \infty$  $\boldsymbol\downarrow$}}
        &  \multirow{3}{*}{\vspace{0pt}\shortstack{Prediction $\hat{d}$\\when true\\$d = \infty$  $\boldsymbol\uparrow$}}
        \\
        &
        &
        &
        &
        \\
        &
        &
        &
        &   \\
        \midrule
        \midrule

        PQE-LH
        & \xmark
        & \round{3.0426766257733107}{3}
        & \round{1.626337456703186}{3}
        & \round{69.94242248535156}{3}
        \\

        PQE-GG
        & \xmark
        & \round{3.908537095412612}{3}
        & \round{1.895107340812683}{3}
        & \round{101.82395477294922}{3}
        \\

        \midrule

        \multirow{2}{*}{\vspace{0pt}\ul{Best} \Uncon Net.}
        & \xmark
        & \round{3.0862420331686735}{3}
        & \round{2.1151447772979735}{3}
        & \round{59.5242919921875}{3}

        \\
        & \cmark
        & \round{2.8128044679760933}{3}
        & \round{2.210947322845459}{3}
        & \round{61.37094268798828}{3}

        \\
        \midrule

        \multirow{2}{*}{\ul{Best} Asym.~Dot Product}
        & \xmark
        & \round{48.105619102716446}{3}
        & 2.520 $\times 10^{11}$  
        & 2.679 $\times 10^{11}$  

        \\
        & \cmark
        & \round{48.102133721113205}{3}
        & 2.299 $\times 10^{11}$  
        & 2.500 $\times 10^{11}$  
        \\
        \midrule

        \ul{Best} Metric Embedding
        & \xmark
        & \round{17.595218867063522}{3}
        & \round{7.539879989624024}{3}
        & \round{53.85003433227539}{3}
        \\
        \midrule

        \ul{Best} DeepNorm
        & \xmark
        & \round{5.071479082107544}{3}
        & \round{2.0852701663970947}{3}
        & \round{120.04515838623047}{3}
        \\
        \midrule

        \ul{Best} WideNorm
        & \xmark
        & \round{3.532807668671012}{3}
        & \round{1.769381594657898}{3}
        & \round{124.65802764892578}{3}
        \\
        \bottomrule
    \end{tabular}%
    }\vspace{-6pt}
    \captionof{table}
      {%
        \Qmet learning on large-scale web graph. ``\ul{Best}'' is selected by \emph{test} MSE \wrt $\gamma$-discounted distances.%
        \label{fig:expr-social-graph}%
      }%
  \end{minipage}\hfill%
  \begin{minipage}[b][][b]{.415\linewidth}
    \centering
    \includegraphics[scale=0.349, trim=10 24 0 10]{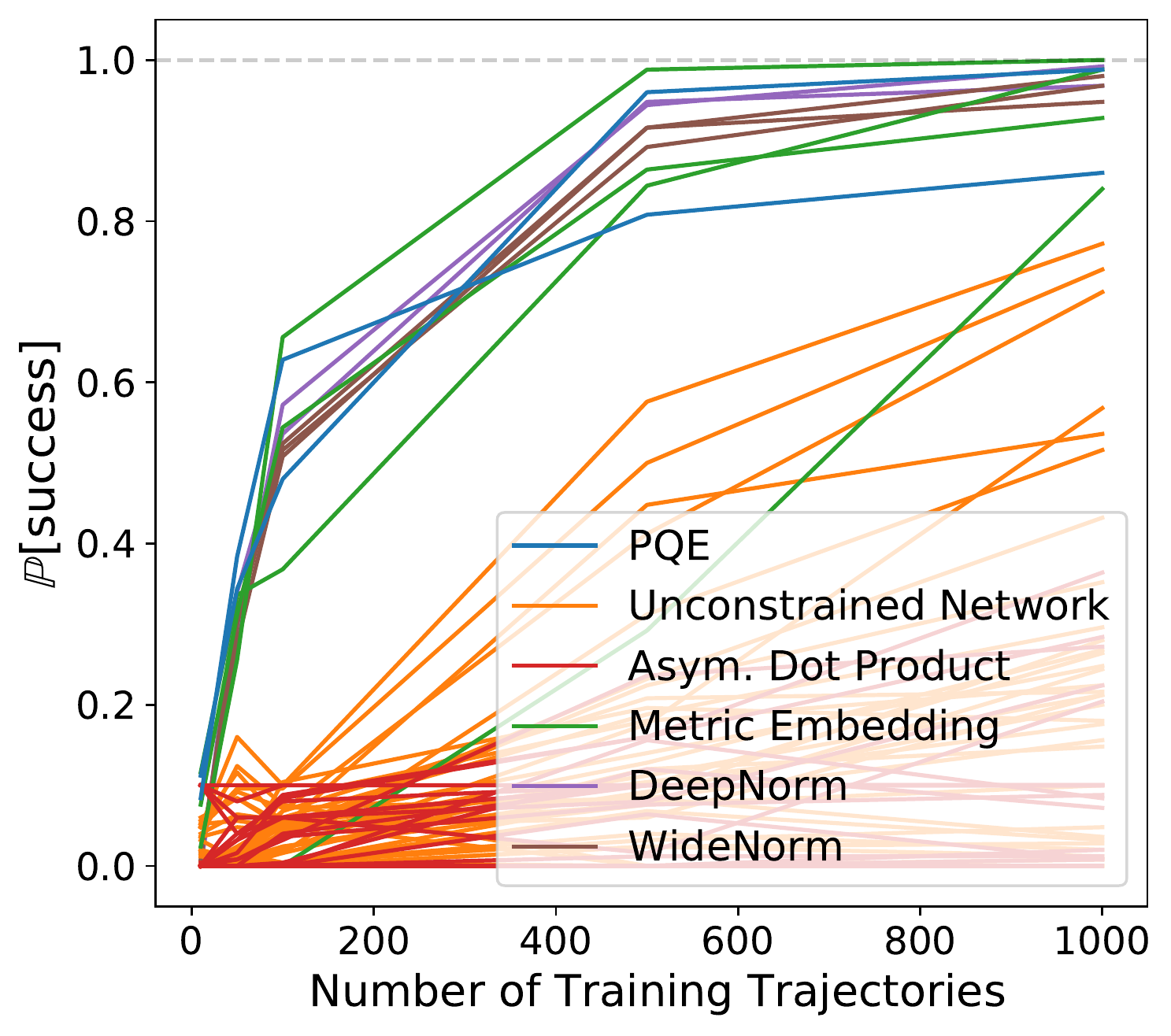}%
    \vspace{-3pt}
    \caption{%
        Offline Q-learning results.%
        \label{fig:expr-offline-q-learning}%
    }
  \end{minipage}
  \vspace{-15pt}
\end{figure}

\myparagraph{Large-scale social graph.}
We choose the $\mathsf{Berkeley\hbox{-}Stanford~Web~Graph}$ \citep{snapnets} as the real-wold social graph for evaluation. This graph consists of $685{,}230$ pages as nodes, and $7{,}600{,}595$ hyperlinks as directed edges. We use $128$-dimensional $\mathsf{node2vec}$ features \citep{grover2016node2vec} and the landmark method \citep{rizi2018shortest} to construct a training set of $2{,}500{,}000$ pairs, and a test set of $150{,}000$ pairs. PQEs generally perform better than other methods, accurately predicting finite distances while predicting high values for infinite distances (see \Cref{fig:expr-social-graph}). DeepNorms and WideNorms learn finite distances less accurately here, and also  do much worse than PQEs on learning the (quasi)metric of an \emph{undirected} social graph (shown in the appendix).

\myparagraph{Offline Q-learning.}
Optimal goal-reaching plan costs in MDPs are \qmets \citep{bertsekas1991analysis,tian2020model} (see also the appendix). In practice, optimizing deep Q-functions often suffers from stability and sample efficiency issues \citep{henderson2018deep,fujimoto2018addressing}.
As a proof of concept, we use PQEs as goal-conditional Q-functions in offline Q-learning, on the grid-world environment with one-way doors built upon \texttt{gym-minigrid} \citep{gym_minigrid} (see \Cref{fig:intro-qmet}~right), following the algorithm and data sampling procedure described in \citet{tian2020model}. Adding strong \qmet structures greatly improves sample efficiency and greedy planning success rates over popular existing approaches such as \uncon networks used in \citet{tian2020model} and asymmetrical dot products used in \citet{schaul2015universal} (see \Cref{fig:expr-offline-q-learning}).
As an interesting observation, some metric embedding formulations work comparably well.

\section{Related Work}\label{sec:related}

\myparagraph{Metric learning.} Metric learning aims to approximate a target metric/similarity function, often via a learned embedding into a metric space. This idea has successful applications in dimensionality reduction \citep{tenenbaum2000global}, information retrieval \citep{wang2014learning}, clustering \citep{xing2002distance}, classification \citep{weinberger2006distance,hoffer2015deep}, 
\etc. While asymmetrical formulations have been explored, they either ignore \qmet constraints \citep{oord2018representation,logeswaran2018efficient,schaul2015universal}, or are not general enough to approximate arbitrary \qmet \citep{balashankar2021learning}, which is the focus of the present paper.

\myparagraph{Isometric embeddings.} Isometric (distance-preserving) embeddings is a highly influential and well-studied topic in mathematics and statistics. Fundamental results, such as Bourgain's random embedding theorem \citep{bourgain1985lipschitz}%
, laid important ground work in understanding and constructing (approximately) isometric embeddings. While most such researches concern metric spaces, \citet{memoli2018quasimetric} study an algorithmic construction of a \qmet embedding via basic blocks called \emph{\qparts}. Their approach requires knowledge of \qmet distances between all pairs and thus is not suitable for learning. Our formulation takes inspiration from the form of their embedding, but is fully learnable with gradient-based optimization over a training subset.

\myparagraph{\Qmets and partial orders.} Partial orders (\qparts) are special cases of \qmets (see \Cref{sec:emb}). A line of machine learning research studies embedding partial order structures into latent spaces for tasks such as relation discovery and information retrieval \citep{vendrov2015order,suzuki2019hyperbolic,hata2020nested,ganea2018hyperbolic}. Unfortunately, unlike PQEs, such formulations do not straightforwardly generalize to arbitrary \qmets, which are more than binary relations. Similar to PQEs, DeepNorm and WideNorm are \qmet embedding approaches learnable with gradient-based optimization \citep{pitis2020inductive}. Theoreically, they universally approximates  a subset of \qmets (ones induced by asymmetrical norms). Despite often using many more parameters, they are restricted to this subset and unable to approximate general \qmets like PQEs do (\Cref{fig:expr-random-graph}).



\section{Implications}
\label{sec:implications}

In this work, we study \qmet learning via both theoretical analysis and empirical evaluations.

Theoretically, we show strong negative results for a common family of learning algorithms, and positive guarantees for our proposed \PQE (PQE). Our results introduce the novel concept of equivariant learning algorithms, which may potentially be used for other learnability analyses with algorithms such as deep neural networks. Additionally, a thorough average-case or data-dependent analysis would nicely complement our results, and may shed light on conditions where algorithms like deep networks can learn decent approximations to \qmets in practice.

PQEs are the first \qmet embedding formulation that can be learned via gradient-based optimization. Empirically, PQEs show promising performance in various tasks. Furthermore, PQEs are fully differentiable, and (implicitly) enforce a \qmet structure in any latent space. They are particularly suited for integration in large deep learning systems, as we explore in the Q-learning experiments. This can potentially open the gate to many practical applications such as better embedding for planning with MDPs, efficient shortest path finding via learned \qmet heuristics, representation learning with \qmet similarities, causal relation learning, \etc.




\newpage\clearpage

\bibliography{reference}
\bibliographystyle{iclr2022_conference}


\newpage\clearpage
\appendix
\section*{Appendix}
\startcontents[sections]
\printcontents[sections]{l}{1}{\setcounter{tocdepth}{2}}
\newpage\clearpage

\begingroup
\newcommand{\appdefn}{1}

\section{Discussions for \texorpdfstring{\Cref*{sec:preliminaries}: \nameref*{sec:preliminaries}}{Preliminaries Section}}

\subsection{\Qmet Spaces}

\begin{repthm}[\Qmet Space]{defn}{defn:qmet-space}{defnQmetSpace}
    A \emph{\qmet space} is a pair $(\mathcal{X}, d)$, where $\mathcal{X}$ is a set of points and $d \colon \mathcal{X} \times \mathcal{X} \rightarrow [0, \infty]$ is the \qmet, satisfying the following conditions: \begin{align}
        \forall x, y \in \mathcal{X}, \qquad & x = y \iff d(x, y) = 0,\hspace{-2.2cm}  \tag{\idtindis} \\[-0.2ex]
        \forall x, y, z \in \mathcal{X}, \qquad & d(x, y) + d(y, z) \geq d(x, z).\hspace{-2.2cm} \tag{Triangle Inequality}
    \end{align}
\end{repthm}

\begin{defn}[\Qpmet Space]\label{defn:qpmet-space}
As a further generalization,  we say $(\mathcal{X}, d)$ is a \emph{\qpmet space} if the \emph{\idtindis} requirement is only satisfied in one direction:
\begin{align}
        \forall x, y \in \mathcal{X}, \qquad & x = y \implies d(x, y) = 0,\hspace{-2.2cm}  \tag{\idtindis} \\[-0.2ex]
        \forall x, y, z \in \mathcal{X}, \qquad & d(x, y) + d(y, z) \geq d(x, z).\hspace{-2.2cm} \tag{Triangle Inequality}
\end{align}
\end{defn}


\subsubsection{Examples of \Qmet Spaces}


\begin{prop}[Expected Hitting Time of a Markov Chain]\label{prop:mc-hitting-t-qmet}
    Let random variables $(X_t)_t$ be a Markov Chain with support $\mathcal{X}$. Then $(\mathcal{X}, d_\mathsf{hitting})$ is a \qmet space, where \begin{equation}
        d_\mathsf{hitting}(s, t) \trieq \expect{\text{time to hit $t$} \given \text{start from $s$}},
    \end{equation}
    where we define the hitting time of $s$ starting from $s$ to be $0$.
\end{prop}
\begin{proof}[Proof of \Cref{prop:mc-hitting-t-qmet}]
    Obviously $d_\mathsf{hitting}$ is non-negative. We then verify the following \qmet space properties: \begin{itemize}
        \item \textbf{\idtindis.}
        By definition, we have, $\forall x, y \in \mathcal{X}$, $ x \neq y$, \begin{align}
            d_\mathsf{hitting}(x, x) & = 0 \\
            d_\mathsf{hitting}(x, y) & \geq 1.
        \end{align}

        \item \textbf{Triangle Inequality.}
        For any $x, y, z \in \mathcal{X}$, we have \begin{align}
            d_\mathsf{hitting}(x, y) + d_\mathsf{hitting}(y, z )
            & =  \expect{\text{time to hit $y$ then hit $z$} \given \text{start from $x$}} \\
            & \geq \expect{\text{time to hit $z$} \given \text{start from $x$}} \\
            & = d_\mathsf{hitting}(x, z).
        \end{align}
    \end{itemize}
    Hence, $(\mathcal{X}, d_\mathsf{hitting})$ is a \qmet space.
\end{proof}

\begin{prop}[Conditional Shannon Entropy]\label{prop:cond-ent-qmet}
    Let $\mathcal{X}$ be the set of random variables (of some probability space). Then $(\mathcal{X}, d_H)$ is a \qpmet space, where \begin{equation}
        d_H(X, Y) \trieq H(Y \given X).
    \end{equation}
    If for all distinct $(X, Y) \in \mathcal{X}\times\mathcal{X}$, $X$ can not be written as (almost surely) a deterministic function of $Y$, then $(\mathcal{X}, d_H)$ is a \qmet space.
\end{prop}
\begin{proof}[Proof of \Cref{prop:cond-ent-qmet}]
    Obviously $d_H$ is non-negative. We then verify the following \qpmet space properties: \begin{itemize}
        \item \textbf{\idtindis.}
        By definition, we have, $\forall X, Y \in \mathcal{X}$, \begin{align}
            d_H(X, X) & = H(X \given X) = 0 \\
            d_H(Y, X) & = H(Y \given X) \geq 0,
        \end{align}
        where $\leq$ is $=$ iff $Y$ is a (almost surely) deterministic function of $X$.

        \item \textbf{Triangle Inequality.}
        For any $X, Y, Z \in \mathcal{X}$, we have \begin{align}
            d_H(X, Y) + d_H(Y, Z)
            & = H(Y \given X)  + H(Z \given Y) \\
            & \geq H(Y \given X)  + H(Z \given XY) \\
            & = H(YZ \given X) \\
            & \geq H(Z \given X) \\
            & = d_H(X, Z).
        \end{align}
    \end{itemize}
    Hence, $(\mathcal{X}, d_H)$ is a \qpmet space, and a \qmet space when the last condition is satisfied.
\end{proof}

\paragraph{Conditional Kolmogorov Complexity}
From algorithmic information theory, the conditional Kolmogorov complexity $K(y \given x)$ also similarly measures ``the bits needed to create $y$ given $x$ as input'' \citep{kolmogorov1963tables}. It is also almost a \qmet, but the exact definition affects some constant/log terms that may make the \qmet constraints non-exact. For instance, when defined with the prefix-free version, conditional Kolmogorov complexity is always strictly positive, even for $K(x \given x) > 0$ \citep{li2008introduction}. One may remedy this with a definition using a universal Turing machine (UTM) that simply outputs the input on empty program. But to make triangle inequality work, one needs to reason about how the input and output parts work on the tape(s) of the UTM. Nonetheless, regardless of the definition details, conditional Kolmogorov complexity do satisfy a triangle inequality up to log terms \citep{grunwald2004shannon}. So intuitively, it behaves roughly like a \qmet defined on the space of binary strings.

\paragraph{Optimal Goal-Reaching Plan Costs in Markov Decision Processes (MDPs)}

We define MDPs in the standard manner: $\mathcal{M} = (\mathcal{S}, \mathcal{A}, \mathcal{R}, \mathcal{P}, \gamma)$ \citep{puterman1994mdp}, where $\mathcal{S}$ is the state space, $\mathcal{A}$ is the action space, $\mathcal{R} \colon \mathcal{S} \times \mathcal{A} \rightarrow \R$ is the reward function, $\mathcal{P} \colon \mathcal{S} \times \mathcal{A} \rightarrow \Delta(\mathcal{S})$ is the transition function (where $\Delta(\mathcal{S})$ is the set of all distributions over $\mathcal{S}$), and $\gamma \in (0, 1)$ is the discount factor.

We define $\Pi$ as the collection of all stationary policies $\pi \colon \mathcal{S} \rightarrow \Delta(\mathcal{A})$ on $\mathcal{M}$. For a particular policy $\pi \in \Pi$, it induces random \emph{trajectories}: \begin{itemize}
    \item \emph{Trajectory} starting from state $s\in \mathcal{S}$ is the random variable \begin{equation}
        \xi_\pi(s) = (s_1, a_1, r_1, s_2, a_2, r_2, \dots),
    \end{equation}
    distributed as \begin{align}
        s_1 & = s \\
        a_i & \sim \mathrlap{\pi(s_i),}\hphantom{\mathcal{P}(s_i, a_i), } \qquad \forall i \geq 1 \\
        s_{i+1} & \sim\mathrlap{\mathcal{P}(s_i, a_i),}\hphantom{\mathcal{P}(s_i, a_i), } \qquad \forall i \geq 1.
    \end{align}
    \item \emph{Trajectory} starting from state-action pair $(s, a) \in \mathcal{S} \times \mathcal{A}$ is the random variable  \begin{equation}
        \xi_\pi(s, a) = (s_1, a_1, r_1, s_2, a_2, r_2, \dots),
    \end{equation}
    distributed as \begin{align}
        s_1 & = s \\
        a_1 & = a \\
        a_i & \sim \mathrlap{\pi(s_i),}\hphantom{\mathcal{P}(s_i, a_i),} \qquad \forall i \geq 2 \\
        s_{i+1} & \sim\mathrlap{\mathcal{P}(s_i, a_i),}\hphantom{\mathcal{P}(s_i, a_i), } \qquad \forall i \geq 1.
    \end{align}
\end{itemize}

\begin{prop}[Optimal Goal-Reaching Plan Costs in MDPs]\label{prop:opt-goal-reaching-plan-costs-qmet}
    Consider an MDP $\mathcal{M} = (\mathcal{S}, \mathcal{A}, \mathcal{R}, \mathcal{P}, \gamma)$. WLOG, assume that $\mathcal{R} \colon \mathcal{S} \times \mathcal{A} \rightarrow (-\infty, 0]$ has only non-positive rewards (\ie, negated costs). Let  $\mathcal{X} = \mathcal{S} \cup (\mathcal{S}\times \mathcal{A})$. Then $(\mathcal{X}, d_\mathsf{sum})$ and  $(\mathcal{X}, d_\gamma)$ are \qpmet spaces, where \begin{align}
        d_\mathsf{sum}(x, y)
        &\trieq \min_{\pi \in \Pi} \expect{\text{total costs from $x$ to $y$ under $\pi$}} \\
        &= \begin{cases}
            \min_{\pi \in \Pi}
            \mathbb{E}_{(s_1, a_1, r_1, \dots) = \xi_\pi(x)} \big[
            -\sum_t r_t \underbrace{\indic{s' \notin \{s_i\}_{i \in [t]}}}_\textup{not reached $s'$ yet}
            \big] & \text{if $\underbrace{\vphantom{\indic{s' \notin \{s_i\}_{i \in [t]}}}y = s' \in \mathcal{S}}_\textup{goal is a state$\vphantom{s'}$}$,} \\
            \min_{\pi \in \Pi}
            \mathbb{E}_{(s_1, a_1, r_1, \dots) = \xi_\pi(x)} \big[
            -\sum_t r_t \underbrace{\indic{(s', a') \notin \{(s_i, a_i)\}_{i \in [t-1]}}}_{\mathclap{\textup{not reached $s'$ \emph{\ul{and}} performed $a'$ yet}}}
            \big] & \text{if $\underbrace{\vphantom{\indic{(s', a') \notin \{(s_i, a_i)\}_{i \in [t-1]}}}y = (s', a') \in \mathcal{S} \times \mathcal{A}}_{\mathclap{\textup{goal is a state-action pair $\vphantom{s'}$}}}$,}
        \end{cases}
    \end{align}
    and \begin{equation}
        d_\gamma(x, y)
        \trieq \log_\gamma \max_{\pi \in \Pi} \expect{\gamma^{\text{total costs from $x$ to $y$ under $\pi$}}}
    \end{equation}
    is defined similarly.

    If the reward function is always \emph{negative}, $(\mathcal{X}, d_\mathsf{sum})$ and  $(\mathcal{X}, d_\gamma)$ are \emph{\qmet} spaces.
\end{prop}

\begin{proof}[Proof of \Cref{prop:opt-goal-reaching-plan-costs-qmet}]
    Obviously both $d_\mathsf{sum}$ and $d_\gamma$ are non-negative, and satisfy \emph{\idtindis} (for \qpmet spaces). For triangle inequality, note that for each $y$, we can instead consider alternative MDPs: \begin{itemize}
        \item If $y = s' \in \mathcal{S}$, modify the original MDP to make $s'$ a sink state, where performing any action yields $0$ reward (\ie, $0$ cost);
        \item If $y = (s', a') \in \mathcal{S} \times \mathcal{A}$, modify the original MDP such that performing action $a'$ in state $s'$ surely transitions to a new sink state, where performing any action yields $0$ reward (\ie, $0$ cost).
    \end{itemize}
    Obviously, both are Markovian. Furthermore, they are Stochastic Shortest Path problems with no negative costs \citep{guillot2020stochastic}, implying that there are Markovian (\ie, stationary) optimal policies (respectively \wrt either minimizing expected total cost or maximizing expected $\gamma^\text{total cost}$). Thus optimizing over the set of stationary policies, $\Pi$, gives the optimal quantity over all possible policies, including concatenation of two stationary policies. Thus the triangle inequality is satisfied by both.

    Hence, $(\mathcal{X}, d_\mathsf{sum})$ and  $(\mathcal{X}, d_\gamma)$ are \qpmet spaces.

    Finally, if the reward function is always \emph{negative}, $x \neq y \implies d_\mathsf{sum}(x, y) > 0 \text{ and }d_\mathsf{\gamma}(x, y) > 0 $, so $(\mathcal{X}, d_\mathsf{sum})$ and  $(\mathcal{X}, d_\gamma)$ are \qmet spaces.
\end{proof}

\begin{rmk}\label{rmk:mdp-qmet}
We make a couple remarks:
\begin{itemize}
    \item Any MDP with a bounded reward function can be modified to have only non-positive rewards by subtracting the maximum reward (or larger);
    \item We have \begin{equation}
        d_\mathsf{sum}(s, (s, a)) = d_\gamma(s, (s, a)) = -\mathcal{R}(s, a).
    \end{equation}
    \item When the dynamics is deterministic, $d_\mathsf{sum} \equiv d_\gamma$, $\forall \gamma \in (0, 1)$.
    \item Unless $y$ is reachable from $x$ with probability $1$ under some policy, $d_\mathsf{sum}(x, y) = \infty$.
    \item Unless $y$ is \emph{unreachable} from $x$ with probability $1$ under \emph{all} policies, $d_\mathsf{sum}(x, y) < \infty$. Therefore, it is often favorable to consider $d_\gamma$ types.
    \item In certain MDP formulations, the reward is stochastic and/or dependent on the reached next state. The above definitions readily extend to those cases.
    \item $\gamma^{d_\gamma((s, a), y)}$ is very similar to Q-functions except that Q-function applies discount based on time, and $\gamma^{d_\gamma((s, a), y)}$ applies discount based on costs. We note that a Q-learning-like recurrence can also be found for $\gamma^{d_\gamma((s, a), y)}$.

    If the cost is constant in the sense for some fixed $c < 0$, $\mathcal{R}(s, a) = c$, $\forall (s, a) \in \mathcal{S} \times \mathcal{A}$, then time and cost are equivalent up to a scale. Therefore,  $\gamma^{d_\gamma((s, a), y)}$ coincides with the optimal Q-functions for the MDPs described in proof, and $\gamma^{d_\gamma(s, y)}$ coincides with the optimal value functions for the respective MDPs.
\end{itemize}
\end{rmk}

\subsubsection{\Qmet \Tw and Graph \Tw}


\begin{repthm}[Treewidth of \Qmet Spaces \citep{memoli2018quasimetric}]{defn}{defn:qmet-tw}{defnQmetTw}
\ifdefined\appdefn%
    Consider representations of a \qmet space $M$ as shortest-path distances on a positively-weighted directed graph.  \emph{\Tw} of $M$ is the minimum over all such graphs' \tws. (Recall that the \tw of a graph (after replacing directed edges with undirected ones) is a measure of its complexity.)
\else%
    Consider a \qmet space $M$ as shortest-path distances on a positively-weighted directed graph.  \emph{\Tw} of $M$ is the minimum over all such graphs' \tws.
\fi
\end{repthm}

Graph \tw is a standard complexity measure of how ``similar'' a graph is to a tree \citep{robertson1984graph}. Informally speaking, if a graph has low \tw, we can represent it as a tree, preserving all connected paths between vertices,  except that in each tree node, we store a small number of vertices (from the original graph) rather than just $1$.

Graph \tw is widely used by the Theoretical Computer Science and Graph Theory communities, since many NP problems are solvable in polynomial time for graphs with bounded \tw  \citep{bertele1973non}.

\subsection{Poisson Processes}
\begin{repthm}[Poisson Process]{defn}{defn:poisson-process}{defnPoissonProcess}
    For nonatomic measure $\mu$ on set $A$, a \emph{Poisson process} on $A$ with \emph{mean measure} $\mu$ is a random countable subset $P \subset A$ (\ie, the  random events / points) such that \begin{itemize}[topsep=-3pt, itemsep=-2.5pt]
        \item for any disjoint measurable subsets $A_1, \dots, A_n$ of $A$, the random variables $N(A_1), \dots, N(A_n)$ are independent, where $\smash{N(B) \trieq \#\{P \cap B\}}$ is 
        the number of points of $P$ in $B$, and
        \item $N(B)$ has the Poisson distribution with mean $\mu(B)$, denoted as $\pois(\mu(B))$.
    \end{itemize}
\end{repthm}

Poisson processes are usually used to model events that randomly happens ``with no clear pattern'', \eg, visible stars in a patch of the sky, arrival times of Internet packages to a data center. These events may randomly happen all over the sky / time. To an extent, we can say that their characteristic feature is a property of statistical independence \citep{kingman2005p}.

To understand this, imagine raindrops hitting the windshield of a car. Suppose that we already know that the rain is heavy, knowing the exact pattern of the raindrops  hitting on the left side of the windshield tells you little about the hitting pattern on the right side. Then, we may assume that, as long as we look at regions that are disjoint on the windshield, the number of raindrops in each region are independent.

This is the fundamental motivation of Poisson processes. In a sense, from this characterization, Poisson processes are inevitable (see Sec.~1.4 of \citep{kingman2005p}).

\subsubsection{Poisson Race Probability $\prob{\pois(\mu_1) \leq \pois(\mu_2)}$ and Its Gradient Formulas}

In \Cref{fact:diff-poi-race} we made several remarks on the Poisson race probability, \ie, for \emph{independent} $X \sim \pois(\mu_1)$, $Y \sim \pois(\mu_2)$, the quantity $\prob{X \leq Y}$. In this section, we detailedly describe how we arrived at those conclusions, and provide the exact gradient formulas for differentiating $\prob{X \leq Y}$ \wrt $\mu_1$ and $\mu_2$.

\paragraph{From Skellam distribution CDF to Non-Central $\chi^2$ distribution CDF.} Distribution of the difference of two independent Poisson random variables is called the \emph{Skellam} distribution \citep{skellam1946frequency}, with its parameter being the rate of the two Poissons. That is, $X - Y \sim \mathrm{Skellam}(\mu_1, \mu_2)$. Therefore, $\prob{X \leq Y}$ is essentially the cumulative distribution function (CDF) of this Skellam at $0$. In Eq.~(4) of \citep{johnson1959poissonchisq}, a connection is made between the CDF of $\mathrm{Skellam}(\mu_1, \mu_2)$ distribution, and the CDF of a non-central $\chi^2$ distribution (which is a non-centered generalization of $\chi^2$ distribution) with two parameters $k > 0$ degree(s) of freedom and non-centrality parameter $\lambda \geq 0$): for integer $n > 0$, \begin{equation}
    \prob{\mathrm{Skellam}(\mu_1, \mu_2) \geq n} = \mathbb{P}\big[ \mathrm{NonCentral}\chi^2(\underbrace{2n\vphantom{\mu_2}}_{\mathclap{\textup{degree(s) of freedom}\qquad\qquad\quad}}, \underbrace{2\mu_2}_{\mathclap{\qquad\qquad\qquad\textup{non-centrality parameter}}}) < 2\mu_1 \big], \label{eq:skellam-ncx2-cdf-link}
\end{equation}%
which can be evaluated using statistical computing packages such as \texttt{SciPy} \citep{2020SciPy-NMeth} and \texttt{CDFLIB} \citep{cdflibcodepage,cdflib}.

\paragraph{Marcum-Q-Function and gradient formulas.} To differentiate through \Cref{eq:skellam-ncx2-cdf-link}, we consider representing the non-central $\chi^2$ CDF as a Marcum-Q-function \citep{marcum1950q}. One definition of the Marcum-Q-function $Q_M \colon \R \times \R \rightarrow \R$ in statistics is \begin{equation}
    Q_M(a, b) \trieq \int_b^\infty x \left(\frac{x}{a}\right)^{M-1}\exp \left(  -\frac{x^2+a^2}{2} \right) I_{M-1}(ax) \diff x,
\end{equation}
where $I_{M-1}$ is the modified Bessel function of order $M - 1$. (When $M$ is non-integer, we refer readers to \citep{Brychkov2012marcumq,marcum1950q} for definitions, which are not relevant to the discussion below.) When used in CDF of non-central $\chi^2$, we have \begin{equation}
    \prob{ \mathrm{NonCentral}\chi^2(k, \lambda) < x}  = 1 - Q_{\frac{k}{2}}(\sqrt{\lambda}, \sqrt{x}).
\end{equation}
Combining with \Cref{eq:skellam-ncx2-cdf-link}, and using the symmetry $\mathrm{Skellam}(\mu_1, \mu_2) \stackrel{d}{=} -\mathrm{Skellam}(\mu_2, \mu_1)$, we have, for integer $n$, \begin{align}
    \prob{X \leq Y + n}
    & = \prob{\mathrm{Skellam}(\mu_1, \mu_2) \leq n} \\
    & = \begin{cases}
        \prob{ \mathrm{NonCentral}\chi^2(-2n, 2\mu_1) < 2\mu_2} & \text{if $n < 0$} \\
        1 - \prob{ \mathrm{NonCentral}\chi^2(2(n+1), 2\mu_2) < 2\mu_1} & \text{if $n \geq 0$} \\
    \end{cases} \\
    & = \begin{cases}
        1 - Q_{-n}(\sqrt{2 \mu_1}, \sqrt{2 \mu_2}) & \text{if $n < 0$} \\
        Q_{n+1}(\sqrt{2 \mu_2}, \sqrt{2 \mu_1})  & \text{if $n \geq 0$}. \\
    \end{cases}
\end{align}
Prior work \citep{Brychkov2012marcumq} provides several derivative formula for the Marcum-Q-Function:
\begin{itemize}
    \item For $n < 0$, we have
    \begin{align}
        \frac{\partial}{\partial \mu_1} \prob{X \leq Y + n}
        & = \frac{\partial}{\partial \mu_1} \left( 1 - Q_{-n}(\sqrt{2 \mu_1}, \sqrt{2 \mu_2}) \right) \\
        & = Q_{-n}(\sqrt{2 \mu_1}, \sqrt{2 \mu_2}) - Q_{-n+1}(\sqrt{2 \mu_1}, \sqrt{2 \mu_2}) \tag{Eq.~(16) of \citep{Brychkov2012marcumq}} \\
        & = -\left(\frac{\mu_2}{\mu_1}\right)^{-\frac{n}{2}} e^{-(\mu_1 + \mu_2)} I_{-n}(2\sqrt{\mu_1 \mu_2}) \tag{Eq.~(2) of \citep{Brychkov2012marcumq}} \\
        & = -\left(\frac{\mu_2}{\mu_1}\right)^{-\frac{n}{2}} e^{-(\sqrt{\mu_1} - \sqrt{\mu_2})^2} I^{(e)}_{-n}(2\sqrt{\mu_1 \mu_2}),
    \end{align}
    where $I_v^{(e)}(x) \trieq e^{-\abs{x}} I_v(x)$ is the exponentially-scaled version of $I_v$ that computing libraries often provide due to its superior numerical precision (\eg, \texttt{SciPy} \citep{2020SciPy-NMeth}),
    \begin{align}
        \frac{\partial}{\partial \mu_2} \prob{X \leq Y + n}
        & = \frac{\partial}{\partial \mu_2} \left( 1 - Q_{-n}(\sqrt{2 \mu_1}, \sqrt{2 \mu_2}) \right) \\
        & = \left(\frac{\mu_2}{\mu_1}\right)^{-\frac{n+1}{2}} e^{-(\mu_1+\mu_2)} I_{-n-1}(2\sqrt{\mu_1\mu_2}) \tag{Eq.~(19) of \citep{Brychkov2012marcumq}} \\
        & = \left(\frac{\mu_2}{\mu_1}\right)^{-\frac{n+1}{2}} e^{-(\sqrt{\mu_1} - \sqrt{\mu_2})^2}  I^{(e)}_{-n-1}(2\sqrt{\mu_1\mu_2}),
    \end{align}

    \item For $n \geq 0$, we have
    \begin{align}
        \frac{\partial}{\partial \mu_1} \prob{X \leq Y + n}
        & = \frac{\partial}{\partial \mu_1} Q_{n+1}(\sqrt{2 \mu_2}, \sqrt{2 \mu_1}) \\
        & = - \left(\frac{\mu_1}{\mu_2}\right)^{n} e^{-(\mu_1+\mu_2)} I_{n}(2\sqrt{\mu_1\mu_2}) \tag{Eq.~(19) of \citep{Brychkov2012marcumq}} \\
        & = - \left(\frac{\mu_1}{\mu_2}\right)^{n} e^{-(\sqrt{\mu_1} - \sqrt{\mu_2})^2} I^{(e)}_{n}(2\sqrt{\mu_1\mu_2}),
    \end{align}
    and,
    \begin{align}
        \frac{\partial}{\partial \mu_2} \prob{X \leq Y + n}
        & = \frac{\partial}{\partial \mu_2} Q_{n+1}(\sqrt{2 \mu_2}, \sqrt{2 \mu_1}) \\
        & = Q_{n+2}(\sqrt{2 \mu_2}, \sqrt{2 \mu_1}) - Q_{n+1}(\sqrt{2 \mu_2}, \sqrt{2 \mu_1}) \tag{Eq.~(16) of \citep{Brychkov2012marcumq}} \\
        & = \left(\frac{\mu_1}{\mu_2}\right)^{\frac{n+1}{2}} e^{-(\mu_1 + \mu_2)} I_{n+1}(2\sqrt{\mu_1 \mu_2}) \tag{Eq.~(2) of \citep{Brychkov2012marcumq}} \\
        & = \left(\frac{\mu_1}{\mu_2}\right)^{\frac{n+1}{2}} e^{-(\sqrt{\mu_1} - \sqrt{\mu_2})^2} I^{(e)}_{n+1}(2\sqrt{\mu_1 \mu_2}).
    \end{align}
\end{itemize}
Setting $n = 0$ gives the proper forward and backward formulas for $\prob{X \leq Y}$.

\section{Proofs, Discussions and Additional Results for \texorpdfstring{\Cref*{sec:theory}: \nameref*{sec:theory}}{Theory Section}}


\paragraph{Assumptions.} Recall that we assumed a \qmet space, which is stronger than a \qpmet space (\Cref{defn:qpmet-space}), with finite distances.  These are rather mild assumptions, since any \qpmet with infinities can always be modified to obey these assumptions by (1) adding a small metric (\eg, $d_\eps(x, y) \trieq \eps \indic{x \neq y}$ with small $\eps > 0$) and (2) capping the infinite distances to a large value higher than any finite distance.

\paragraph{Worst-case analysis. }In this work we focus on the \emph{worst-case} scenario, as is common in standard (quasi)metric embedding analyses \citep{bourgain1985lipschitz,johnson1984extensions,indyk2001algorithmic,memoli2018quasimetric}. Such results are important because embeddings are often used as heuristics in downstream tasks (\eg, planning) which are sensitive to any error. While our negative result readily extends to the average-case scenario (since the error (\dis or \vio) is arbitrary), we leave a thorough average-case analysis as future work.

\paragraph{Data-independent bounds. }
We analyze possible \emph{data-independent} bounds for various algorithms. In this sense, the positive result for PQEs (\Cref{thm:pqe-general-low-dis-vio}) is really strong, showing good guarantees \emph{regardless data \qmet}. The negative result (\Cref{thm:orth-equi-bad}) is also revealing, indicating that a family of algorithms should probably not be used, unless we know something more about data. \emph{Data-independent} bounds are often of great interest in machine learning (\eg, concepts of VC-dimension \citep{vapnik2015uniform} and PAC learning \citep{valiant1984theory}). An important future work is to explore data-dependent results, possibly via defining a \qmet complexity metric that is both friendly for machine learning analysis, and connects well with combinatorics measures such as \qmet \tw.

\paragraph{\Vio and \dis metrics.} The optimal \vio has value $1$.  Specifically, it is $1$ iff $\hat{d}$ is a \qmet on $\mathcal{X}$  (assuming \emph{non-negativity}).
\Dis (over training set) and \vio together quantify how well $\hat{d}$ learns a \qmet consistent with the training data. A predictor can fit training data well (low \dis), but ignores basic \qmet constraints on heldout data (high \vio). Conversely, a predictor can perfectly obey the training data constraints (low \vio), but doesn't actually fit training data well (high \dis). Indeed, (assuming \emph{non-negativity} and \emph{\idtindis}), perfect \dis (value $1$) and violation (value $1$) imply that $\hat{d}$ is a \qmet consistent with training data.

\paragraph{Relation with classical in-distribution generalization studies.}Classical generalization studies the prediction error over the underlying data distribution, and often involves complexity of the hypothesis class and/or training data \citep{vapnik2015uniform,mcallester1999some}. Our focus on \qmet constraints violation is, in fact, not an orthogonal problem, but potentially a core part of in-distribution generalization for this setting. Here, the underlying distribution is supported on all pairs of $\mathcal{X} \times \mathcal{X}$. Indeed, if a learning algorithm has large \dis, it must attain large prediction error on $S \subset \mathcal{X} \times \mathcal{X}$; if it has large \vio, it must violates the \qmet constraints and necessarily admits bad prediction error on some pairs (whose true distances obey the \qmet constraints). \Cref{thm:dis-vio-gen} (proved below) formalizes this idea, where we characterize generalization with the \dis \emph{over all possible pairs in $\mathcal{X} \times \mathcal{X}$}.

\subsection{%
\texorpdfstring{\Cref*{thm:dis-vio-gen}: \nameref*{thm:dis-vio-gen}}{\Dis and \Vio Lower-Bound Generalization}%
}
\begin{repthm}[\Dis and \Vio Lower-Bound Generalization Error]{thm}{thm:dis-vio-gen}{thmDisVioGen}
For non-negative $\hat{d}$, $\dis(\hat{d}) \geq \max(\dis_S(\hat{d}), \sqrt{\smash{\vio(\hat{d})}\vphantom{d^{2^2}}})$, where $\dis(\hat{d})$ captures generalization over the entire $\mathcal{X}$ space.%
 \vspace{-2pt}%
\end{repthm}

\subsubsection{%
Proof%
}
\begin{proof}[Proof of \Cref{thm:dis-vio-gen}]

It is obvious that \begin{equation}
    \dis(\hat{d}) \geq \dis_S(\hat{d}). \label{eq:dis-bound-gen}
\end{equation}

Therefore, it remains to show that $\dis(\hat{d}) \geq \sqrt{\vio(\hat{d})}$.

WLOG, say $\vio(\hat{d}) > 1$. Otherwise, the statement is trivially true.

By the definition of \vio (see \Cref{defn:qmet-vio}), we have, for some $x, y, z \in \mathcal{X}$, with $\hat{d}(x,z) > 0$, \begin{equation}
    \frac{\hat{d}(x,z)}{\hat{d}(x,y) + \hat{d}(y,z)} = \vio(\hat{d}).
\end{equation}

If $\hat{d}(x,y) + \hat{d}(y,z) = 0$, then we must have one of the following two cases: \begin{itemize}
    \item If $d(x,y) > 0$ or $d(y,z)> 0$, the statement is true because $\dis(\hat{d}) = \infty$.
    \item If $d(x,y) = d(y,z)= 0$, then $d(x,z) = 0$ and the statement is true since $\dis(\hat{d}) \geq \frac{\hat{d}(x,z)}{d(x,z)} = \infty$.
\end{itemize}

It is sufficient to prove the case that $\hat{d}(x,y) + \hat{d}(y,z) > 0$. We can derive \begin{align}
    \hat{d}(x,z)
    & = \vio(\hat{d}) \left( \hat{d}(x,y) + \hat{d}(y,z) \right) \\
    & \geq \frac{\vio(\hat{d})}{\dis(\hat{d})} \left(\vphantom{\hat{d}} d(x,y) + d(y,z) \right) \\
    & \geq \frac{\vio(\hat{d})}{\dis(\hat{d})}\ d(x,z). \label{eq:vio-gen-bound-dist-ratio}
\end{align}

If $d(x,z) = 0$, then $\dis(\hat{d}) = \infty$ and the statement is trivially true.

If $d(x,z) > 0$, above \Cref{eq:vio-gen-bound-dist-ratio} implies \begin{equation}
    \dis(\hat{d}) \geq \frac{\hat{d}(x,z)}{d(x,z)} \geq \frac{\vio(\hat{d})}{\dis(\hat{d})} \implies \dis(\hat{d}) \geq \sqrt{\vio(\hat{d})}.  \label{eq:vio-bound-gen}
\end{equation}

Combining \Cref{eq:dis-bound-gen,eq:vio-bound-gen} gives the desired statement.

\end{proof}

\subsection{\texorpdfstring{\Cref*{lemma:orth-equi-ex}: \nameref*{lemma:orth-equi-ex}}{Algorithms Equivariant to Orthogonal Transforms}}
\begin{repthm}[Examples of \orthequi Algorithms]{lemma}{lemma:orth-equi-ex}{lemmaOrthEquiEx}
$k$-nearest-neighbor with Euclidean distance, dot-product kernel ridge regression (including min-norm linear regression and MLP trained with squared loss in NTK regime) are \orthequi.%
 \vspace{-1pt}
\end{repthm}

Recall the definition of Equivariant Learning Transforms.
\begin{repthm}[Equivariant Learning Algorithms]{defn}{defn:equi-alg}{defnEquiAlg}
    Given training set $\mathcal{D} = \{(z_i, y_i)\}_i \subset \mathcal{Z} \times \mathcal{Y}$, where $z_i$ are inputs and $y_i $ are targets, a learning algorithm $\mathsf{Alg}$ produces a function $\mathsf{Alg}(\mathcal{D}) \colon \mathcal{Z} \rightarrow Y$ such that $\mathsf{Alg}(\mathcal{D})(z')$ is the function's prediction on sample $z'$.
    Consider $\mathcal{T}$ a set of transformations $\mathcal{Z} \rightarrow \mathcal{Z}$. $\mathsf{Alg}$ is equivariant to $\mathcal{T}$ iff for all transform $T \in \mathcal{T}$, training set $\mathcal{D}$, $\mathsf{Alg}(\mathcal{D}) = \mathsf{Alg}(T \mathcal{D})\circ T$,
    where $T \mathcal{D} = \{(T z, y) \colon (z, y) \in \mathcal{D}\}$ is the training set with transformed inputs.
\end{repthm}
\subsubsection{Proof}

\begin{proof}[Proof of \Cref{lemma:orth-equi-ex}]

We consider the three algorithms individually: \begin{itemize}
    \item
    \textbf{$k$-nearest neighbor with Euclidean distance.}

    It is evident that if a learning algorithm only depend on pairwise dot products (or distances), it is equivariant to orthogonal transforms, which preserve dot products (and distances). $k$-nearest-neighbor with Euclidean distance  only depends on pairwise distances, which can be written in terms of dot products: \begin{equation}
        \norm{x - y}_2^2 = x\T x + y\T y - 2 x\T y.
    \end{equation}
    Therefore, it is equivariant to orthogonal transforms.

    \item
    \textbf{Dot-product kernel ridge regression.}
    
    Since orthogonal transforms preservers dot-products, dot-product kernel ridge regression is equivariant to them.

    As two specific examples, let's look at linear regression and NTK for fully-connected MLPs.
    
    \begin{itemize}
    \item
    \textbf{Min-norm least-squares linear regression.}

    Recall that the solution to min-norm least-squares linear regression $Ax = b$ is given by Moore–Penrose pseudo-inverse $x = A^+ b$. For any matrix $A \in \R^{m\times n}$ with SVD $U \Sigma V^* = A$, and $T \in O(n)$ (where $O(n)$ is the orthogonal group in dimension $n$), we have \begin{equation}
        (AT\T)^+ = (U \Sigma V^*T\T)^+ = T V\Sigma^+ U^* = T A^+,
    \end{equation}
    where we used $T^* = T\T$ for $T \in O(n)$.
    The solution for the transformed data $AT\T$ and $b$ is thus \begin{equation}
        (AT\T)^+ b = T A^+ b.
    \end{equation}
    Thus, for any new data point $\tilde{x} \in \R^n$ and its transformed version $T \tilde{x} \in \R^n$,\begin{equation}
        \underbrace{(T\tilde{x})\T (AT\T)^+ b}_{\mathclap{\text{transformed problem prediction}}} = \tilde{x}\T T\T T A^+ = \underbrace{\tilde{x} A^+\vphantom{(T\tilde{x})\T (AT\T)^+ b}}_{\mathclap{\text{original problem prediction}}}.
    \end{equation}
    Hence, min-norm least-squares linear regression is equivariant to orthogonal transforms.
    \item
    \textbf{MLP trained with squared loss in NTK regime.}

    We first recall the NTK recursive formula from \citep{jacot2018neural}.

    Denote the NTK for a MLP with $L$ layers with the scalar kernel $\Theta^{(L)} \colon \R^d \times \R^d \rightarrow \R$. Let $\beta > 0$ be the (fixed) parameter for the bias strength in the network model, and $\sigma$ be the activation function.
    Given $x, z \in \R^d$, it can be recursively defined as following. For $h\in[L]$,  \begin{equation}
        \Theta^{(h)}(x, z) \trieq\Theta^{(h-1)}(x, z) \dot{\Sigma}^{(h)}(x, z) + \Sigma^{(h)}(x, z),
    \end{equation}
    where \begin{align}
        \Sigma^{(0)}(x, z) &= \frac{1}{d} x\T z + \beta^2, \\
        \Lambda^{(h-1)}(x, z) & = \begin{pmatrix}
            \Sigma^{(h-1)}(x,x) & \Sigma^{(h-1)}(x,z) \\
            \Sigma^{(h-1)}(z,x) &  \Sigma^{(h-1)}(z,z)
        \end{pmatrix}, \\
        \Sigma^{(h)}(x, z) &= c\cdot \expect[(u, v) \sim \mathcal{N}(0, \Lambda^{(h-1)})]{\sigma(u) \sigma(v)} + \beta^2, \\
        \dot\Sigma^{(h)}(x, z) &= c\cdot \expect[(u, v) \sim \mathcal{N}(0, \Lambda^{(h-1)})]{\dot\sigma(u) \dot\sigma(v)},
    \end{align}
    for some constant $c$.

    It is evident from the recursive formula, that $\Theta^{(h)}(x, z)$ only depends on $x\T x$, $z\T z$ and $x\T z$. Therefore, the NTK is \emph{invariant} to orthogonal transforms.

    Furthermore, training an MLP in NTK regime is the same as kernel regression with the NTK \citep{jacot2018neural}, which has a unique solution only depending on the kernel matrix on training set, denoted as $K_\mathsf{train} \in \R^{n\times n}$, where $n$ is the training set size. Specifically, for training data $\{(x_i, y_i)\}_{i \in [n]}$, the solution $f_\mathsf{NTK}^* \colon \R \rightarrow \R$ can be written as \begin{equation}
        f_\mathsf{NTK}^*(x) = \begin{pmatrix}
            \Theta^{(L)}(x, x_1) & \Theta^{(L)}(x, x_2) & \cdots & \Theta^{(L)}(x, x_n)
        \end{pmatrix} K_\mathsf{train}^{-1} y,
    \end{equation}
    where $y = \begin{pmatrix}
        y_1 & y_2 & \dots & y_n
    \end{pmatrix}$ is the vector of training labels.

    Consider any orthogonal transform $T \in O(d)$, and the NTK regression trained on the transformed data $\{(T x_i, y_i)\}_{i \in [n]}$. Denote the solution as $f_{\mathsf{NTK},T}^* \colon \R \rightarrow \R$. As we have shown, $K_\mathsf{train}^{-1}$ is invariant to such transforms, and remains the same. Therefore, \begin{align}
        f_{\mathsf{NTK},T}^*(T x)
        & = \begin{pmatrix}
            \Theta^{(L)}(T x, T x_1) & \Theta^{(L)}(T x, T x_2) & \cdots & \Theta^{(L)}(T x, T x_n)
        \end{pmatrix} K_\mathsf{train}^{-1} y \\
        & = \begin{pmatrix}
            \Theta^{(L)}(x, x_1) & \Theta^{(L)}(x, x_2) & \cdots & \Theta^{(L)}(x, x_n)
        \end{pmatrix} K_\mathsf{train}^{-1} y \\
        & = f_\mathsf{NTK}^*(x).
    \end{align}

    Hence, MLPs trained (with squared loss) in NTK regime is equivariant to orthogonal transforms.

    Furthermore, we note that there are many variants of MLP NTK formulas depending on details such as the particular initialization scheme and bias settings. However, they usually only lead to slight changes that do not affect our results. For example, while the above recursive NTK formula are derived assuming that the bias terms are initialized with a normal distribution \citep{jacot2018neural}, the formulas for initializing bias as zeros \citep{geifman2020similarity} does not affect the dependency only on dot product, and thus our results still hold true.
    \end{itemize}
\end{itemize}

These cases conclude the proof.
\end{proof}

\subsection{\texorpdfstring{\Cref*{thm:orth-equi-bad}: \nameref*{thm:orth-equi-bad}}{Failure of Algorithms Equivariant to Orthogonal Transforms}}

\begin{repthm}[Failure of \orthequi Algorithms]{thm}{thm:orth-equi-bad}{thmQmetGenOrthEqui}
Let $(f_n)_n$ be an \emph{arbitrary} sequence of large values. There is an infinite sequence of \qmet spaces $((\mathcal{X}_n, d_n))_n$ with $\size{\mathcal{X}_n} = n$, $\mathcal{X}_n \subset \R^n$ such that, over a random training set $S$ of size $m$, any \orthequi algorithm outputs a predictor $\smash{\hat{d}}$ that
\begin{itemize}[topsep=-4pt, itemsep=-4pt]
    \item $\smash{\hat{d}}$ fails \emph{non-negativity}, or
    \item $\smash{\max(\dis_S(\hat{d}),\vio(\hat{d})) \geq f_n}$~~(\ie, $\smash{\hat{d}}$ approximates training $S$ badly or  is far from a \qmet),
\end{itemize}
with probability $1/2 - o(1)$,
as long as $S$ does not contain almost all pairs $\smash{1 - m / n^2 = \omega(n^{-1/3})}$, and does not only include few pairs $\smash{m / n^2 = \omega(n^{-1/2})}$.%
 \vspace{-1pt}
\end{repthm}

Recall that the little-Omega notation means $f = \omega(g) \iff g = o(f)$.

\begin{figure}
    \centering
\ifdefined\appdefn%
\vspace{-14pt}%
\else
\vspace{-29pt}%
\fi
\scalebox{0.89}{
\begin{tikzpicture}[bayes_net, node distance = 0.5cm, every node/.style={inner sep=0}]
    \node[main_node, minimum size=0.82cm] (1y) {$y$};
    \node[main_node, minimum size=0.82cm] (1x) [above left = -0.02cm and 2.4cm of 1y] {$x$};
    \node[main_node, minimum size=0.82cm] (1z) [below left = -0.02cm and 2.4cm of 1y] {$z$};
    \node[main_node, minimum size=0.82cm] (1yp) [above right = -0.02cm and 2.4cm of 1x] {$y'$};
    \node[main_node, minimum size=0.82cm] (1wp) [above right = -0.02cm and 2.4cm  of 1y] {$w'$};
    \node[main_node, minimum size=0.82cm] (1w) [below right = -0.02cm and 2.4cm of 1y] {$w$};
    \node[] (1ineq) [below = 0.8cm of 1y, inner sep=4pt, rounded corners=0.3cm, fill=black, opacity=0.08, text opacity=1, align=center,
    minimum width=7.07cm
    ]
    {\small
    $\displaystyle
        \vio(\hat{d})
        \geq \frac{\hat{d}(x, z)}{\hat{d}(x, y) + \hat{d}(y, z)}
        \geq \frac{c\vphantom{\hat{d}(y, z)}}{\dis_S(\hat{d})(\dis_S(\hat{d})+\hat{d}(y, z))}
        $%
    };

    \node[] (1tr) [below left = 0cm and -7.5cm of 1ineq, inner sep=4pt, rounded corners=0.3cm] {$\displaystyle\begin{aligned}
        \text{Training (\raisebox{2pt}{\protect\tikz{\protect\draw[line width=1.25pt, ->, thick, >=stealth'] (0, 0) -- (0.65, 0);}}\hspace{-1pt}) : }\hphantom{d}
        d(x, z) &= c,\hphantom{1l}
        d(w, z) = 1, \\[-0.6ex]
        d(x, {\color{blue} y}) &= 1,\hphantom{cl}
        d(y, {\color{blue} w'}) = 1. \\[-0.3ex]
        \text{Test (\raisebox{2pt}{\protect\tikz{\protect\draw[line width=1.25pt, ->, thick, >=stealth', dashed] (0, 0) -- (0.65, 0);}}\hspace{-1pt}) : }\hphantom{d} \hat{d}(y, z) & = {\color{red} ?}
    \end{aligned}$};

    \tikzset{myptr/.style={decoration={markings,mark=at position 1 with %
        {\arrow[scale=0.825,>=stealth']{>}}},postaction={decorate}}}
    \path[]
    (1x) edge[-, myptr, line width=1.4pt] [right] node [left=3pt] {$c$} (1z)
    (1w) edge[-, myptr, line width=1.4pt] [right] node [below=3pt] {$1$} (1z)
    (1x) edge[-, myptr, line width=1.4pt] [right] node [midway, above=3pt] {$1$} (1y)
    (1y) edge[-, myptr, line width=1.4pt] [right] node [pos=0.45, above=3pt] {$1$} (1wp)
    (1y) edge[-, myptr, dashed, line width=1.4pt] [right] node [above=3pt] {\color{red} $?$} (1z)
    ;

    \node[main_node, minimum size=0.82cm] (2y) [right = 7.25cm of 1y] {$y$};
    \node[main_node, minimum size=0.82cm] (2x) [above left = -0.02cm and 2.4cm of 2y] {$x$};
    \node[main_node, minimum size=0.82cm] (2z) [below left = -0.02cm and 2.4cm of 2y] {$z$};
    \node[main_node, minimum size=0.82cm] (2yp) [above right = -0.02cm and 2.4cm of 2x] {$y'$};
    \node[main_node, minimum size=0.82cm] (2wp) [above right = -0.02cm and 2.4cm  of 2y] {$w'$};
    \node[main_node, minimum size=0.82cm] (2w) [below right = -0.02cm and 2.4cm of 2y] {$w$};

    \node[] (2ineq) [below = 0.8cm of 2y, inner sep=4pt, rounded corners=0.3cm, fill=black, opacity=0.08, text opacity=1, align=center,
    minimum width=7.07cm
    ]
    {
    \small
    $\displaystyle
     \vio(\hat{d})
        \geq \frac{\hat{d}(y, z)}{\hat{d}(y, w) + \hat{d}(w, z)}
        \geq \frac{\hat{d}(y, z)}{2\cdot\dis_S(\hat{d})}
    $
    };

    \node[] (2tr) [below left = 0cm and -7cm of 2ineq, inner sep=4pt, rounded corners=0.3cm] {$\displaystyle\begin{aligned}
        \text{Training (\raisebox{2pt}{\protect\tikz{\protect\draw[line width=1.25pt, ->, thick, >=stealth'] (0, 0) -- (0.65, 0);}}\hspace{-1pt}) : }\hphantom{d}
        d(x, z) &= c,\hphantom{1l}
        d(w, z) = 1, \\[-0.6ex]
        d(x, {\color{blue} y'}) &= 1,\hphantom{cl}
        d(y, {\color{blue} w}) = 1. \\[-0.3ex]
        \text{Test (\raisebox{2pt}{\protect\tikz{\protect\draw[line width=1.25pt, ->, thick, >=stealth', dashed] (0, 0) -- (0.65, 0);}}\hspace{-1pt}) : }\hphantom{d} \hat{d}(y, z) & = {\color{red} ?}
    \end{aligned}$};

    \path[]
    (2x) edge[-, myptr, line width=1.4pt] [right] node [left=3pt] {$c$} (2z)
    (2w) edge[-, myptr, line width=1.4pt] [right] node [below=3pt] {$1$} (2z)
    (2x) edge[-, myptr, line width=1.4pt] [right] node [midway, above=3pt] {$1$} (2yp)
    (2y) edge[-, myptr, line width=1.4pt] [right] node [pos=0.45, above=3pt] {$1$} (2w)
    (2y) edge[-, myptr, dashed, line width=1.4pt] [right] node [above=3pt] {\color{red} $?$} (2z)
    ;


    \begin{scope}[on background layer]
        \fill[fill=black, opacity=0.08] \convexpath{1z,1x,1y}{0.53cm};
        \fill[fill=black, opacity=0.08] \convexpath{2y,2w,2z}{0.53cm};
    \end{scope}
\end{tikzpicture}
}
\vspace{-22pt}
\caption{
Two training sets
pose incompatible constraints (\raisebox{-2pt}{\protect\tikz{\fill[fill=black, opacity=0.15]  circle(1ex);}})
for the test pair distance $d(y, z)$.
%
With one-hot features, an orthogonal transform can exchange $(*, {\color{blue}y})\leftrightarrow (*, {\color{blue}y'})$ and $(*, {\color{blue}w}) \leftrightarrow (*, {\color{blue}w'})$, leaving the test pair $(y, z)$ unchanged, but transforming the training set from one scenario to the other. Given either set, an \orthequi algorithm must attain same training \dis and predict identically on $(y, z)$. For appropriate $c$, this implies large \dis (not fitting training set) or \vio (not approximately a quasimetric) in one of these cases.
}
\ifdefined\appdefn%
\else
\vspace{-1pt}%
\fi
    \label{fig:app:orth-equi-fail-simple}
\end{figure}

\subsubsection{Proof}

\paragraph{Proof strategy.} In our proof below, we will extend the construction discussed in \Cref{sec:theory-qmet-vio-orth-equi} to large \qmet spaces (reproduced here as \Cref{fig:app:orth-equi-fail-simple}). To do so, we \begin{enumerate}
    \item Construct large \qmet spaces containing many copies of the (potentially failing) structure in \Cref{fig:app:orth-equi-fail-simple}, where we can consider training sets of certain properties such that \begin{itemize}
        \item we can pair up such training sets,
        \item an algorithm equivariant to orthogonal transforms must fail on one of them,
        \item for each pair, the two training sets has equal probability of being sampled;
    \end{itemize}

    Then, it remains to show that with probability $1-o(1)$ we end up with a training set of such properties.
    \item Consider sampling training set as independently collecting each pair with a certain probability $p$, and carefully analyze the conditions to sample a training set with the special properties with high probability $1-o(1)$.
    \item Extend to fixed-size training sets and show that, under similar conditions, we sample a training set with the special properties with high probability $1-o(1)$.
\end{enumerate}

In the discussion below and the proof, we will freely speak of infinite distances between two elements of $\mathcal{X}$, but really mean a very large value (possibly finite). This allows us to make the argument clearer and less verbose. Therefore, we are not restricting the applicable settings of \Cref{thm:orth-equi-bad} to \qmets with (or without) infinite distances.

In \Cref{sec:theory-qmet-vio-orth-equi}, we showed how orthogonal-transform-equivariant algorithms can not predict $\hat{d}(y, z)$ differently for the two particular \qmet spaces and their training sets shown in \Cref{fig:app:orth-equi-fail-simple}.

But are these the only bad training sets? Before the proof, let us consider what kinds of training sets are bad for these two \qmet spaces. Consider the \qmets $d_\lft$ and $d_\rgt$ over $\mathcal{X} \trieq \{x,y,y',z,w,w'\}$, with distances as shown in the left and right parts of  \Cref{fig:app:orth-equi-fail-simple}, where we  assume that the unlabeled pairs have infinite distances except in the \ul{left pattern} $d(x, w') \leq 2$, and in the \ul{both patterns}  $d(y, z)$ has some appropriate value consistent with the respective triangle inequality.

Specifically, we ask: \begin{itemize}
    \item
    For what training sets $S_\lft \subset \mathcal{X} \times \mathcal{X}$ can we interchange $y \leftrightarrow y'$ and $w \leftrightarrow w'$ on 2nd input to obtain a valid training set for $d_\rgt$, regardless of $c$?
    \item
    For what training sets $S_\rgt \subset \mathcal{X} \times \mathcal{X}$ can we interchange $y \leftrightarrow y'$ and $w \leftrightarrow w'$ on 2nd input to obtain a valid training set for $d_\lft$, regardless of $c$?
\end{itemize}

Note that if $S_\lft$ (or $S_\rgt$) satisfies its condition, the predictor $\hat{d}$ from an algorithm equivariant to orthogonal transforms must (1) predict $\hat{d}(y, z)$ identically and (2) attain the same training set \dis on it and its transformed training set. As we will see in the proof for \Cref{thm:orth-equi-bad}, this implies large \dis or \vio for appropriate $c$.

Intuitively, all we need is that the transformed data do not break \qmet constraints. However, its conditions are actually nontrivial as we want to set $c$ to arbitrary: \begin{itemize}

    \item We can't have $(x, w) \in S_\rgt$ because it would be transformed into $(x, w')$ which has $d_\lft(x, w') \leq 2$. Then $d_\rgt(x, w) \leq 2$ and then restricts the possible values of $c$ due to triangle inequality with $d_\rgt(w, z) = 1$. For similar reasons, we can't have $(x, w') \in S_\lft$. In fact, we can't have a path of finite total distance from $x$ to $w$ (or $w'$) in $S_\rgt$ (or $S_\lft$).

    \item We can not have $(y', y') \in S_{(\cdot)}$ (which has distance $0$), which would get transformed into $(y', y)$ with distance $0$, which (on the other pattern) would restrict the possible values of $c$ due to triangle inequality. For similar reasons $(w', w')$, and cycles containing $y'$ or $w'$ with finite total distance, should be avoided.
    
    \item For the theoretical analysis, we assumed that the truth $d$ is a \qmet rather than just being a \qpmet. The difference is that \qpmet additionally allows two distinct elements to have $0$ distance. This assumptions allows us to freely talk about distance ratios for defining \dis and \vio.
    
    For this particular reason, we can't allow $(y, y')$, $(y', y)$, $(w, w')$, $(w', w)$, $(y, y)$ or $(w, w)$, as they break this assumption. However, with metrics more friendly to zero distances (than \dis and \vio, which are based on distance ratios), it might be possible to allow them and obtain better bounds in the second-moment argument below in the proof for \Cref{thm:orth-equi-bad}.
\end{itemize}

With these understandings of the pattern shown in \Cref{fig:app:orth-equi-fail-simple}, we are ready to discuss the constructed \qmet space and training sets.

\begin{proof}[Proof of \Cref{thm:orth-equi-bad}] Our proof follows the outline listed above.
\begin{enumerate}
    \item \textbf{Construct large \qmet spaces containing many copies of the (potentially failing) structure in \Cref{fig:app:orth-equi-fail-simple}.  }

    For any $n > 0$, consider the following \qmet space $(\mathcal{X}_n, d_n)$ of size $n$, with one-hot features. WLOG, assume $n = 12 k$ is a multiple of $12$. If it is not, set at most $11$ elements to have infinite distance with every other node. This won't affect the asymptotics. Let the $n = 12k$ elements of the space be \begin{alignat}{7}
        \mathcal{X}_n =
        &\{
        x^\lft_1, \dots, x^\lft_{k}, &&
        x^\rgt_1, \dots, x^\rgt_{k}, &&
        w^\lft_1, \dots, w^\lft_{k}, &&
        w^\rgt_1, \dots, w^\rgt_{k}, \notag\\
        &\hphantom{\{}
        y^\lft_1, \dots, y^\lft_{k}, &&
        y^\rgt_1, \dots, y^\rgt_{k}, &&
        w'^\lft_1, \dots, w'^\lft_k, &&
        w'^\rgt_{k+1}, \dots, w'^\rgt_{2k}, \notag\\
        &\hphantom{\{}
        y'^\lft_1, \dots, y'^\lft_{k}, &&
        y'^\rgt_{k+1}, \dots y'^\rgt_{2k}, &&
        z_1, \dots, z_k, &&
        z_{k+1}, \dots, z_{2k}\},
    \end{alignat}

    with \qmet distances, $\forall \ci, \cj$, \begin{alignat}{3}
        d_n(x^\lft_\ci, z_\cj) & = d_n(x^\rgt_\ci, z_\cj) &&= c \\
        d_n(w^\lft_\ci, z_\cj) & = d_n(w^\rgt_\ci, z_\cj) &&= 1 \\
        d_n(x^\lft_\ci, y^\lft_\ci) & = d_n(x^\rgt_\ci, y'^\rgt_\ci) &&= 1 \\
        d_n(y^\lft_\ci, w'^\lft_\ci) & = d_n(y^\rgt_\ci, w^\rgt_\ci) &&= 1 \\
        d_n(x^\lft_\ci, w'^\lft_\ci) & = 2 \\
        d_n(y^\lft_\ci, z_\cj) & = c \\
        d_n(y^\rgt_\ci, z_\cj) & = 2,
    \end{alignat}
    where subscripts are colored to better show when they are the same (or different), unlisted distances are infinite (except that $d_n(u, u) = 0, \forall u \in \mathcal{X}$). Essentially, we equally divide the $12k$ nodes into $6$ ``types'', $\{x, y, w, z, w', y'\}$, corresponding to the $6$ nodes from \Cref{fig:app:orth-equi-fail-simple}, where each type has half of its nodes corresponding to the \ul{left pattern} (of \Cref{fig:app:orth-equi-fail-simple}), and the other half corresponding to the \ul{right pattern}, except for the $z$ types.

    Furthermore, \begin{itemize}
        \item Among the \ul{left-pattern} nodes, each set with the same subscript are bundled together in the sense that $x^\lft_\ci$ only has finite distance to $y^\lft_\ci$ which only has finite distance to $w'^\lft_\ci$ (instead of other $y^\lft_j$'s or $w'^\lft_k$'s). However, since distance to/from $y^\lft_\ci$ and $w^\lft_\ci$ are infinite anyways, we can pair \begin{equation}
            (x^\lft_\ci, y^\lft_\ci, w'^\lft_\ci, y'^\lft_\cj, w^\lft_\cl, z_\ch)
        \end{equation} for any $\ci, \cj, \cl, \ch$, to obtain a \ul{left pattern}.

        \item Among the \ul{right-pattern} nodes, each set with the same subscript are bundled together in the sense that $x^\rgt_\ci$ only has finite distance to $y'^\rgt_\ci$, and $y^\rgt_\cj$ which only has finite distance to $w^\rgt_\cj$ (instead of other $y'^\rgt_j$'s or $w^\rgt_k$'s). However, since are distances are infinite anyways, we can pair \begin{equation}
            (x^\rgt_\ci, y'^\rgt_\ci, y^\rgt_\cj, w^\rgt_\cj, w'^\rgt_\cl, z_\ch)
        \end{equation} for any $\ci, \cj, \cl, \ch$, to obtain a \ul{right pattern}.
    \end{itemize}

    We can see that $(\mathcal{X}, d)$ indeed satisfies all \qmet space requirements (\Cref{defn:qmet-space}), including triangle inequalities (\eg, by, for each $(a, b)$ with finite distance $d_n(a, b) < \infty$, enumerating finite-length paths from $a$ to $b$).

    Now consider the sampled training set $S$.  \begin{itemize}
        \item We say $S$ is \emph{bad} on a \ul{left pattern} specified by  $\cil, \cjl, \cll, \chl$, if
        \begin{align}
            S & \supset \{
            (x^\lft_\cil, z_\chl),
            (x^\lft_\cil, y^\lft_\cil),
            (y^\lft_\cil, w'^\lft_\cil),
            (w^\lft_\cll, z_\chl)
            \} \label{eq:left-pattern-require} \\
            \emptyset = S & \cap \{
            (y^\lft_\cil, z_\chl),
            (y^\lft_\cil, y^\lft_\cil),
            (w^\lft_\cll, w^\lft_\cll),
            (y'^\lft_\cjl, y'^\lft_\cjl),
            (w'^\lft_\cil, w'^\lft_\cil), \notag\\
            & \mathbin{\hphantom{\cap \{ }}
            (x^\lft_\cil, w'^\lft_\cil),
            (y^\lft_\cil, y'^\lft_\cjl),
            (w^\lft_\cll, w'^\lft_\cil),
            (y'^\lft_\cjl, y^\lft_\cil),
            (w'^\lft_\cil, w^\lft_\cll)
            \} \label{eq:left-pattern-prohibit}
        \end{align}

        \item We say $S$ is \emph{bad} on a \ul{right pattern} specified by $\cir, \cjr, \clr, \chr$, if
        \begin{align}
            S & \supset \{
            (x^\rgt_\cir, z_\chr),
            (x^\rgt_\cir, y'^\rgt_\cir),
            (y'^\rgt_\cjr, w^\rgt_\cjr),
            (w^\rgt_\cjr, z_\chr)
            \} \label{eq:right-pattern-require} \\
           \emptyset =  S & \cap \{
            (y^\rgt_\cjr, z_\chr),
            (y^\rgt_\cjr, y^\rgt_\cjr),
            (w^\rgt_\cjr, w^\rgt_\cjr),
            (y'^\rgt_\cir, y'^\rgt_\cir),
            (w'^\rgt_\clr, w'^\rgt_\clr), \notag\\
            & \mathbin{\hphantom{\cap \{ }}
            (x^\rgt_\cir, w'^\rgt_\cjr),
            (y^\rgt_\cjr, y'^\rgt_\cir),
            (w^\rgt_\cjr, w'^\rgt_\clr),
            (y'^\rgt_\cir, y^\rgt_\cjr),
            (w'^\rgt_\clr, w^\rgt_\cjr)
            \} \label{eq:right-pattern-prohibit}
        \end{align}
    \end{itemize}

    Most importantly, \begin{itemize}
        \item If $S$ is bad on a \ul{left pattern} specified by  $\cil, \cjl, \cll, \chl$, consider the orthogonal transform that interchanges $y^\lft_\cil \leftrightarrow y'^\lft_\cjl$ and $w^\lft_\cll \leftrightarrow w'^\lft_\cil$ on 2nd input. In $S$, the possible transformed pairs are \begin{alignat}{7}
            d(x^\lft_\cil, y^\lft_\cil) & = 1
            && \hphantom{-}\longrightarrow\hphantom{-} &
            d(x^\lft_\cil, y'^\lft_\cjl) & = 1,
            \tag{known in $S$}\\
            d(y^\lft_\cil, w'^\lft_\cil) & = 1
            && \hphantom{-}\longrightarrow\hphantom{-} &
            d(y^\lft_\cil, w^\lft_\cll) & = 1
            ,\tag{known in $S$} \\
            d(u, y^\lft_\cil) & = \infty
            && \hphantom{-}\longrightarrow\hphantom{-} &
            d(u, y'^\lft_\cjl) & = \infty
            , \tag{poissble in $S$ for some $u\neq x^\lft_\cil$} \\
            d(u, y'^\lft_\cjl) & = \infty
            && \hphantom{-}\longrightarrow\hphantom{-} &
            d(u, y^\lft_\cil) & = \infty
            , \tag{poissble in $S$ for some $u$} \\
            d(u, w'^\lft_\cil) & = \infty
            && \hphantom{-}\longrightarrow\hphantom{-} &
            d(u, w^\lft_\cll) & = \infty
            , \tag{poissble in $S$ for some $u \notin \{x^\lft_\cil, y^\lft_\cil\}$} \\
            d(u, w^\lft_\cll) & = \infty
            && \hphantom{-}\longrightarrow\hphantom{-} &
            d(u, w'^\lft_\cil) & = \infty
            . \tag{poissble in $S$ for some $u$}
        \end{alignat}

        The crucial observation is that the transformed training set just look like one sampled from a \qmet space where \begin{itemize}
            \item the \qmet space has one less set of \ul{left-pattern} elements,
            \item the \qmet space has one more set of \ul{right-pattern} elements, and
            \item transformed training set is \emph{bad} on that extra \ul{right pattern} (given by the extra set of \ul{right-pattern} elements),
        \end{itemize}
        which can be easily verified by comparing the transformed training set with the requirements in \Cref{eq:right-pattern-require,eq:right-pattern-prohibit}.

        \item Similarly, if $S$ is bad on a \ul{right pattern} specified by $\cir, \cjr, \clr, \chr$, consider the orthogonal transform that interchanges $y^\rgt_\cjr \leftrightarrow y'^\rgt_\cir$ and $w^\rgt_\cjr \leftrightarrow w'^\rgt_\clr$ on 2nd input. In $S$ the possible transformed pairs are \begin{alignat}{7}
            d(x^\rgt_\cir, y'^\rgt_\cir) & = 1
            && \hphantom{-}\longrightarrow\hphantom{-} &
            d(x^\rgt_\cir, y^\rgt_\cjr) & = 1,
            \tag{known in $S$} \\
            d(y^\rgt_\cjr, w^\rgt_\cjr) & = 1
            && \hphantom{-}\longrightarrow\hphantom{-} &
            d(y^\rgt_\cjr, w'^\rgt_\clr) & = 1
            ,\tag{known in $S$} \\
            d(u, y^\rgt_\cjr) & = \infty
            && \hphantom{-}\longrightarrow\hphantom{-} &
            d(u, y'^\rgt_\cir) & = \infty
            , \tag{poissble in $S$ for some $u$} \\
            d(u, y'^\rgt_\cir) & = \infty
            && \hphantom{-}\longrightarrow\hphantom{-} &
            d(u, y^\rgt_\cjr) & = \infty
            , \tag{poissble in $S$ for some $u\neq x^\rgt_\cir$} \\
            d(u, w'^\rgt_\clr) & = \infty
            && \hphantom{-}\longrightarrow\hphantom{-} &
            d(u, w^\rgt_\cjr) & = \infty
            , \tag{poissble in $S$ for some $u$} \\
            d(u, w^\rgt_\cjr) & = \infty
            && \hphantom{-}\longrightarrow\hphantom{-} &
            d(u, w'^\rgt_\clr) & = \infty
            . \tag{poissble in $S$ for some $u \notin \{x^\rgt_\cir, y^\rgt_\cjr\}$}
        \end{alignat}

        Again, the crucial observation is that the transformed training set just look like one sampled from a \qmet space where \begin{itemize}
            \item the \qmet space has one less set of \ul{right-pattern} elements,
            \item the \qmet space has one more set of \ul{left-pattern} elements, and
            \item transformed training set is \emph{bad} on that extra \ul{left pattern} (given by the extra set of \ul{left-pattern} elements),
        \end{itemize}
        which can be easily verified by comparing the transformed training set with the requirements in \Cref{eq:left-pattern-require,eq:left-pattern-prohibit}.
    \end{itemize}

    Therefore, when $S$ is bad on \emph{both a \ul{left pattern} and a \ul{right pattern}} (necessarily on disjoint sets of pairs), we consider the following orthogonal transform composed of: \begin{enumerate}
        \item both transforms specified above (which only transforms 2nd inputs),

        (so that after this we obtain \emph{another possible training set of same size from the \qmet space that is only different up to some permutation of $\mathcal{X}$})

        \item a permutation of $\mathcal{X}$ (on both inputs) so that the bad \ul{left-pattern} nodes and the bad \ul{right-pattern} nodes exchange features,
    \end{enumerate}
    This transforms gives \emph{another possible training set of same size from the \ul{same} \qmet space, also is bad on a \ul{left pattern} and a \ul{right pattern}}. Moreover, with a particular way of select bad patterns (\eg, by the order of the subscripts), this process is \emph{reversible}. Therefore, we have defined a way to pair up all such bad training sets.

    Consider the predictors $\hat{d}_\mathsf{before}$ and $\hat{d}_\mathsf{after}$ trained on these two training sets (before and after transform) with an learning algorithm equivariant to orthogonal transforms.  Assuming that they satisfy non-negativity and \idtindis, we have, \begin{itemize}
        \item The predictors have the same \dis over respective training sets.

        Therefore we denote this \dis as $\dis_S(\hat{d})$ without specifying the predictor $\hat{d}$ or training set $S$.

        \item the predictors must predict the same on heldout pairs in the sense that \begin{align}
            \hat{d}_\mathsf{before}(y_\cil^\lft, z_\chl) & =  \hat{d}_\mathsf{after}(y_\cjr^\rgt, z_\chr) \\
            \hat{d}_\mathsf{before}(y_\cjr^\rgt, z_\chr) & =  \hat{d}_\mathsf{after}(y_\cil^\lft, z_\chl).
        \end{align}

        Focusing on the first, we denote \begin{equation}
            \hat{d}(y, z) \trieq \hat{d}_\mathsf{before}(y_\cil^\lft, z_\chl)  =  \hat{d}_\mathsf{after}(y_\cjr^\rgt, z_\chr)
        \end{equation}  without specifying the predictor $\hat{d}$ or the specific $y$ and $z$.
    \end{itemize}


    However, the \qmet constraints on heldout pairs $(y_\cil^\lft, z_\chl)$ and $(y_\cjr^\rgt, z_\chr)$ are completely different (see the left vs.~right part of \Cref{fig:app:orth-equi-fail-simple}). Therefore, as shown in \Cref{fig:app:orth-equi-fail-simple}, assuming \emph{non-negativity}, \emph{one of the two predictors} must have total \vio at least \begin{equation}
        \vio(\hat{d})
            \geq \max\left(
            \frac{c}{\dis_S(\hat{d})(\dis_S(\hat{d})+\hat{d}(y, z))}_{\textstyle,}\hphantom{1}
            \frac{\hat{d}(y, z)}{2\cdot\dis_S(\hat{d})}
            \right). \label{eq:equi-orth-total-voi-geq-max-estd}
    \end{equation}

    Fixing a large enough $c$, two terms in the $\max$ of \Cref{eq:equi-orth-total-voi-geq-max-estd} can equal for some  $\hat{d}(y, z)$, and are respectively decreasing and increasing in $\hat{d}(y, z)$. In that case, we have \begin{equation}
        \vio(\hat{d}) \geq \frac{\delta}{2\cdot\dis_S(\hat{d})},
    \end{equation}for $\delta > 0$ such that \begin{equation}
        \frac{c}{\dis_S(\hat{d})(\dis_S(\hat{d})+\delta)} = \frac{\delta}{2\cdot\dis_S(\hat{d})}.
    \end{equation}
    Solving the above quadratic equation gives \begin{equation}
        \delta = \frac{-\dis_S(\hat{d})+ \sqrt{\dis_S(\hat{d})^2 + 8c} }{2},
    \end{equation}
    leading to  \begin{equation}
        \vio(\hat{d}) \geq \frac{-1+ \sqrt{1 + 8c / \dis_S(\hat{d})^2} }{4}.
    \end{equation}

    Therefore, choosing $c \geq f_n^2(4f_n + 1)^2$ gives \begin{align}
        & \hphantom{{}\implies{}} \dis_S(\hat{d}) \leq f_n \\
        & \implies \vio(\hat{d})
        \geq \frac{-1+ \sqrt{1 + 8c / \dis_S(\hat{d})^2} }{4} \\
        & \hphantom{{}\implies \vio(\hat{d})}
        \geq \frac{-1+ \sqrt{1 + 8 f_n^2(4f_n + 1)^2 / f_n^2} }{4} \\
        & \hphantom{{}\implies \vio(\hat{d})}
        = \frac{-1+ \sqrt{1 + 8 (4f_n + 1)^2} }{4} \\
        & \hphantom{{}\implies \vio(\hat{d})}
        \geq \frac{-1+ 4f_n + 1}{4} \\
        & \hphantom{{}\implies \vio(\hat{d})}
        = f_n.
    \end{align}


    Hence, for training sets that are \emph{bad} on \emph{both a \ul{left pattern} and a \ul{right pattern}}, we have shown a way to pair them up such that \begin{itemize}
        \item each pair of training sets have the same size, and
        \item the algorithm fail on one of each pair by producing a distance predictor that \begin{itemize}
            \item has either \dis over training set $\geq f_n$, or \vio $\geq f_n$, and
            \item has test MSE $\geq f_n$.
        \end{itemize}
    \end{itemize}

    \begin{rmk} \label{rmk:suffice-to-show-S-bad-left-right-whp}
        Note that all training sets of size $m$ has equal probability of being sampled. Therefore, to prove the theorem, it suffices to show that with probability $1 - o(1)$, we can sample a training set of size $m$ that is \emph{bad} on \emph{both a \ul{left pattern} and a \ul{right pattern}}.
    \end{rmk}

    \item \textbf{Consider sampling training set as individually collecting each pair with a certain probability $p$, and carefully analyze the conditions to sample a training set with the special properties with high probability $1-o(1)$.}

    In probabilistic methods, it is often much easier to work with independent random variables. Therefore, instead of considering uniform sampling a training set $S$ of fixed size $m$, we consider including each pair in $S$ with probability $p$, chosen independently. We will first show result based on this sampling procedure via a second moment argument, and later extend to the case with a fixed-size training set.

    First, let's define some notations that ignore constants:\begin{align}
        f \sim g &\iff f = (1 + o(1)) g \\
        f \ll g &\iff f = o(g).
    \end{align}

    We start with stating a standard result from the second moment method \citep{alon2004probabilistic}.
    \begin{cor}[Corollary~4.3.5 of \citep{alon2004probabilistic}]\label{cor:second-moment-sym}
        Consider random variable $X = X_1 + X_2 + \dots + X_n$, where $X_i$ is the indicator random variable for event $A_i$. Write $i \sim j$ if $i \neq j$ and the pair of events $(A_i, A_j)$ are not independent. Suppose the following quantity does not depend on $i$: \begin{equation}
            \Delta^* \trieq \sum_{j \sim i} \prob{A_j \given A_i}.\label{eq:second-moment-delta-star}
        \end{equation}
        If $\expect{X} \rightarrow \infty$ and $\Delta^* \ll \expect{X}$, then $X  \sim \expect{X}$ with probability $1 - o(1)$.
    \end{cor}

    We will apply this corollary to obtain conditions on $p$ such that $S$ with probability $1 - o(1)$ is \emph{bad} on some \ul{left pattern}, and conditions such that $S$ with probability $1 - o(1)$ is \emph{bad} on some \ul{right pattern}. A union bound would then give the desired result.

    \begin{itemize}
        \item \textbf{$S$ is \emph{bad} on some \ul{left pattern}.}

        Recall that a \ul{left pattern} is specified by $\cil, \cjl, \cll, \chl$ all $\in [k]$: \begin{equation}
            (x^\lft_\cil, y^\lft_\cil, w'^\lft_\cil, y'^\lft_\cjl, w^\lft_\cll, z_\chl)
        \end{equation}

        Therefore, we consider $k^4 = (\frac{n}{12})^4$ events of the form \begin{equation}
            A_{\cil, \cjl, \cll, \chl} \trieq \{\text{$S$ is bad on the \ul{left pattern} at $\cil, \cjl, \cll, \chl$}\}.
        \end{equation}

        Obviously, these events are symmetrical, and the $\Delta^*$ in \Cref{eq:second-moment-delta-star} does not depend on $i$.

        By the \qmet space construction and the requirement for $S$ to be bad on a \ul{left pattern} in \Cref{eq:left-pattern-require,eq:left-pattern-prohibit}, we can see that $(\cil, \cjl, \cll, \chl) \sim (\cipl, \cjpl, \clpl, \chpl)$ only if $\cil = \cipl$ or $\cjl=\cjpl$ or $\cll=\clpl$ or $\chl=\chpl$.

        Therefore, we have \begin{align}
            \expect{X} & \sim n^4 p^4 (1-p)^{10} \tag{include $4$ pairs \& exclude $10$ pairs}\\
            \Delta^*
            & \ll
            n^3 p^4 (1-p)^9 \tag{share $\cjl$} \\
            & \hphantom{{}\ll{}} +
            n^3 p^2 (1-p)^7 \tag{share $\cil$} \\
            & \hphantom{{}\ll{}} +
            n^3 p^4 (1-p)^9 \tag{share $\cll$} \\
            & \hphantom{{}\ll{}} +
            n^3 p^4 (1-p)^{10} \tag{share $\chl$} \\
            & \hphantom{{}\ll{}} +
            n^2 p^2 (1-p)^{4} \tag{share $\cjl,\cil$} \\
            & \hphantom{{}\ll{}} +
            n^2 p^4 (1-p)^{8} \tag{share $\cjl,\cll$} \\
            & \hphantom{{}\ll{}} +
            n^2 p^4 (1-p)^{9} \tag{share $\cjl,\chl$} \\
            & \hphantom{{}\ll{}} +
            n^2 p^2 (1-p)^{4} \tag{share $\cil,\cll$} \\
            & \hphantom{{}\ll{}} +
            n^2 p (1-p)^{6} \tag{share $\cil,\chl$} \\
            & \hphantom{{}\ll{}} +
            n^2 p^3 (1-p)^{9} \tag{share $\cll,\chl$} \\
            & \hphantom{{}\ll{}} +
            n (1-p)^{3} \tag{share $\cil,\cll,\chl$} \\
            & \hphantom{{}\ll{}} +
            n p^3 (1-p)^{8} \tag{share $\cjl,\cll,\chl$} \\
            & \hphantom{{}\ll{}} +
            n p (1-p)^{3} \tag{share $\cjl,\cil,\chl$} \\
            & \hphantom{{}\ll{}} +
            n p^2 (1-p) \tag{share $\cjl,\cil,\cll$} \\
            & \sim
            n^3 p^2(1-p)^7 + n^2(p^2(1-p)^4 + p(1-p)^6)  \\
            & \hphantom{{}\ll{}} +
            n((1-p)^3+p^2(1-p)).
        \end{align}

        Therefore, to apply \Cref{cor:second-moment-sym}, we need to have \begin{align}
            n^4p^4(1-p)^{10} & \rightarrow \infty \\
            n^3 p^2(1-p)^7 & \ll n^4p^4(1-p)^{10} \\
            n^2(p^2(1-p)^4 + p(1-p)^6) & \ll n^4p^4(1-p)^{10} \\
            n((1-p)^3+p^2(1-p)) & \ll n^4p^4(1-p)^{10},
        \end{align}
        which gives \begin{align}
            p & \gg n^{-1/2} \\
            1 - p & \gg n^{-1/3}
        \end{align}
        as a sufficient condition to for $S$ to be bad on some \ul{left pattern} with probability $1 - o(1)$.

        \item \textbf{$S$ is \emph{bad} on some \ul{right pattern}.}

        Recall that a \ul{right pattern} is specified by $\cir, \cjr, \clr, \chr$ all $\in [k]$: \begin{equation}
            (x^\rgt_\cir, y'^\rgt_\cir, y^\rgt_\cjr, w^\rgt_\cjr, w'^\rgt_\clr, z_\chr)
        \end{equation}

        Similarly, we consider $k^4 = (\frac{n}{12})^4$ events of the form \begin{equation}
            A_{\cir, \cjr, \clr, \chr} \trieq \{\text{$S$ is bad on the \ul{left pattern} at $\cir, \cjr, \clr, \chr$}\}.
        \end{equation}

        Again, these events are symmetrical, and $\Delta^*$ in \Cref{eq:second-moment-delta-star} does not depend on $i$.

        Similarly, we have \begin{align}
            \expect{X} & \sim n^4 p^4 (1-p)^{10} \tag{include $4$ pairs \& exclude $10$ pairs}\\
            \Delta^*
            & \ll
            n^3 p^3 (1-p)^9 \tag{share $\cir$} \\
            & \hphantom{{}\ll{}} +
            n^3 p^3 (1-p)^8 \tag{share $\cjr$} \\
            & \hphantom{{}\ll{}} +
            n^3 p^4 (1-p)^{10} \tag{share $\chr$} \\
            & \hphantom{{}\ll{}} +
            n^3 p^4 (1-p)^{9} \tag{share $\clr$} \\
            & \hphantom{{}\ll{}} +
            n^2 p^2 (1-p)^{4} \tag{share $\cir,\cjr$} \\
            & \hphantom{{}\ll{}} +
            n^2 p^2 (1-p)^{9} \tag{share $\cir,\chr$} \\
            & \hphantom{{}\ll{}} +
            n^2 p^3 (1-p)^{8} \tag{share $\cir,\clr$} \\
            & \hphantom{{}\ll{}} +
            n^2 p^2 (1-p)^{7} \tag{share $\cjr,\chr$} \\
            & \hphantom{{}\ll{}} +
            n^2 p^3 (1-p)^{5} \tag{share $\cjr,\clr$} \\
            & \hphantom{{}\ll{}} +
            n^2 p^4 (1-p)^{9} \tag{share $\chr,\clr$} \\
            & \hphantom{{}\ll{}} +
            n p^2 (1-p)^{4} \tag{share $\cjr,\chr,\clr$} \\
            & \hphantom{{}\ll{}} +
            n p^2 (1-p)^{8} \tag{share $\cir,\chr,\clr$} \\
            & \hphantom{{}\ll{}} +
            n p^2 (1-p) \tag{share $\cir,\cjr,\clr$} \\
            & \hphantom{{}\ll{}} +
            n (1-p) \tag{share $\cir,\cjr,\chr$} \\
            & \sim
            n^3 p^3(1-p)^8 + n^2p^2(1-p)^4  \\
            & \hphantom{{}\ll{}} +
            n(1-p).
        \end{align}

        Therefore, to apply \Cref{cor:second-moment-sym}, we need to have \begin{align}
            n^4p^4(1-p)^{10} & \rightarrow \infty \\
            n^3 p^3(1-p)^8 & \ll n^4p^4(1-p)^{10} \\
            n^2 p^2(1-p)^4 & \ll n^4p^4(1-p)^{10} \\
            n(1-p) & \ll n^4p^4(1-p)^{10},
        \end{align}
        which gives \begin{align}
            p & \gg n^{-3/4} \\
            1 - p & \gg n^{-1/3}
        \end{align}
        as a sufficient condition to for $S$ to be bad on some \ul{right pattern} with probability $1 - o(1)$.
    \end{itemize}

    So, by union bound, as long as \begin{align}
        p & \gg n^{-1/2} \label{eq:good-p}\\
        1 - p & \gg n^{-1/3}, \label{eq:good-1mp}
    \end{align}
    $S$ is bad on some \ul{left pattern} \emph{and} some \ul{right pattern} with probability $1 - o(1)$.

    \item \textbf{Extend to fixed-size training sets and show that, under similar conditions, we sample a training set with the special properties with high probability $1-o(1)$.}

    To extend to fixed-size training sets, we consider the following alteration procedure: \begin{enumerate}
        \item Sample training set $S$ by independently include each pair with probability $p \trieq \frac{m+ \delta}{n^2}$, for some $\delta > 0$.
        \item Show that with high probability $1 - o(1)$, we end up with $[m, m + 2\delta]$ pairs in $S$.
        \item Make sure that $p$ satisfy \Cref{eq:good-p} and \Cref{eq:good-1mp} so that $S$ is bad  on some \ul{left pattern} \emph{and} some \ul{right pattern} with high probability $1-o(1)$.
        \item Randomly discard the additional pairs, and show that with high probability $1-o(1)$ this won't affect that $S$ is bad  on some \ul{left pattern} \emph{and} some \ul{right pattern}.
    \end{enumerate}

    We now consider each step in details: \begin{enumerate}
        \item \textbf{Sample training set $S$ by independently include each pair with probability $p \trieq \frac{m+ \delta}{n^2}$, for some $\delta > 0$.}

        For $p \trieq \frac{m+ \delta}{n^2}$, the number of pairs in the training set is distributed as \begin{equation}
            \mathrm{Binomial}(n^2, \frac{m+\delta}{n^2}).
        \end{equation}

        \item \textbf{Show that with high probability $1 - o(1)$, we end up with $[m, m + 2\delta]$ pairs in $S$.}

        Standard Binomial concentration tells us that, \begin{equation}
            \delta \gg n \sqrt{p (1-p)} \implies \prob{\mathrm{Binomial}(n^2, \frac{m+\delta}{n^2}) \notin [m, m + 2\delta]} \rightarrow 0,
        \end{equation}
        which can be satisfied if
        \begin{equation}
            \delta \gg n.
        \end{equation}

        \item \textbf{Make sure that $p$ satisfy \Cref{eq:good-p} and \Cref{eq:good-1mp} so that $S$ is bad  on some \ul{left pattern} \emph{and} some \ul{right pattern} with high probability $1-o(1)$.}

        Therefore, we want \begin{align}
            \frac{m+\delta}{n^2} & \gg n^{-1/2} \\
            1 - \frac{m+\delta}{n^2} & \gg n^{-1/3}.
        \end{align}

        \item \textbf{Randomly discard the additional pairs, and show that with high probability $1-o(1)$ this won't affect that $S$ is bad on some \ul{left pattern} \emph{and} some \ul{right pattern}.}


        Consider any specific bad \ul{left pattern} \emph{and a} \ul{right pattern} in $S$. It is sufficient that we don't break these two patterns during discarding.

        Since we only discard pairs, it suffices to only consider the pairs we want to preserve, which are a total of $8$ pairs across two patterns.

        Each such pair is discarded the probability $\leq \frac{2 \delta}{m}$, since we remove at most $2\delta$ pairs. By union bound, \begin{equation}
            \prob{\text{all $8$ pairs are preserved}} \geq 1 - \frac{16 \delta}{m}.
        \end{equation}

        Hence, it suffices to make sure that \begin{equation}
            \delta \ll m.
        \end{equation}
    \end{enumerate}

    Collecting all requirements, we have \begin{align}
            \delta & \gg n \\
            \frac{m+\delta}{n^2} & \gg n^{-1/2} \\
            1 - \frac{m+\delta}{n^2} & \gg n^{-1/3} \\
            \delta &\ll m.
    \end{align}

    Assume that \begin{align}
        \frac{m}{n^2} & \gg n^{-1/2} \\
        1 - \frac{m}{n^2} & \gg n^{-1/3}.
    \end{align}It can be easily verified that using $\delta \trieq n^{1.1}$ satisfies all conditions.

    Hence, for a uniformly randomly sampled training set $S$ with size $m$,  $S$ is bad  on some \ul{left pattern} \emph{and} some \ul{right pattern} with high probability $1-o(1)$, as long as \begin{align}
        \frac{m}{n^2} & \gg n^{-1/2} \\
        1 - \frac{m}{n^2} & \gg n^{-1/3}.
    \end{align}

    This is exactly the condition we need to prove the theorem (see \Cref{rmk:suffice-to-show-S-bad-left-right-whp}).
\end{enumerate}
This concludes the proof.
\end{proof}

\subsubsection{Discussions}
\paragraph{Training set size dependency. } Intuitively, when the training set has almost all pairs, \vio can be lowered by simply fitting training set well; when it is small and sparse, the learning algorithm may have an easier job finding some consistent \qmet. \Cref{thm:orth-equi-bad} shows that, outside these two cases, algorithms equivariant to orthogonal transforms can fail. Note that for the latter case, \Cref{thm:orth-equi-bad} requires the training fraction to decrease slower than $n^{-1/2}$, which rules out training sizes that is linear in $n$. We leave improving this result as future work. Nonetheless,  \Cref{thm:orth-equi-bad} still covers common scenarios such as a fixed fraction of all pairs, and highlights that a training-data-agnostic result (such as the ones for PQEs) is not possible for these algorithms.

\paragraph{Proof techniques.}
In embedding theory, it is quite standard to analyze \qmets as directed graphs due to their lack of nice metric structure.
In the proof for \Cref{thm:orth-equi-bad}, we used abundant techniques from the probabilistic method, which are commonly used for analyzing graph properties in the asymptotic case,
including \Cref{cor:second-moment-sym} from the second moment technique, and the alteration technique to extend to fixed-size training sets.
While such techniques may be new in learning theory, they are standard  for characterizing asymptotic probabilities on graphs, which \qmets are often analyzed as \citep{charikar2006directed,memoli2018quasimetric}.

To provide more intuition on why these techniques are useful here, we note that the construction of a training set of pairs is essentially like constructing an \ER random graph on $n^2$ vertices. \ER (undirected) random graphs come in two kinds: \begin{itemize}
    \item Uniformly sampling a fixed number of $m$ edges;
    \item Adding an edge between each pair with probability $p$, decided independently.
\end{itemize}
The latter, due to its independent decisions, is often much easy to analyze and preferred by many. The alteration technique (that we used in the proof) is also a standard way to transfer a result on a random graph of the latter type, to a random graph of the former type \citep{bollobas2001random}. Readers can refer to \citep{alon2004probabilistic,bollobas2001random,erdos59a} for more in-depth treatment of these topics.

\paragraph{Generalization to other transforms.}
The core of this construction only relies on the ability to swap (concatenated) inputs between $(x,y) \leftrightarrow (x,y')$ and between $(y,w)\leftrightarrow(y,w')$ via a transform. For instance, here the orthogonal transforms satisfy this requirement on one-hot features. Therefore, the result can also be generalized to other transforms and features with the same property. Our stated theorem focuses on orthogonal transforms because they correspond to several common learning algorithms (see \Cref{lemma:orth-equi-ex}). If a learning algorithm is equivariant to some other transform family, it would be meaningful to generalize this result to that transform family, and obtain a similar negative result. We leave such extensions as future work.

\subsubsection{Corollary of \Dis and \Vio for \Uncon MLPs}
\begin{cor}[\Dis and \Vio of \Uncon MLPs]\label{cor:mlp-vio-bad}
    Let $(f_n)_n$ be an arbitrary sequence of desired \vio values. There is an infinite collection of \qmet spaces $((\mathcal{X}_n, d_n))_{n=1,2,\dots}$ with $\size{\mathcal{X}_n} = n$, $\mathcal{X}_n \subset \R^n$ such that MLP trained with squared loss in NTK regime converges to a function $\hat{d}$ that either \begin{itemize}[topsep=-4pt, itemsep=-4pt]
        \item fails non-negativity, or
        \item $\vio(\hat{d}) \geq f_n$,
    \end{itemize}
    with probability $1/2 - o(1)$ over the random training set $S$ of size $m$, as long as $S$ does not contain almost all pairs $1 - m / n^2 = \omega(n^{-1/3})$, and does not only include few pairs $m / n^2 = \omega(n^{-1/2})$.
\end{cor}

\begin{proof}[Proof of \Cref{cor:mlp-vio-bad}]
    This follows directly from \Cref{thm:orth-equi-bad} and standard NTK convergence results obtained from the kernel regression optimality and the positive-definiteness of the NTK. In particular, Proposition~2 of \citep{jacot2018neural} claims that the NTK is positive-definite when restricted to a hypersphere. Since the construction in proof of \Cref{thm:orth-equi-bad} uses one-hot features, the input (concatenation of two features) lie on the hypersphere with radius $\sqrt{2}$. Hence, the NTK is guaranteed positive definite.
\end{proof}

\subsubsection{Empirical Verification of the Failure Construction}

\begin{figure}
    \begin{subfigure}[t]{0.48\linewidth}
        \centering
        \includegraphics[trim=5 8 0 0, clip, scale=0.5]{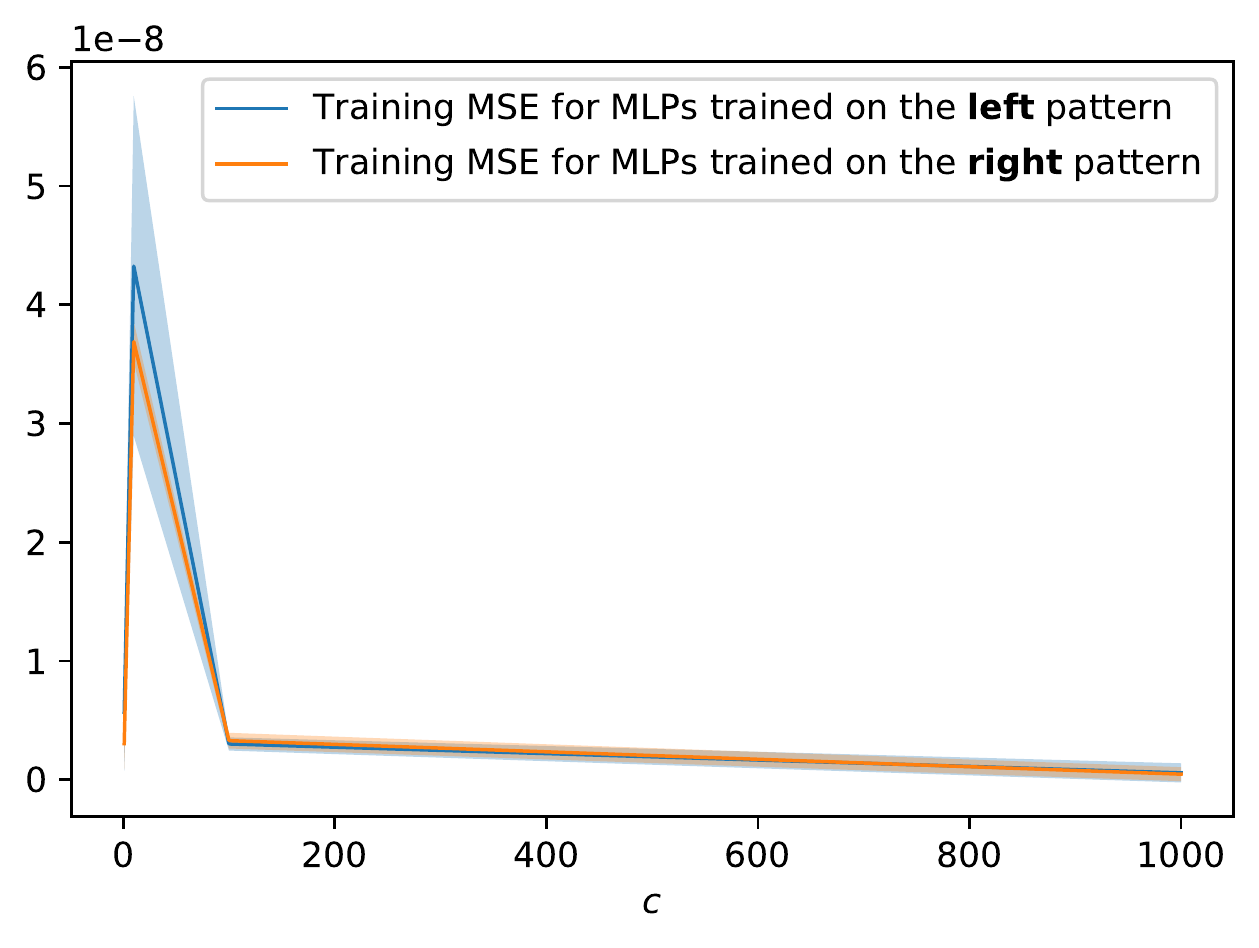}
        \caption{Training losses for varying $c$. Note the scale of the vertical axis.}
    \end{subfigure}\hfill
    \begin{subfigure}[t]{0.48\linewidth}
        \centering
        \includegraphics[trim=0 8 0 0, clip, scale=0.5]{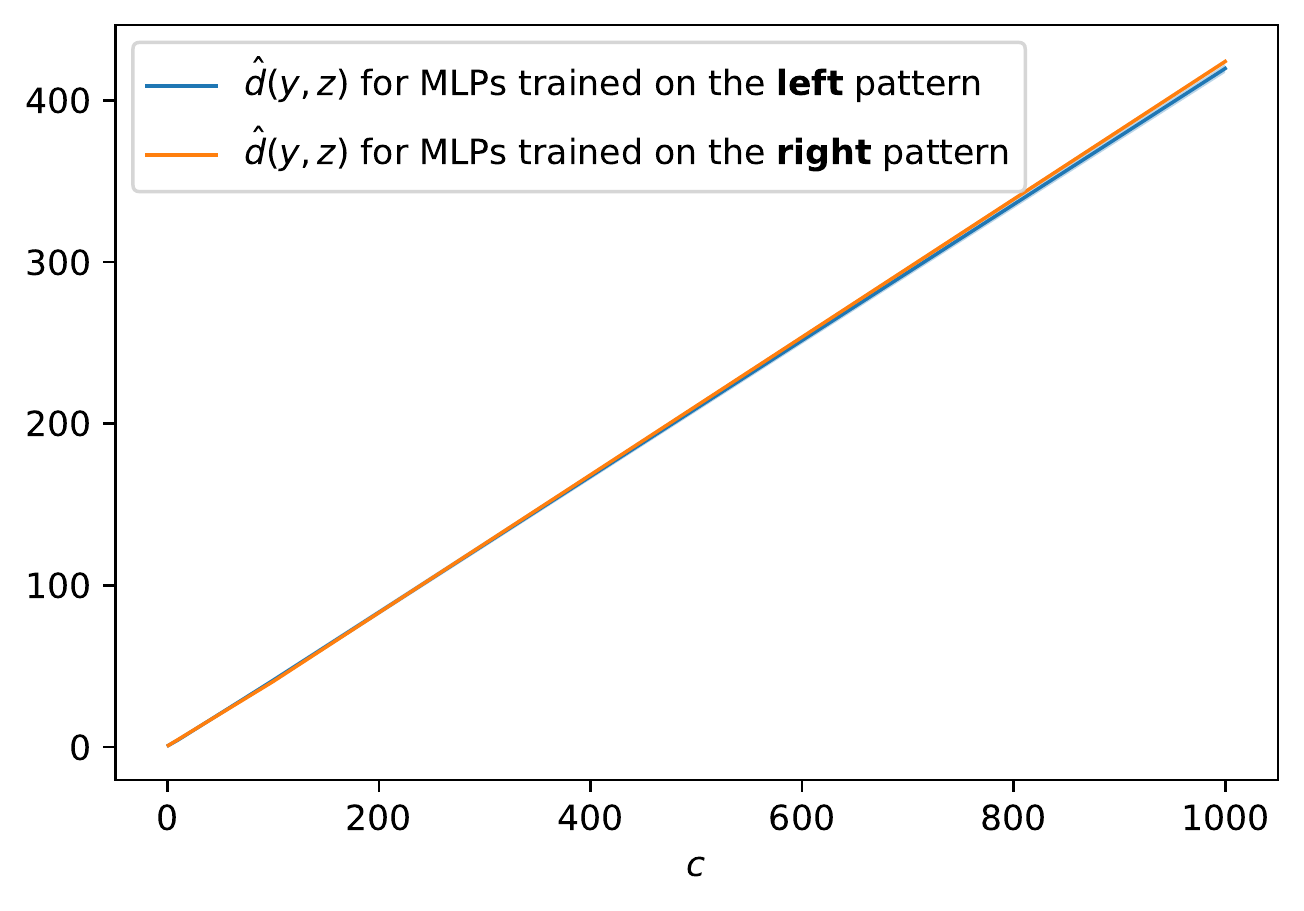}
        \caption{Prediction on heldout pair $\hat{d}(y, z)$ for varying $c$.}
    \end{subfigure}
    \caption{Training \uncon MLPs on the toy failure construction discussed in \Cref{sec:theory-qmet-vio-orth-equi} (reproduced as \Cref{fig:app:orth-equi-fail-simple}). Two patterns in the construction have different constraints on distance of the heldout pair $(y, z)$. Plots show mean and standard deviations over $5$ runs. \textbf{Left:} All training conclude with small training error. \textbf{Right:} Trained MLPs predict identically for both patterns. Here standard deviation is small compared to mean and thus not very visible.
    }    \label{fig:toy-failure-verify}
\end{figure}

We train \uncon MLPs on the toy failure construction discussed in \Cref{sec:theory-qmet-vio-orth-equi} (reproduced as \Cref{fig:app:orth-equi-fail-simple}). The MLP uses 12-1024-1 architecture with ReLU activations, takes in the concatenated one-hot features, and directly outputs predicted distances. Varying $c \in \{1, 10, 100, 1000\}$, we train the above MLP $5$ times on each of the two patterns in \Cref{fig:app:orth-equi-fail-simple}, by regressing towards the training distances via MSE loss.

In \Cref{fig:toy-failure-verify}, we can see that all training runs conclude with small training error, and indeed the trained MLPs predict very similarly on the heldout pair, regardless whether it is trained on the left or right pattern of \Cref{fig:app:orth-equi-fail-simple}, which restricts the heldout pair distance differently.

This verifies our theory (\Cref{thm:orth-equi-bad} and \Cref{cor:mlp-vio-bad}) that algorithms equivariant to orthogonal transforms (including MLPs in NTK regime) cannot distinguish these two cases and thus must fail on one of them.

\section{Proofs and Discussions for \texorpdfstring{\Cref*{sec:emb}: \nameref*{sec:emb}}{PQE Section}}

\subsection{Non-differentiability of Continuous-Valued Stochastic Processes}

In this section we formalize the argument presented in \Cref{sec:continuous-nondiff} to show why continuous-valued stochastic processes lead to non-differentiability. \Cref{fig:continuous-nondiff-intuition} also provides a graphical illustration of the general idea.

\begin{restatable}[\Qmet Embeddings with Continuous-Valued Stochastic Processes are not Differentiable]{prop}{propQmetEmbContDistnNonDiff}
\label{prop:qmet-emb-cont-distn-nondiff}
Consider any $\R^k$-valued stochastic process
$\{R(u)\}_{u \in \R^d}$ such that $u \neq u' \implies \prob{R(u) = R(u')} < c$ for some universal constant $c < 1$. Then $\prob{R(u) \leq R(u')}$ is not differentiable at any $u=u'$.
\end{restatable}

\begin{figure}
    \begin{subfigure}[t]{0.48\linewidth}
        \centering
        \includegraphics[scale=0.243, page=1, trim=90 65 70 120, clip]{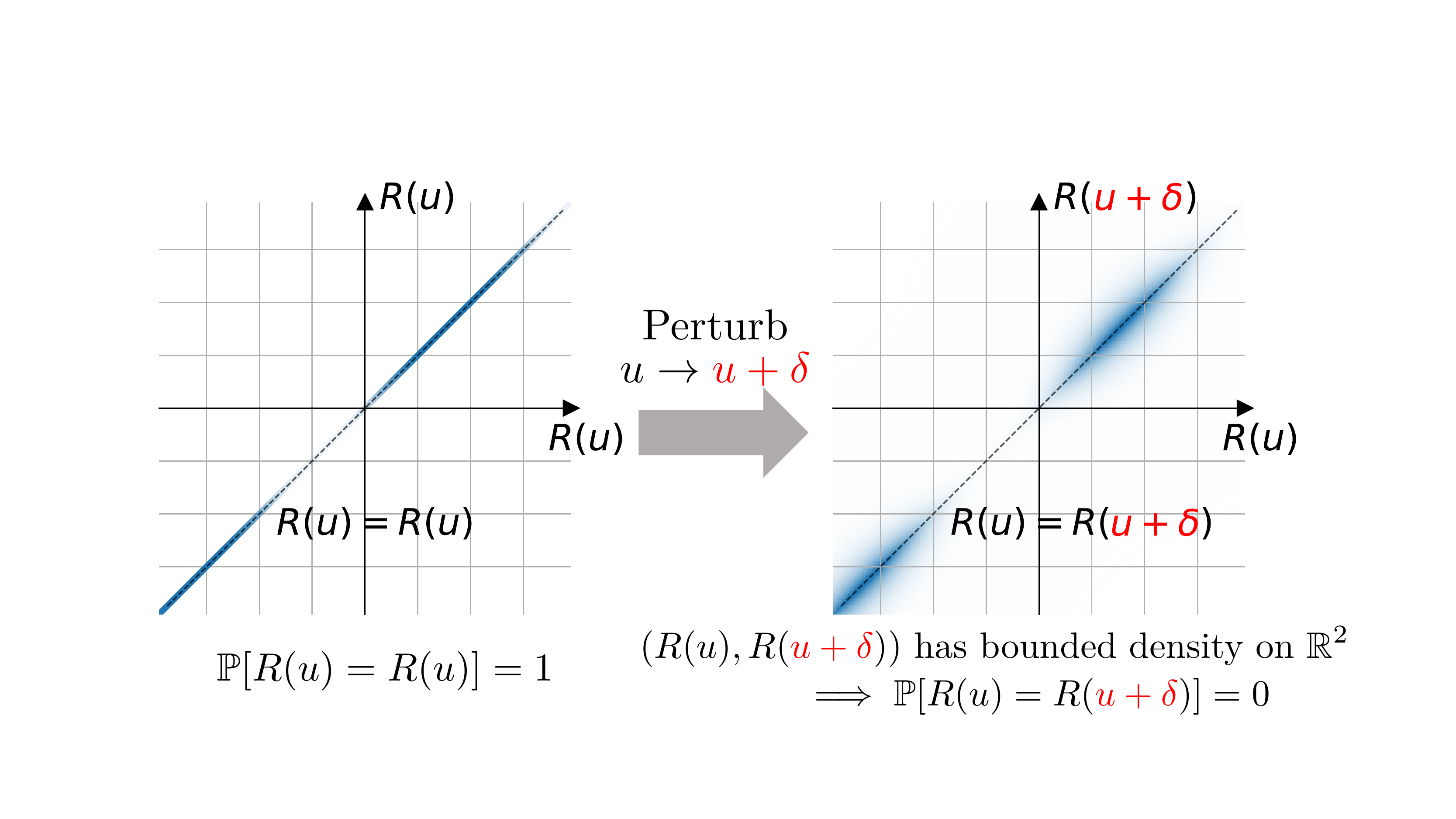}\vspace{-4pt}
        \caption{Continuous-valued stochastic process.}
    \end{subfigure}\hfill
    \begin{subfigure}[t]{0.5\linewidth}
        \centering
        \includegraphics[scale=0.243, page=2, trim=90 65 70 120, clip]{figures/continuous_failure/figures.pdf}\vspace{-4pt}
        \caption{Discrete-valued stochastic process.}
    \end{subfigure}
    \caption{Bivariate distributions from different stochastic processes.
    \textbf{Left:} In a continuous-valued process (where $(N_\theta, N_{\theta'})$ has bounded density if $\theta \neq \theta'$), perturbing one $\theta \rightarrow {\color{red} \theta +\eps}$ leaves $\prob{N_{\theta} = N_{\color{red}\theta + \epsilon}} = 0$. Then one of $\mathbb{P}\big[ N_{\theta} \leq N_{\color{red}\theta + \eps} \big]$ and $\mathbb{P}\big[ N_{\color{red}\theta + \eps} \leq N_{\theta} \big]$ must be far away from $1$ (as they sum to $1$), breaking differentiability at either $\prob{N_{\theta} \leq N_{\theta}} = 1$ or $\prob{ N_{\color{red}\theta + \epsilon} \leq N_{\color{red}\theta + \epsilon}} = 1$. \textbf{Right:} For discrete-valued processes, most probability can still be left on $N_{\theta} = N_{\color{red}\theta + \epsilon}$ and thus do not break differentiability.
    }
    \label{fig:continuous-nondiff-intuition}
\end{figure}
\begin{proof}[Proof of \Cref{prop:qmet-emb-cont-distn-nondiff}]
    Assume that the quantity is differentiable. Then it must be continuous in $u$ and $v$.

    We will use the $(\eps, \delta)$-definition of continuity.

    At any $u \in \R^d$, consider small $\eps \in (0, \frac{1-c}{3})$. By continuity, since \begin{equation}
        \prob{R(u) \leq R(u)}=\prob{ R({\color{red}u + \delta}) \leq R({\color{red}u + \delta})} = 1
    \end{equation} we can find $\eps \in \R^d$ such that \begin{align}
        \prob{R(u) \leq R({\color{red}u + \delta})} &\geq 1 - \eps \\
        \prob{R({\color{red}u + \delta}) \leq R(u) } &\geq 1 - \eps.
    \end{align}

    However, by assumption, $\prob{R(u) = R({\color{red}u + \delta})} < c$. Therefore, \begin{align}
        \prob{R(u) \leq R({\color{red}u + \delta})} & \geq 1 - \eps \\
        \prob{R({\color{red}u + \delta}) < R(u) } & \geq 1 - \eps - c,
    \end{align}
    which implies
    \begin{equation}
        1 = \prob{R(u) \leq R({\color{red}u + \delta})} + \prob{R({\color{red}u + \delta}) < R(u) } \geq 2 - 2\eps - c \geq \frac{5}{3} - \frac{2}{3} c >  1.
    \end{equation}

    By contradiction, the quantity must not be differentiable at any $u = u'$.
\end{proof}


\subsection{\texorpdfstring{\PQEGG}{PQEGG}: Gaussian-based Measure and Gaussian Shapes}\label{sec:gaussian-gaussian}

In \Cref{sec:pqelh}, we presented the following \PQELH formulation for Lebesgue measures and half-lines: \begin{align}
    d_z^\PQELH(u, v) \trieq \sum_i \alpha_i \cdot \Big(1 - \exp \big(-\sum_j(u_{i,j} - v_{i,j})^+\big)\Big). \tag{\ref{eq:pqe-lebesgue-halfline}}
\end{align}
Here, $u_{i,j}$ and $v_{i,j}$ receive zero gradient when $u_{i,j} \leq v_{i,j}$.

\paragraph{Gaussian shapes parametrization.}
We therefore consider a set parametrization where no one set is entirely contained in a different set--- the regions regions $\subset \R^2$ between an axis and a 1D Gaussian density function of fixed variance $\sigma^2_\mathsf{shape}=1$. That is, for each given $u \in R$, we consider sets \begin{equation}
    A_\mathcal{N}(\mu) \trieq \{(a, b) \colon b \in [0, f_\mathcal{N}(a; \mu, 1)]\},
\end{equation}where $f_\mathcal{N}(b; \mu, \sigma^2)$ denotes the density of 1D Gaussian $\mathcal{N}(\mu, \sigma^2)$ with mean $\mu$ and variance $\sigma^2$ evaluated at $b$. Since the Gaussian density function have unbounded support, these sets, which are translated versions of each other, never have one set fully contained in another. For latent $u \in \R^{h \times k}$ reshaped as 2D, our set parametrizations are,\begin{equation}
    u \rightarrow A_{i,j}(u) \trieq A_\mathcal{N}(u_{i,j}), \mathrlap{\qquad\quad i \in [h], j \in [k].}
\end{equation}


\paragraph{A Gaussian-based measure.}
These subsets of $\R^2$ always have Lebesgue measure $1$, which would make PQE symmetrical (if used with a (scaled) Lebesgue measure). Thus, we use an alternative $\R^2$ measure given by the product of a $\R$ Lebesgue measure on the $b$-dimension (\ie, dimension of the function value of the Gaussian density) and a $\R$ Gaussian measure on the $a$-dimension (\ie, dimension on the input of the Gaussian density) centered at $0$ with \emph{learnable} variances $(\sigma_\mathsf{measure}^2)_{i,j}$. To avoid being constrained by the bounded total measure of $1$, we also optimize learnable positive scales $c_{i,j} > 0$. Hence, the each Poisson process has a mean measure as the product of a $\R$ Lebesgue measure and a $\R$ Gaussian with learnable standard deviation, then scaled with a learnable scale.

Note that the Gaussian measure should not be confused with the Gaussian shape. Their parameters also are fully independent with one another.

\paragraph{Computing measures of Gaussian shapes and their intersections.} The intersection of two such Gaussian shapes is formed by two Gaussian tail shapes, reflected around the middle point of the two Gaussian means (since they have the same standard deviation $\sigma^\mathsf{shape}=1$). Hence, it is sufficient to describe how to integrate a Gaussian density on a Gaussian measure over an interval. Applying this with different intervals would give the measure of the intersection, and the measures of the two Gaussian shapes. Omit indices $i,j$ for clarity. Formally, we integrate the Gaussian density $f_\mathcal{N}(a; u, \sigma^2_\mathsf{shape})$ over the centered Gaussian measure with variance  $\sigma_\mathsf{measure}^2$, which has density $f_\mathcal{N}(a; 0, \sigma_\mathsf{measure}^2)$: \begin{equation}
    \int c\cdot f_\mathcal{N}(a; u, \sigma^2_\mathsf{shape}) f_\mathcal{N}(a; 0, \sigma_\mathsf{measure}^2) \diff a,
\end{equation}
which is also another Gaussian integral (\eg, considered as integrating the product measure along the a line of the form $y = x + u$). After standard algebraic manipulations (omitted here), we obtain \begin{align}
    & \hphantom{{}={}} \int c\cdot f_\mathcal{N}(a; u, \sigma^2_\mathsf{shape}) f_\mathcal{N}(a; 0, \sigma_\mathsf{measure}^2) \diff a \\
    & = \frac{c\cdot \exp \left(- u^2/ \sigma^2_\mathsf{total} \right)}{\sqrt{2 \pi \sigma^2_\mathsf{total}}}  \int f_\mathcal{N}\left(a; u \frac{\sigma_\mathsf{measure}^2}{\sigma^2_\mathsf{total}}, \frac{\sigma^2_\mathsf{shape} \sigma_\mathsf{measure}^2}{\sigma^2_\mathsf{total} }\right) \diff a,
\end{align}
for \begin{equation}
    \sigma^2_\mathsf{total} \trieq \sigma^2_\mathsf{shape} + \sigma_\mathsf{measure}^2.
\end{equation}
This can be easily evaluated using statistical computing packages that supports computing the error function and/or Gaussian CDF. Moreover, this final form is also readily differentiable with standard gradient formulas. To summarize, \begin{itemize}
    \item each set $A(u)$ has total measure \begin{equation}
        \frac{c}{\sqrt{2 \pi \sigma^2_\mathsf{total}}} \exp \left(-u^2/ \sigma^2_\mathsf{total} \right);
    \end{equation}
    \item the intersection of $A({v})$ and $A({u_2})$, for $v \leq u_2$ has measure \begin{align}
        & \hphantom{{}+{}} \frac{c\cdot \exp \left(-u_2^2/ \sigma^2_\mathsf{total} \right)}{\sqrt{2 \pi \sigma^2_\mathsf{total}}}  \int_{-\infty}^{\frac{v + u_2}{2}} f_\mathcal{N}\left(a; u_2 \frac{\sigma_\mathsf{measure}^2}{\sigma^2_\mathsf{total}}, \frac{\sigma^2_\mathsf{shape} \sigma_\mathsf{measure}^2}{\sigma^2_\mathsf{total} }\right) \diff a \\
        & + \frac{c\cdot \exp \left(-v^2/ \sigma^2_\mathsf{total} \right)}{\sqrt{2 \pi \sigma^2_\mathsf{total}}}  \int^{+\infty}_{\frac{v + u_2}{2}} f_\mathcal{N}\left(a; v \frac{\sigma_\mathsf{measure}^2}{\sigma^2_\mathsf{total}}, \frac{\sigma^2_\mathsf{shape} \sigma_\mathsf{measure}^2}{\sigma^2_\mathsf{total} }\right) \diff a.
    \end{align}
\end{itemize}

\paragraph{Interpretation and representing any total order.} Consider two Gaussian shapes $A({v})$ and $A({u_2})$. Note that the Gaussian-based measure $\mu_\mathsf{Gaussian}$ is symmetric around and centered at $0$. Therefore, \begin{align}
    \abs{v} < \abs{u_2}
    & \implies
    \mu_\mathsf{Gaussian}(A(v)) > \mu_\mathsf{Gaussian}(A(u_2)) \\
    & \implies  \mu_\mathsf{Gaussian}(A(v) \setminus A(u_2)) >  \mu_\mathsf{Gaussian}(A(u_2) \setminus A(v)).
\end{align}
Moreover, scaling the rates of a Poisson makes it more concentrated (as a Poisson's mean grows as the square of its standard deviation) so that $\lim_{c\rightarrow \infty} \prob{\pois(c\mu_1) \leq \pois(c \mu_2)} = \indic{\mu_1 < \mu_2}$ for $\mu_1 \neq \mu_2$. Then any total order can be represented as the limit of a Poisson process with Gaussian shapes, with the shapes' having their means arranged according to the total order, as the scale on the Gaussian-based measure grows to infinity.



\subsection{Theoretical Guarantees for PQEs}

\begin{repthm}[\Dis and \vio of PQEs]{thm}{thm:pqe-lhgg-low-dis-vio}{thmPQELHGGLowDisVio}
Under the assumptions of \Cref{sec:theory}, \emph{any} \qmet space with size $n$ and treewidth $t$ admits a \PQELH and a \PQEGG with \dis $\smash{\bigO(t \log^2 n)}$ and \vio $1$, with an expressive encoder (\eg, a ReLU network with $\geq3$ layers and polynomial width).
\end{repthm}

In \Cref{sec:pqe-guarantees}, we presented the above theoretical \dis and \vio guarantees for \PQELH and \PQEGG. Furthermore, we commented that the same guarantees apply to more generally to PQEs satisfying a mild condition. Here, we first precisely describe this condition, show that \PQELH and \PQEGG do satisfy it, state and prove the general result, and then show the above as a straightforward corollary.

\subsubsection{The Concentration Property}\label{sec:require-qpart-distn-concentrate}

Recall that PQEs are generally defined with measures $\mu$ and set parametrizations $A$ as \begin{equation}
    d_z^{\mathsf{PQE}}(u, v;\mu,A,\alpha)
    \trieq \sum_i \alpha_i \cdot \expect[\pi_z \sim \Pi_z^{\mathsf{PQE}}(\mu_i, A_i)]{\pi_z(u, v)}, \tag{\ref{eq:pqe}}
\end{equation}
where
\begin{equation}
\mathbb{E}_{\pi_z \sim \Pi_z^{\mathsf{PQE}}(\mu, A)}[\pi_z(u, v)]\trieq 1 - \prod_j \prob{N_j(A_j(u)) \leq N_j(A_j(v))}. \tag{\ref{eq:pqe-qpart-expect}}
\end{equation}

Because the measures $\mu$ and set parametrizations $A$ themselves may have parameters (\eg, as in \PQEGG), we consider them as classes of PQEs. \Eg, \PQEGG is a class of PQEs such that the $\mu$ is the specific Gaussian-based form, and $A$ is the specific Guassian-shape.

\begin{defn}[Concetration Property of PQEs]\label{defn:pqe-concentration}
Consider a PQE class with $h$ mixtures of \qpart distributions, each from $k$ Poisson processes.
We say that it has concentration property if it satisfies the following.
Consider any finite subset of $\mathcal{X}' \subset \mathcal{X}$, and arbitrary function $g \colon \mathcal{X} \rightarrow \R^{h\times k}$. There exists a sequence of $((f^{(n)}, \mu^{(n)}, A^{(n)})_n$ such that \begin{itemize}
    \item $f^{(n)} \colon \mathcal{X}' \rightarrow \R^d$,
    \item $\mu^{(n)}, A^{(n)}$ are valid members of this PQE,
    \item $ \expect[\pi_z \sim \Pi_z^{\mathsf{PQE}}(\mu_i, A_i)]{\pi_z(f^{(n)}(x'), f^{(n)}(y'))}$ uniformly converges to $1 - \prod_{j} \indic{g(x)_{i,j} \leq g(y)_{i,j}}$, over all mixtures $i$ and pairs $x, y \in \mathcal{X}'$.
\end{itemize}
\end{defn}

\paragraph{A sufficient condition.}

It suffices to make the probabilities\begin{equation}
    (x, y, i, j) \rightarrow \prob{N_j(A_j(u)) \leq N_j(A_j(v))}, \label{eq:xyij-poisson-race-probs}
\end{equation}
along some PQE sequence
uniformly converge to the indicators \begin{equation}
    (x, y, i, j) \rightarrow  \indic{g(x')_{i,j} \leq g(y')_{i,j}}. \label{eq:xyij-indics}
\end{equation} This is sufficient since product of bounded functions is uniformly convergent, if each function is.
Both statements below together form \textbf{a sufficient condition} for \Cref{eq:xyij-poisson-race-probs} to uniformly converge to \Cref{eq:xyij-indics}: \begin{enumerate}
    \item \label{enum:pqe-cond} For any $g$, there exists a specific PQE of this class satisfying \begin{itemize}
        \item Measures (of set differences) are consistent with $g$ with some margin $\eps > 0$: $\forall i \in [h], j \in [k], x \in \mathcal{X}', y \in \mathcal{X}'$, \begin{align*}
            g(x)_{i, j} < g(y)_{i, j} &\iff \mu_{i,j}(A_{i,j}({f(x)}) \setminus A_{i,j}({f(y)})) + \eps < \mu_{i,j}(A_{i,j}({f(y)}) \setminus A_{i,j}({f(x)})) \\
            g(x)_{i, j} = g(y)_{i, j} &\iff \mu_{i,j}(A_{i,j}({f(x)}) \setminus A_{i,j}({f(y)})) = \mu_{i,j}(A_{i,j}({f(y)}) \setminus A_{i,j}({f(x)})) = 0.
        \end{align*}
        \item Either of the following: \begin{itemize}
            \item One side must be zero: $\forall i \in [h], j \in [k], x \in \mathcal{X}, y \in \mathcal{X}$,
            \begin{equation}
                \left( \mu_{i,j}(A_{i,j}({f(x)}) \setminus A_{i,j}({f(y)}))\right) \left(\mu_{i,j}(A_{i,j}({f(y)}) \setminus A_{i,j}({f(x)}))\right) = 0,
            \end{equation}
            \item Max measure is bounded by some constant $c > 0$: \begin{equation}
                \max_{x,y,i,j} \mu_{i,j}(A_{i,j}({f(x)}) \setminus A_{i,j}({f(y)})) \leq c.
            \end{equation}
        \end{itemize}
    \end{itemize}
    \item For any given specific PQE of this class, for any positive scale $d > 0$, there is another PQE (with same formulation) whose measures (of set differences) equal exactly those of the given PQE scaled by $d$.
\end{enumerate}
We now show that this is a sufficient condition. Note that a Poisson distribution has standard deviation equal to square root of its mean. This means that as we scale the rate of a Poisson, it becomes more concentrated. Applying to Poisson race probability, we have, for $0 \leq \mu_1 + \eps < \mu_2$,
\begin{itemize}
     \item one direction of Poisson race probability: \begin{align}
        & \hphantom{{}\geq{}} \prob{\pois(d \cdot \mu_1) \leq \pois(d \cdot \mu_2)} \\
        & \geq \prob{\abs{\pois(d \cdot \mu_2) - \pois(d \cdot \mu_1) - d (\mu_2 - \mu_1)} \leq d (\mu_2 - \mu_1)} \\
        & \geq 1 - \frac{\mu_1 + \mu_2}{d (\mu_2 - \mu_1)^2} \\
        & \geq \begin{cases}
            1 - \frac{2}{d\eps} & \text{if $\mu_1 = 0$} \\
            1  - \frac{2c}{d \eps^2} & \text{if $\mu_2 < c$};
        \end{cases}
    \end{align}

     \item the other direction of Poisson race probability: \begin{align}
        & \hphantom{{}\leq{}} \prob{\pois(d \cdot \mu_2) \leq \pois(d \cdot \mu_1)} \\
        & \leq \prob{\abs{\pois(d \cdot \mu_2) - \pois(d \cdot \mu_1) - d (\mu_2 - \mu_1)} \geq d (\mu_2 - \mu_1)} \\
        & \leq \frac{\mu_1 + \mu_2}{d (\mu_2 - \mu_1)^2} \\
        & \leq \begin{cases}
            \frac{2}{d\eps} & \text{if $\mu_1 = 0$} \\
            \frac{2c}{d \eps^2} & \text{if $\mu_2 < c$}.
        \end{cases}
    \end{align}
\end{itemize}
Therefore, applying to scaled versions of the PQE from \Cref{enum:pqe-cond} above, we have thus obtained the desired sequence, where \Cref{eq:xyij-poisson-race-probs} uniformly converges to \Cref{eq:xyij-indics} with rate $\bigO(1/d)$.

\begin{lemma}\label{lemma:pqelhgg-concentrate}
\PQELH and \PQEGG both have the concentration property.
\end{lemma}
\begin{proof}[Proof of \Cref{lemma:pqelhgg-concentrate}]
We show that both classes satisfy the above sufficient condition.
\begin{itemize}
    \item \PQELH: Lebesgue measure $\lambda$ and half-lines.

    WLOG, since $\mathcal{X}$ is countable, we assume that $g$ satisfies \begin{equation}
        g(x)_{i,j} \neq g(y)_{i,j} \implies \abs{g(x)_{i,j} - g(y)_{i,j}} > 1, \qquad\quad \forall i \in [h], j \in [k], x \in \mathcal{X}', y \in \mathcal{X}'.
    \end{equation} The encoder in \Cref{enum:pqe-cond} above $f \colon \mathcal{X} \rightarrow \R^{h\times k}$ can simply be $g$. We then have \begin{equation}
        \mu_{i,j}(A_{i,j}({f(y)}) \setminus A_{i,j}({f(x)}))
        =
        \mathrm{Leb} ((-\infty, g(y)] \setminus (-\infty, g(x)]) \\
        = (g(y)_{i,j} - g(x)_{i,j})^+.
    \end{equation} This ensures that one side is always zero. Furthermore, scaling can be done by simply scaling the encoder $f$. Hence, \PQELH satisfies this constraint.

    \item \PQEGG: Gaussian-based measure and Gaussian shapes (see \Cref{sec:gaussian-gaussian}).

        Because $\mathcal{X}'$ is finit, we can have positive constant margin for the PQE requirements in \Cref{enum:pqe-cond}. (Infinite $\mathcal{X}'$ does not work because the total measure is finite (for a specific \PQEGG with specific values of the scaling).) Concretely, we satisfy both requirements via \begin{itemize}
        \item in descending order of $g(\cdot)_{i,j}$ we assign Gaussian shapes increasingly further from the origin;
        \item scaling comes from that we allow scaling the Gaussian-based measure.
    \end{itemize} Hence, \PQEGG satisfies this constraint for finite $\mathcal{X}$.

\end{itemize}
\end{proof}

\subsubsection{A General Statement}

We now state the general theorem for PQEs with the above concentration property.
\begin{thm}[\Dis and \vio of PQEs (General)]\label{thm:pqe-general-low-dis-vio}
Consider any PQE class with the concentration property.
Under the assumptions of \Cref{sec:theory}, \emph{any} \qmet space with size $n$ and treewidth $t$ admits such a PQE with \dis $\smash{\bigO(t \log^2 n)}$ and \vio $1$, with an expressive encoder (\eg, a ReLU network with $\geq3$ hidden layers, $\bigO(n)$ hidden width, and  $\bigO(n^2)$ \qpart distributions, each with $\bigO(n)$ Poisson processes.).
\end{thm}

Before proving this more general theorem, let us extend a result from \citet{memoli2018quasimetric}.

\begin{restatable}[\Qmet Embeddings with Low \Dis; Adapted from Corollary~2 in \citet{memoli2018quasimetric}]{lemma}{lemmaGeneralConvCombinationQpartsLowDis}
\label{lemma:general-conv-combination-qparts-low-dis}
Let $M = (X, d)$ be a \qpmet space with \tw $t$, and $n = \size{X}$. Then $M$ admits an embedding into a convex combination (\ie, scaled mixture) of $\bigO(n^2)$ \qparts with \dis $\bigO(t \log^2 n)$.
\end{restatable}

\begin{proof}[Proof of \Cref{lemma:general-conv-combination-qparts-low-dis}]

The \dis bound is proved in Corollary~2 in \citep{memoli2018quasimetric}, which states that any \qpmet space with $n$ elements and $t$ \tw admits an embedding into a convex combination of \qparts with \dis $\bigO(t \log^2 n)$.

To see that $n^2$ \qparts suffice, we scrutinize their construction of \qparts in Algorithm~2 of \citep{memoli2018quasimetric}, reproduced below as \Cref{alg:random-qpart-tw-graph}.

\begin{algorithm}[H]
    \caption{Random \qpart of a bounded treewidth graph. Algorithm~2 of \citep{memoli2018quasimetric}.}
    \label{alg:random-qpart-tw-graph}
    \small
    \textbf{Input:} A digraph $G$ of \tw $t$, a hierarchical tree of separators of $G$ $(H, f)$ with width $t$, and $r > 0$.\\
    \textbf{Output:} A random $r$-bounded \qpart $R$.
    \begin{algorithmic}
        \State \textbf{Initialization:} Set $G^* = G$, $H^* = H$ and $R = E(G)$. Perform the following recursive algorithm on $G^*$ and $H^*$.
        \State \textbf{Step 1.} Pick $z \in [0, r/2]$ uniformly at random.
        \State \textbf{Step 2.} If $\size{V(G^*)}\leq 1$,terminate the current recursive call. Otherwise pick the set of vertices $K=G^*$. Let $H_1, \dots, H_m$ be the sub-trees of $H^*$ below $\mathsf{root}(H^*)$  that are hierarchical trees of separators of $C_1,\dots, C_m$ respectively.
        \State \textbf{Step 3.} For all $(u, v)\in E(G^*)$ remove $(u,v)$ from $R$ if one of the following holds: \begin{enumerate}[label=(\alph*),leftmargin=3.8em]
            \item $d_G(u,x)>z$ and $d_G(v, x) \leq z$ for some vertex $x \in K$.
            \item $d_G(x,v)>z$ and $d_G(x, u) \leq z$ for some vertex $x \in K$.
        \end{enumerate}
        \State \textbf{Step 4.} For all $i \in \{1, \dots, m\}$ perform a recursive call of Steps 2-4 setting $G^*=G^*[C_i]$ and $H^*=H_i$.
        \State \textbf{Step 5.} Once all branches of the recursive terminate, enforce transitivity on $R$: For all $u,v,w\in V(G)$ if $(u, v) \in R$ and $(v, w) \in R$, add $(u,w)$ to $R$.
    \end{algorithmic}
\end{algorithm}

Many concepts used in \Cref{alg:random-qpart-tw-graph} are not relevant for our purpose (\eg, $r$-bounded \qpart). Importantly, we observe that for a given \qmet space, the produced \qpart is entirely determined by the random choice of $z$ in Step~1, which is only used to compare with distance values between node pairs. Note that there are $n^2$ node pairs, whose minimum distance is exactly $0$ (\ie, distance from a node to itself). Since $z \geq 0$, there are at most $n^2$ choices of $z$ that lead to at most $n^2$ different \qparts, for all possible values of $r$.

The construction used to prove Corollary~2 of \citep{memoli2018quasimetric} uses exactly \qparts given by this algorithm. Therefore, the lemma is proved.
\end{proof}

\Cref{lemma:general-conv-combination-qparts-low-dis} essentially proves the first half of \Cref{thm:pqe-general-low-dis-vio}. Before proving the full \Cref{thm:pqe-general-low-dis-vio}, we restate the following result from \citep{hiraguchi1951dimension}, which gives us a bound on how many total orders are needed to represent a general partial order (\ie, \qpart).

\begingroup  
\let\repthmXRRR\undefined
\begin{repthm}[Hiraguchi's Theorem \citep{hiraguchi1951dimension,bogart1973maximal}]{thm}{thm:hiraguchi-po-dim}{thmHiraguchiPODim}
Let $(X, P)$ be a partially ordered set such that $\size{X} \geq 4$. Then there exists a mapping $f \colon X \rightarrow \R^{\floor{\size{X} / 2}}$ such that \begin{equation}
    \forall x, y \in X, \qquad x P y \iff f(x) \leq f(y)~\coorwise. \hphantom{\iff\quad}
\end{equation}
\end{repthm}

\endgroup

\begin{proof}[Proof of \Cref{thm:pqe-general-low-dis-vio}]
    It immediately follows from \Cref{lemma:general-conv-combination-qparts-low-dis} and \Cref{thm:hiraguchi-po-dim} that any \qmet space with $n$ elements and \tw $t$ admits an embedding with \dis $\bigO(t \log^2 n)$ into a convex combination of $n^2$ \qparts, each represented with an intersection of $\bigO(n)$ total orders.

    Because the PQE class has concentration property, for any finite \qmet space, we can simply select a PQE that is close enough to the desired convex combination of $n^2$ \qparts, to obtain  \dis $\bigO(t \log^2 n)$.
    Since each Poisson process in PQE takes a constant number of latent dimensions, we can have such a PQE with $\bigO(n^3)$-dimensional latents and $n^2$ \qpart distributions.

    It remains only to prove that we can compute such required latents using the described architecture.

    Consider any $x \in \mathcal{X} \subset \R^d$. Since $\mathcal{X}$ is finite, we can always find direction $u_x \in \R^d$ such that $\forall y \in \mathcal{X} \setminus \{x\}$, $y\T u_x \neq x\T u_x$. That is, $x$ has a unique projection onto $u_x$. Therefore, we can have $c, b_+, b_- \in \R$ such that \begin{align}
        c \cdot u_x\T x + b_+ & = 1 \\
        -c \cdot u_x\T x + b_- & = 1,
    \end{align}
    but for $y \in \mathcal{X} \setminus \{x\}$, we have, for some $a > 0$, either \begin{align}
        c \cdot u_x\T y + b_+ & = -a \\
        -c \cdot u_x\T y + b_- & = a + 2,
    \end{align}
    or \begin{align}
        c \cdot u_x\T y + b_+ & = a + 2 \\
        -c \cdot u_x\T y + b_- & = -a.
    \end{align}

    Then, consider computing two of the first layer features as, on input $z$,  \begin{equation}
        [\mathrm{ReLU}(c \cdot u_x\T z + b_+) \quad
        \mathrm{ReLU}(-c \cdot u_x\T z + b_-)],
    \end{equation}
    which, if $z = x$,  is $[1, 1]$; if $z \neq x$, is either $[0, 2 + a]$ or $[2 + a, 0]$, for some $a > 0$.

    Then, one of the second layer features may sum these two features and threshold it properly would single out $x$, \ie, activate only when input is $x$.

    After doing this for all $x \in \mathcal{X}$, we obtain an $n$-dimensional second layer feature space that is just one-hot features.

    The third layer can then just be a simple embedding look up, able to represent any embedding, including the one allowing a PQE to have \dis  $\bigO(t \log n)$, as described above.

    Because \qmet embeddings naturally have \vio $1$, this concludes the proof.
\end{proof}

\subsubsection{Proof of \texorpdfstring{\Cref*{thm:pqe-lhgg-low-dis-vio}: \nameref*{thm:pqe-lhgg-low-dis-vio}}{Theorem: Distortion and Violation of PQEs}}
\begin{proof}[Proof of \Cref{thm:pqe-lhgg-low-dis-vio}]
    \Cref{lemma:pqelhgg-concentrate} and \Cref{thm:pqe-general-low-dis-vio} imply the result. To see that polynomial width is sufficient, note that the hidden width are polynomial by \Cref{thm:pqe-general-low-dis-vio}, and that the embedding dimensions needed to represent each of the $\bigO(n^3)$ Poisson processes is constant $1$ in both \PQELH and \PQEGG. Hence the latent space is also polynomial. This concludes the result.
\end{proof}

\subsubsection{Discussions}
\label{sec:discussion-qe-dis-bound}

\paragraph{Dependency on $\log n$.}
$\log n$ dependency frequently occurs in \dis results. Perhaps the most well-known ones are Bourgain's Embedding Theorem \citep{bourgain1985lipschitz} and the Johnson-Lindenstrauss Lemma \citep{johnson1984extensions}, which concern \emph{metric} embeddings into Euclidean spaces.

\paragraph{Dependency on \tw $t$. }
\Tw $t$ here works as a complexity measure of the \qmet. We will use a simple example to illustrate why low-\tw is easy. Consider the extreme case where the \qmet is the shortest-path distance on a tree, whose each edge is converted into two opposing directed ones and assigned arbitrary non-negative weights. Such a \qmet space has \tw $1$ (see \Cref{defn:qmet-tw}). On a tree, \begin{enumerate}[topsep=-3pt, itemsep=-1pt]
    \item the shortest path between two points is fixed, regardless of the weights assigned,
    \item for each internal node $u$ and one of its child $c$, the followings are \qparts: \begin{align*}
        d'_{01}(x, y) &\trieq \indic{\text{shortest path from $x$ to $y$ passes $(u, c)$}} \\
        d''_{01}(x, y) &\trieq \indic{\text{shortest path from $x$ to $y$ passes $(c, u)$}}.
    \end{align*}
\end{enumerate}
Hence it can be \emph{exactly} represented as a convex combination of \qparts. However, both of observations becomes false when the graph structure becomes more complex (higher \tw) and the shortest paths can are less well represented as tree paths of the tree composition.

\paragraph{Comparison with \uncon MLPs.}
\Cref{thm:pqe-general-low-dis-vio} requires a poly-width encoder to achieve low \dis. This is comparable with deep \uncon MLPs trained in NTK regime, which can reach $0$ training error (\dis $1$ on training set) in the limit but also requires polynomial width \citep{arora2019exact}.


\paragraph{\Qpmets and infinite distances. } \Cref{thm:pqe-general-low-dis-vio} relies on our assumptions that $(\mathcal{X}, d)$ is not a \qpmet space and has all finite distances. In fact, if we allow a PQE to have infinite convex combination weights, it can readily represent \qpmet spaces with infinite distances. Additionally, PQE can still well approximate the \qmet space with infinities replaced with any sufficiently large finite value (\eg, larger than the maximum finite distance). Thus, this limit is generally not important in practice (\eg, learning $\gamma$-discounted distances), where a large value and infinity are usually not treated much differently.

\paragraph{Optimizing \qmet embeddings.}
From \Cref{thm:pqe-general-low-dis-vio}, we know that optimizing PQEs over the training set $S$ \wrt \dis achieves low \dis (and optimal \vio by definition).  While directly optimizing \dis (or error on log distance or distance ratios, equivalently) seems a valid choice, such objectives do not always train stably in practice, with possible infinities and zeros. Often more stable losses are used, such as MSE over raw distances or $\gamma$-discounted distances $\gamma^d$, for $\gamma \in (0, 1)$. These objectives do not directly relate to \dis, except for some elementary loose bounds. To better theoretically characterize their behavior, an alternative approach with an average-case analysis might be necessary.

\subsection{Implementation of \PQEs (PQEs)}\label{sec:pqe-impl-full}

\Cref{sec:general-pqe} mentioned a couple implementation techniques for PQEs. In this section, we present them in full details.

\subsubsection{Normalized Measures}

Consider a PQE whose each of $j$ expected \qparts is defined via  $k$ Poisson processes, with set parametrizations $u \rightarrow A_{i,j}(u), i \in [h], j \in [k]$. To be robust to the choice of $k$, we instead use the normalized set parametrizations $A'_{i,j}$: \begin{equation}
    A'_{i,j}(u) \trieq A_{i,j}(u) / k, \mathrlap{\qquad\qquad i \in [h], j \in [k].}
\end{equation}

This does not change the PQE's concentration property (\Cref{defn:pqe-concentration}) or its theoretical guarantees (\eg, \Cref{thm:pqe-lhgg-low-dis-vio,thm:pqe-general-low-dis-vio}).

\subsubsection{Outputting $\gamma$-Discounted Distances}\label{sec:learn-gamma-discount-pqe-distances}

Recall the PQE \qmet formulation in \Cref{eq:pqe}, for $\alpha_i \geq 0$, and encoder $f \colon \mathcal{X} \rightarrow \R^d$: \begin{equation}
    \hat{d}(x, y) \trieq
    \sum_{i} \alpha_i \bigg(1 - \prod_{j}
    \mathbb{P}\left[\pois(\mu_{i,j}(A_{i,j}({f(x)}) \setminus A_{i,j}({f(y)}))) \leq \pois(\mu_{i,j}(A_{i,j}({f(y)}) \setminus A_{i,j}({f(x)})))\right]
    \bigg). \tag{\ref{eq:pqe}}
\end{equation}

With discount factor $\gamma \in (0, 1)$, we can write the $\gamma$-discounted PQE distance as \begin{equation}
    \gamma ^ {\hat{d}(x, y)} =
    \prod_{i} (\underbrace{\gamma^{\alpha_i}}_{\mathclap{\vphantom{2^{2}}\qquad\qquad\qquad\qquad\qquad\textup{a scalar that can take value in any $(0, 1)$}}})^{ 1 - \prod_{j}
    \mathbb{P}\left[\pois(\mu_{i,j}(A_{i,j}({f(x)}) \setminus A_{i,j}({f(y)}))) \leq \pois(\mu_{i,j}(A_{i,j}({f(y)}) \setminus A_{i,j}({f(x)})))\right]}.
\end{equation}

Therefore, instead of learning $\alpha_i \in [0, \infty)$, we can learn bases $\beta_i \in (0, 1)$ such and define the $\gamma$-discounted PQE distance as \begin{equation}
    \gamma ^ {\hat{d}(x, y)} \trieq
    \prod_{i} \beta_i^{ 1 - \prod_{j}
    \mathbb{P}\left[\pois(\mu_{i,j}(A_{i,j}({f(x)}) \setminus A_{i,j}({f(y)}))) \leq \pois(\mu_{i,j}(A_{i,j}({f(y)}) \setminus A_{i,j}({f(x)})))\right]}.
\end{equation}

These bases $\beta_i \in (0, 1)$ can be parametrized via a sigmoid transform. Consider \qmet learning \wrt errors on $\gamma$-discounted distances (\eg, MSE).  Unlike the parametrization with directly learning the convex combination weights $\alpha_i$'s, such a parametrization (that learns the bases $\beta_i$'s) does not explicitly include $\gamma$ and thus can potentially be more stable for a wider range of $\gamma$ choices.

\paragraph{Initialization.} Consider learning bases $\beta_i$'s via a sigmoid transform: learning $b_i$ and defining $\beta_i \trieq \sigma(b_i)$. We must take care in initializing these $b_i$'s so that $\sigma(b_i)$'s are not too close to $0$ or $1$, since we take a product of powers with these bases. To be robust to different $h$ numbers of \qpart distributions, we initialize the each $b_i$ to be from the uniform distribution \begin{equation}
    \mathcal{U}[\sigma^{-1}(0.5^{2/h}), \sigma^{-1}(0.75^{2/h})], \label{eq:sigmoid-trick-init-distn}
\end{equation}
which means that, at initialization, \begin{equation}
    \prod_{i \in [h]} \beta_i^{0.5} = \prod_{i \in [h]} \sigma(b_i)^{0.5} \in [0.5, 0.75],
\end{equation}
providing a good range of initial outputs, assuming that the exponents (expected outputs of \qpart distributions) are close to $0.5$. Alternatively, $b_i$'s maybe parametrized by a deep linear network, a similar initialization is employed. See \Cref{sec:deep-linear-net} below for details.

\subsubsection{Learning Linear/Convex Combinations with Deep Linear Networks}\label{sec:deep-linear-net}

Deep linear networks have the same expressive power as regular linear models, but enjoy many empirical and theoretical benefits in optimization \citep{saxe2013exact,pennington2018emergence,huh2021low}. Specifically, instead of directly learning a matrix $\in \R^{m\times n}$, a deep linear network (with bias) of $l$ layers learns a sequence of matrices\begin{align}
    M_1 & \in \R^{m_1\times n} \\
    M_2 & \in \R^{m_2\times m_1} \\
    \vdots & \hphantom{{}\in{}} \vdots \\
    M_{l-1} & \in \R^{m_{l-1}\times m_{l-2}} \\
    M_l & \in \R^{m\times m_{l-1}} \label{eq:deep-linear-net-mats}\\
    B & \in \R^{m \times n},  \label{eq:deep-linear-net-bias}
\end{align}
where the linear matrix can be obtained with \begin{equation}
    M_l\ M_{l-1} \dots M_2\ M_1 + B,
\end{equation}
and we require \begin{equation}
    \min(m_1, m_2, \dots, m_{l-1}) \geq \min(m, n).
\end{equation}

In our case, the convex combination weights for the \qpart distributions often need to be large, in order to represent large \qmet distances; in Poisson process mean measures with learnable scales (\eg, the Gaussian-based measure described in \Cref{sec:gaussian-gaussian}), the scales may also need to be large to approximate particular \qparts (see \Cref{sec:require-qpart-distn-concentrate}).

Therefore, we choose to use deep linear networks to optimize these parameters. In particular, \begin{itemize}
    \item \textbf{For the convex combination weights for $h$ \qpart distributions,}   \begin{itemize}
        \item When learning the convex combination weights $\{\alpha_i\}_{i\in[h]}$, we use a deep linear network to parametrize a matrix $\in \R^{1 \times h}$ (\ie, a linear map from $\R^h$ to $\R$), which is then viewed as  a vector $\in \R^h$ and applied an element-wise square transform $a \rightarrow a^2$ to obtain non-negative weights $\alpha \in [0, \infty)^h$;

        \item When learning the bases for discounted \qmet distances $\beta_i$'s (see \Cref{sec:learn-gamma-discount-pqe-distances}), we use a deep linear network to parametrize a matrix $\in \R^{h \times 1}$, which is then viewed as  a vector $\in \R^h$ and applied an element-wise sigmoid transform $a \rightarrow \sigma(a)$  to obtain bases $\beta \in (0, 1)^{h}$.

        Note that here we parametrize a matrix $\in \R^{h \times 1}$ rather than $\R^{1 \times h}$ as above for $\alpha_i$'s. The reason for this choice is entirely specific to the initialization scheme we use (\ie, (fully-connected layer weight matrix initialization, as discussed below). Here the interpretation of a linear map is no longer true. If we use $\R^{1 \times h}$, the initialization method would lead to the entries distributed with variance roughly $1/n$, which only makes sense if they are then added together. Therefore, we use $\R^{h \times 1}$, which would lead to constant variance.
    \end{itemize}

    \item \textbf{For scales of the Poisson process mean measure, such as \PQEGG,} we consider a slightly different strategy.

    Consider a PQE formulation with $h \times k$ independent Poisson processes, from which we form $h$ \qpart distributions, each from $k$ total orders parametrized by $k$ Poisson processes. The Poisson processes are defined on sets \begin{equation}
        \{A_{i,j}\}_{i\in[h], j\in[k]},
    \end{equation}use mean measures \begin{equation}
        \{\mu_{i,j}\}_{i\in[h], j\in[k]},
    \end{equation}
    and set parametrizations \begin{equation}
        \{ u \rightarrow A_{i,j}(u)\}_{i\in[h], j\in[k]},
    \end{equation}
    to compute quantities \begin{equation}
        \mu_{i,j}(A_{i,j}(u) \setminus A_{i,j}(v))\qquad \text{for $u \in \R^d, v \in \R^d, i\in[h], j\in[k]$}. \label{eq:scaled-measure-set-diffs}
    \end{equation}

    \myparagraph{Scaling each mean measure independently.}
    Essentially, adding learnable scales (of mean measures) $w \in [0, \infty)^{h\times k}$ (or, equivalently, $\{w_{i,j} \in [0, \infty)\}_{i, j}$) gives a scaled set of measures \begin{equation}
        \{w_{i,j} \cdot \mu_{i,j}\}_{i\in[h], j\in[k]}.
    \end{equation}

    This means that the quantities in \Cref{eq:scaled-measure-set-diffs} becomes respectively scaled as \begin{equation}
        w_{i,j} \cdot \mu_{i,j}(A_{i,j}(u) \setminus A_{i,j}(v))\qquad \text{for $u \in \R^d, v \in \R^d, i\in[h], j\in[k]$}.
    \end{equation}

    \myparagraph{Convex combinations of \emph{all} measures.}
    However, we can be more flexible here, and allow not just scaling each measure independently, but also \emph{convex combinations of all measures}. Instead of having $w$ as a collection of $h \times k$ scalar numbers $\in [0, \infty)$, we have a collection of $(h \times k)$ vectors each having length $(h \times k)$ (or $h \times k$-shape tensors) \begin{equation}
        \{w_{i,j} \in [0, \infty)^{h \times k}\}_{i\in[h], j\in[k]},
    \end{equation}
    and have the quantities in \Cref{eq:scaled-measure-set-diffs} respectively scaled and combined as \begin{equation}
        \sum_{i', j'} w_{i,j,i',j'} \cdot \mu_{i',j'}(A{i',j'}(u) \setminus A_{i',j'}(v))\qquad \text{for $u \in \R^d, v \in \R^d, i\in[h], j\in[k]$}.
    \end{equation}
    Note that these still are valid Poisson processes for a PQE. Specifically, the new Poisson processes now all use the same set parametrization (as the collection of original ones), with different measures (as different weighted combinations of the original measures).
    This generalizes the case where each mean measure is scaled independently (as $w$ can be diagonal).

    Therefore, we will apply this more general strategy using \textbf{convex combinations of \emph{all} measures}.

    Similarly to learning the convex combination weights of \qpart distributions, we collapse a deep linear network into a tensor $\in \R^{h\times k \times h \times k}$, and apply an element-wise square $a \rightarrow a^2$, result of which is used as the convex combination weights  $w$ to ensure non-negativity.
\end{itemize}

\paragraph{Initialization.}
For initializing the matrices $(M_1, M_2, \dots, M_l)$ of a deep linear network (\Cref{eq:deep-linear-net-mats}), we use the standard weight matrix initialization of fully-connected layers in PyTorch \citep{paszke2019pytorch}. The bias matrix $B$ (\Cref{eq:deep-linear-net-bias}) is initialized to all zeros.

When used for learning the bases for discounted \qmet distances $\beta_i$'s (as described in \Cref{sec:learn-gamma-discount-pqe-distances}), we have a deep linear network parametrizing a matrix $\in \R^{h \times 1}$, initialized in the same way as above (including initializing $B$ as all zeros). Consider the matrix up to before the last one: \begin{equation}
    M^* \trieq M_{l-1} \cdot M_2\ M_1 \in \R^{m_{l-1} \times 1}.
\end{equation} $M^*$ is essentially a projection to be applied on each row of the last matrix $M_l \in \R^{h \times m_{l-1}}$, to obtain $b_i$ (which is then used to obtain bases  $\beta_i \trieq \sigma(b_i)$). Therefore, we simply rescale the $M^*$ subspace for each row of $M_l$ and keep the orthogonal space intact, such that the projections would be distributed according to the distribution specified in \Cref{eq:sigmoid-trick-init-distn}: \begin{equation}
    \mathcal{U}[\sigma^{-1}(0.5^{2/h}), \sigma^{-1}(0.75^{2/h})], \tag{\ref{eq:sigmoid-trick-init-distn}},
\end{equation}
which has good initial value properties, as shown in \Cref{sec:learn-gamma-discount-pqe-distances}.

\subsubsection{Choosing $h$ the Number of \Qpart Distributions and $k$ the Number of Poisson Processes for Each \Qpart Distribution}\label{sec:param-choose-h-k}

A PQE (class) is defined with $h \times k$ independent Poisson processes with means $\{\mu_{i,j}\}_{i\in[h],j\in[k]}$ along with $h \times k$ set parametrizations $\{A_{i,j}\}_{i\in[h],j\in[k]}$. For $k$ pairs of means and set parametrizations, we obtain a random \qpart. A mixture (convex combination) of the resulting $h$ random \qparts gives the \qmet. The choices of $\mu$ and $A$ are flexible. In this work we explore \PQELH and \PQEGG as two options, both using essentially the same measure and parametrization across all $i, j$ (up to individual learnable scales). These two instantiations both perform well empirically. In this section we aim to provide some intuition on choosing these two hyperparameters $h$ and $k$.

\paragraph{$h$ the Number of \Qpart Distributions}

Theoretical result \Cref{thm:pqe-general-low-dis-vio}  suggest thats, for  a \qmet space with $n$ elements, $n^2$ \qpart distributions suffice to learn a low \dis embedding. Since this is a worst-case result, the practical scenario may require much fewer \qparts. For instance, \Cref{sec:discussion-qe-dis-bound} shows that $\bigO(n)$ \qparts is sufficient for any \qmet space with a tree structure. In our experiments, $h \in [8, 128]$ \qpart distributions are used.

\paragraph{$k$ the Number of Poisson Processes for Each \Qpart Distribution (Random Partial Order)}

It is well-known that such intersection of sufficiently many total orders can represent \emph{any} partial order \citep{trotter1995partially,hiraguchi1951dimension}. This idea is equivalent with the dominance drawing dimension of directed graphs \citep{ortali2019multidimensional}, which concerns an order embedding of the vertices to preserve the poset specified by the reachability relation. In this graph theoretical view, several results are known.  \citep{FELSNER20101097} prove that planar graphs have at most $8$ dimension. \citep{ortali2019multidimensional} show that the dimension of any graph with $n$ vertices is at most $\min(w_P, \frac{n}{2})$, where $w_P$ the maximum size of a set of incomparable vertices. A simpler and more fundamental result can be traced to \citeauthor{hiraguchi1951dimension} from \citeyear{hiraguchi1951dimension}:

\begingroup  
\let\repthmXRRR\undefined
\begin{repthm}[Hiraguchi's Theorem \citep{hiraguchi1951dimension,bogart1973maximal}]{thm}{thm:hiraguchi-po-dim}{thmHiraguchiPODim}
Let $(X, P)$ be a partially ordered set such that $\size{X} \geq 4$. Then there exists a mapping $f \colon X \rightarrow \R^{\floor{\size{X} / 2}}$ such that \begin{equation}
    \forall x, y \in X, \qquad x P y \iff f(x) \leq f(y)~\coorwise. \hphantom{\iff\quad}
\end{equation}
\end{repthm}

\endgroup

\Cref{thm:hiraguchi-po-dim} states that $\frac{n}{2}$ dimensions generally suffice for any poset of size $n \geq 4$.

In our formulation, this means that using $k = \frac{n}{2}$ Poisson processes (giving $\frac{n}{2}$ random total orders) will be maximally expressive. In practice, this is likely unnecessary and sometimes impractical. In our experiments, we choose a small fixed number $k=4$.

\section{Experiment Settings and Additional Results}

\paragraph{Computation power.}All our experiments run on a single GPU and finish within 3 hours. GPUs we used include NVIDIA 1080, NVIDIA 2080 Ti, NVIDIA 3080 Ti, NVIDIA Titan Xp, NVIDIA Titan RTX, and NVIDIA Titan V.

\subsection{Experiments from \texorpdfstring{\Cref*{sec:failure-uncon-metric-emb}: \nameref*{sec:failure-uncon-metric-emb}}{Toy Example Section}}\label{sec:3-node-expr-settings}

In \Cref{sec:failure-uncon-metric-emb} and \Cref{fig:3-node}, we show experiment results on a simple $3$-element \qmet space.

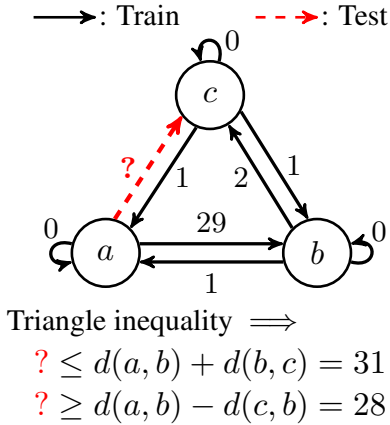
\begin{figure}
    \centering%
    \scalebox{1.25}{%
       \begin{tikzpicture}[bayes_net, node distance = 1.5cm, every node/.style={inner sep=0}]
            \node[main_node, minimum size=0.72cm] (a) {$a$};
            \node[main_node, minimum size=0.72cm] (b) [right = 1.5cm of a] {$b$};
            \node[main_node, minimum size=0.72cm] (c) [above right = 1.15cm and 0.58cm of a] {$c$};

            \node[] (legend) [above = 0.36cm of c, inner sep=0pt] {\small

            \raisebox{2pt}{\protect\tikz{\protect\draw[line width=1.5pt, ->, thick, >=stealth'] (0, 0) -- (0.7, 0);}}\hspace{-0.5pt}: Train\hspace{5ex}
            \raisebox{2pt}{\protect\tikz{\protect\draw[line width=1.5pt, ->, thick, >=stealth', dashed, red] (0, 0) -- (0.7, 0);}}\hspace{-0.5pt}: Test
            };

            \node[] (constraints) [below = 1.9cm of c, inner sep=0pt, align=left] {\shortstack{\small
            Triangle inequality $\implies$\hphantom{wqewcm}\\[0.1ex]
            $
                \begin{aligned}
                    {\color{red} ? } & \leq d(a, b) + d(b, c) = 31 \\[-1ex]
                    {\color{red} ? } & \geq d(a, b) - d(c, b) = 28
                \end{aligned}
            $}};
            \node[] (filler) [below = 0.05cm of constraints, inner sep=0pt] {};

            \tikzset{myptr/.style={decoration={markings,mark=at position 1 with %
                {\arrow[scale=0.825,>=stealth']{>}}},postaction={decorate}}}
            \path[]
            (a) edge[out=165,in=195,-, myptr, line width=1pt,looseness=4] [right] node [above=4pt] {\small$0$} (a)
            (c) edge[out=75,in=105,-, myptr, line width=1pt,looseness=4] [right] node [right=4pt] {\small$0$} (c)
            (b) edge[out=345,in=15,-, myptr, line width=1pt,looseness=4] [right] node [above right=3pt and -0.5pt] {\small$0$} (b)
            (a.15) edge[-, myptr, line width=1pt] [right] node [midway, above=3pt] {\small$29$} (b.165)
            (b.135) edge[-, myptr, line width=1pt] [right] node [midway, below left=-3pt and 3pt] {\small$2$} (c.302)
            (b.195) edge[-, myptr, line width=1pt] [right] node [midway, below=3pt] {\small$1$} (a.-15)
            (c.332) edge[-, myptr, line width=1pt] [right] node [midway, right=3pt] {\small$1$} (b.105)
            (c.247) edge[-, myptr, line width=1pt] [right] node [midway, right=3pt] {\small$1$} (a.45)
            (a.75) edge[-, myptr, dashed, line width=1.3pt, red] [right] node [left=3pt] {\color{red} \small$\boldsymbol{?}$} (c.217)
            ;
        \end{tikzpicture}%
    }
    \caption{The $3$-element \qmet space, and the training pairs.%
    Training set contains all pairs except for $(a, c)$. Arrows show \qmet distances (rather than edge weights of some graph).}    \label{fig:3-node-qmet-space}
\end{figure}

\paragraph{\Qmet space.} The \qmet space has $3$ elements with one-hot features $\in \R^3$. The\qmet and training pairs are shown in \Cref{fig:3-node-qmet-space}.

\paragraph{\Uncon network.}
The unconstrained network has architecture 6-128-128-32-1, with ReLU activations.

\paragraph{Metric embedding.}
The embedding space is $32$-dimensional, upon which corresponding metric is applied. The encoder network has architecture 6-128-128-32, with ReLU activations.

\paragraph{Asymmetric dot products.}
The embedding space is $32$-dimensional. The two inputs are encoded with a \emph{different} encoder of architecture 6-128-128-32, with ReLU activations. Then the dot product of the two $32$-dimensional vector is taken, which parametrizes a distance estimate

\paragraph{\PQEs.}
The embedding space is $32$-dimensional, which parametrizes $8$ \qmet distributions, each from $4$ independent Poisson processes using (scaled) Lebesgue measure and half-lines. We use deep linear networks, as described in \Cref{sec:deep-linear-net}. A deep linear network (without bias) of architecture 8-32-32-1 parametrizes the convex combination weights $\{\alpha_i\}_{i \in [8]}$. Another deep linear network (without bias) of architecture  32-64-64-32 parametrizes convex combination weights of the mean measures $d \in [0, \infty)^{32 \times 32}$. Note that these \emph{do not} give many more effective parameters to PQEs as they are equivalent with simple linear transforms.

\paragraph{Optimization.}
All models are trained \wrt MSE on distances with the Adam optimizer \citep{kingma2014adam} with learning rate $0.0003$ for $1000$ iterations (without mini-batching since the training set has size $8$).

\paragraph{Additional results.}
Results with additional formulations (together with the ones presented in \Cref{fig:3-node}) are shown in \Cref{fig:3-node-additional}.
\begin{figure}
    \begin{subfigure}[t]{0.315\linewidth}
        \centering
        \includegraphics[scale=0.52]{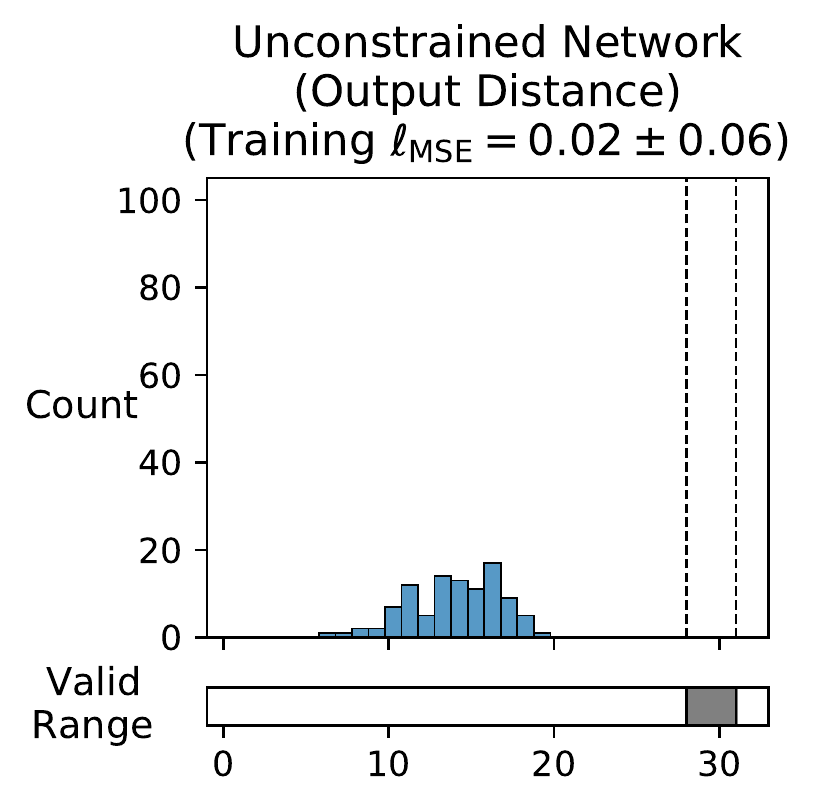}
        \caption{\Uncon network that directly predicts distance.}
    \end{subfigure}\hfill
    \begin{subfigure}[t]{0.315\linewidth}
        \centering
        \includegraphics[scale=0.52]{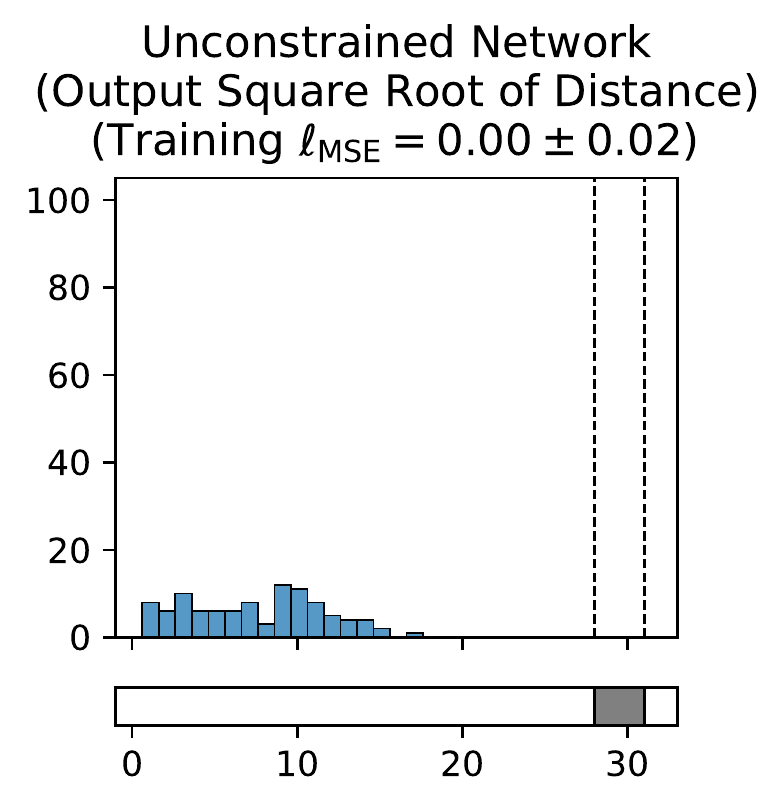}
        \caption{\Uncon network that predicts distance with a square $a \rightarrow a^2$ transform.}
    \end{subfigure}\hfill
    \begin{subfigure}[t]{0.315\linewidth}
        \centering
        \includegraphics[scale=0.52]{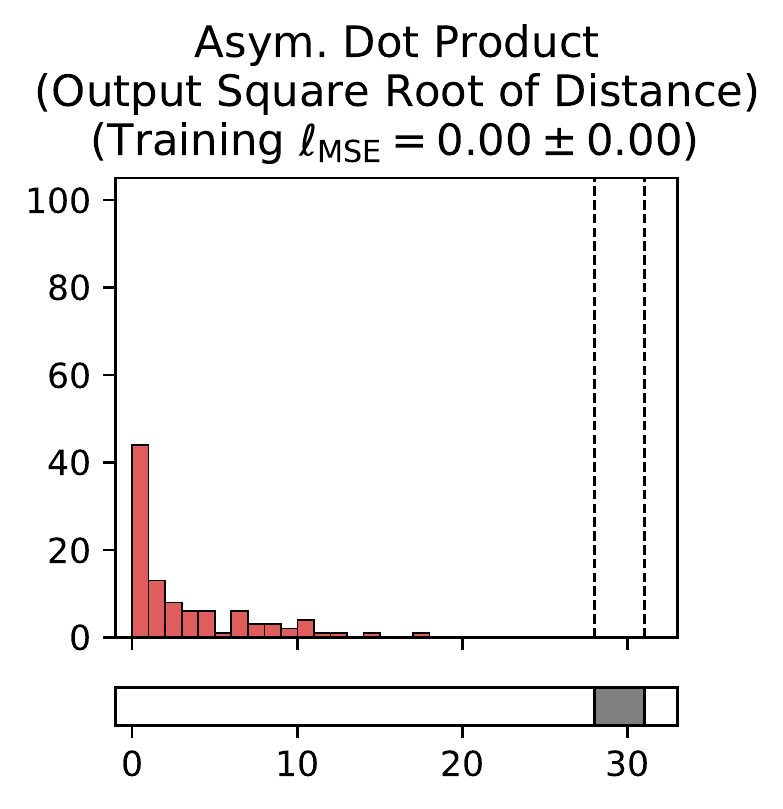}
        \caption{Asymmetrical dot product that predicts distance with a square $a \rightarrow a^2$ transform.}
    \end{subfigure}\\
    \begin{subfigure}[t]{0.315\linewidth}
        \centering
        \includegraphics[scale=0.52]{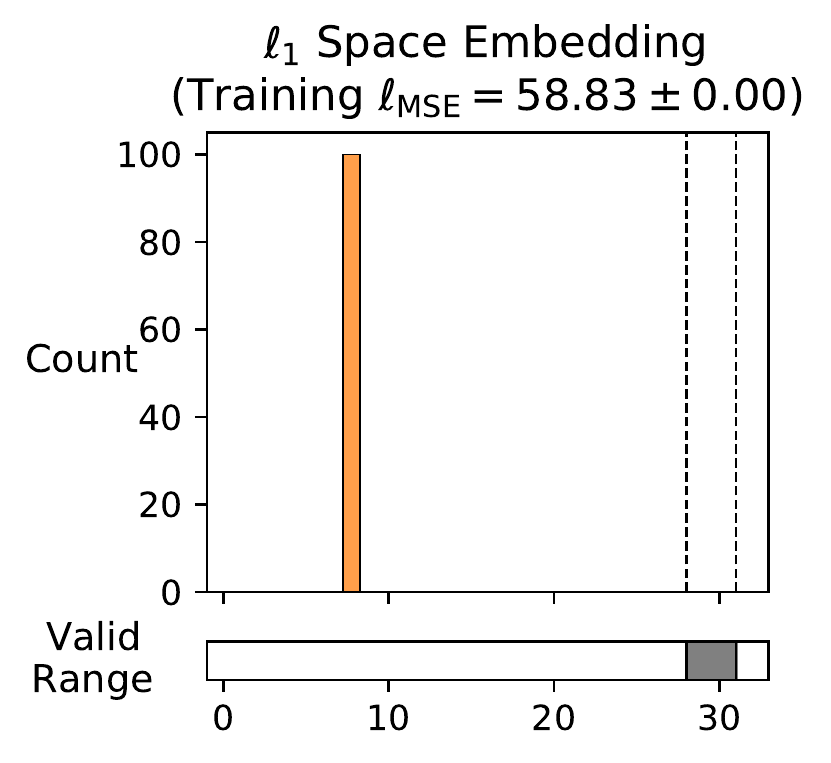}
        \caption{Metric embedding into an $\ell_1$ space.}
    \end{subfigure}\hfill
    \begin{subfigure}[t]{0.315\linewidth}
        \centering
        \includegraphics[scale=0.52]{figures/3_nodes/encl2.pdf}
        \caption{Metric embedding into an Euclidean space.}
    \end{subfigure}\hfill
    \begin{subfigure}[t]{0.315\linewidth}
        \centering
        \includegraphics[scale=0.52]{figures/3_nodes/pqe.pdf}
        \caption{\PQE specified in \Cref{sec:3-node-expr-settings}.}
    \end{subfigure}

    \caption{Training different formulations to fit training pairs distances via MSE, and using them to predict on the test pair. Plots show distribution of the prediction over $100$ runs. Standard deviations of the training error are shown.
    }    \label{fig:3-node-additional}
\end{figure}

\subsection{Experiments from \texorpdfstring{\Cref*{sec:expr}: \nameref*{sec:expr}}{Experiments Section}}

\paragraph{Triangle inequality regularizer.} For methods that do not inherently respect triangle inequalities (\eg, \uncon networks and asymmetrical dot products), we explore training with a regularizer that encourages following these inequalities. By sampling random triplets uniformly over the training set, the regularizer is formulated as, \begin{equation}
    \mathbb{E}_{x,y,z}\big[\max(0, \gamma^{\hat{d}(x, y) + \hat{d}(y, z)} - \gamma^{\hat{d}(x, z)})^2\big],
\end{equation}%
where the $\gamma$-discounted terms and the squared form allows easier balancing with the training loss, which, across all experiments, are MSEs on some $\gamma$-discounted distances.

\paragraph{PQE settings.}
Across all experiments of this section, when given an encoder architecture mapping input to an $\R^d$ latent space, we construct PQEs according to the following general recipe, to obtain the two PQEs settings used across all experiments: \PQELH (PQE with Lebesgue measure and half-lines) and \PQEGG (PQE with Gaussian-based measure and Gaussian shapes, see see \Cref{sec:gaussian-gaussian}): \begin{itemize}
    \item (Assuming $d$ is a multiple of $4$,) We use $h \trieq = d/4$ \qpart distributions, each given by $k \trieq 4$ Poisson processes;
    \item A deep linear network (see \Cref{sec:deep-linear-net}), is used for parametrizing the convex combination weights $\alpha \in \R^{d/4}$ or the bases $\beta \in \R^{d/4}$ (see \Cref{sec:learn-gamma-discount-pqe-distances}), we follow the initialization and parametrization described in \Cref{sec:deep-linear-net}, with hidden sizes $[n_\mathsf{hidden}, n_\mathsf{hidden}, n_\mathsf{hidden}]$ (\ie, $4$ matrices/layers), where $n_\mathsf{hidden} \trieq \max(64, 2^{1 + \ceil{\log_2 (d/4)}})$.
    \item  For \PQEGG, \begin{itemize}
        \item The learnable $\sigma^2_\mathsf{measure} \in (0, \infty)^d$ (one for each Poisson Process) is achieved by optimizing the log variance, which is initialized as all zeros.
        \item The Gaussian-based measures need learnable scales. We use a deep linear network to parametrize the $[0, \infty)^{d\times d}$ weights for the convex combinations of measures, as described in \Cref{sec:deep-linear-net}. Similarly, it has hidden sizes $[n_\mathsf{hidden}, n_\mathsf{hidden}, n_\mathsf{hidden}]$ (\ie, $4$ matrices/layers), where $n_\mathsf{hidden} \trieq \max(64, 2^{1 + \ceil{\log_2 d}})$.
    \end{itemize}
\end{itemize}
Note that the PQEs add only a few extra effective parameters on top of the encoder ($d$ for \PQELH, and $d+d^2$ for \PQEGG), as the deep linear networks do not add extra effective parameters.

\paragraph{Mixed space metric embedding settings.}
Across all experiments of this section, when given an encoder architecture mapping input to an $\R^d$ latent space, we construct the metric embedding into mixed space as follows:\begin{itemize}
    \item (Assuming $d$ is a multiple of $4$,) We use (1) a $(d/2)$-dimensional Euclidean space (2) a $(d/4)$-dimensional $\ell_1$ space, and a $(d/4)$-dimensional spherical distance space (without scale).
    \item Additionally, we optimize three scalar values representing the log weights of the convex combination to mix these spaces.
\end{itemize}

\paragraph{DeepNorm and WideNorm method overview and parameter count comparison with PQEs.}Both DeepNorm and WideNorm parametrize asymmetrical norms. When used to approximate \qmets, they are applied as $\hat{d}(x, y) \trieq f_\mathsf{AsymNorm}(f_\mathsf{Enc}(x) - f_\mathsf{Enc}(y))$, where $f_\mathsf{Enc}$ is the encoder mapping from data space to an $\R^d$ latent space and $f_\mathsf{AsymNorm}$ is either the DeepNorm or the WideNorm predictor on that latent space \citep{pitis2020inductive}. \begin{itemize}
    \item \textbf{DeepNorm} is a modification from Input Convex Neural Network (ICNN; \citet{amos2017input}), with restricted weight matrices and activation functions for positive homogeneity (a requirement of asymmetrical norms), and additional concave function for expressivity.

    For an input latent space of $\R^d$, consider an $n$-layer DeepNorm with width $w$ (\ie, ICNN output size) and the suggested intermediate MaxReLU activation and MaxMean final aggregation (see \citep{pitis2020inductive} for details of these functions). This DeepNorm predictor $f_\mathsf{DeepNorm}$ (on latent space) has \begin{align*}
        \text{\#parmaters of $f_\mathsf{DeepNorm}$}
        & = \underbrace{n \times (d \times w)}_{\text{$U$ matrices from input to each layer}} \\
        & + \underbrace{(n - 1) \times w^2}_{\text{$W$ matrices between neighboring layer activations}}  \\
        & + \underbrace{n \times w}_{\text{intermediate MaxReLU activations}}  \\
        & + \underbrace{w \times (4 + 5)}_{\text{concave function (with $5$ components) parameters}}  \\
        & + \underbrace{1}_{\text{final MaxMean aggregation}},
    \end{align*}
    which is on the order of $\bigO(nw\max(d, w))$. In the common case where the hidden size $w$ is  chosen to be on the same magnitude as $d$, this becomes $\bigO(nd^2)$.

    \item \textbf{WideNorm} is based on the observation that \begin{equation}
        x \rightarrow \norm{W\ \mathrm{ReLU}(x \mathbin{::} -x)}_2
    \end{equation}
    is an asymmetric norm when $W$ is non-negative, where $\mathbin{::}$ denotes vector concatenation. WideNorm then learns many such norms each with a different $W$ matrix parameter,  before (again) feeding the norm values into a concave function and aggregating them together with MaxMean.

    For an input latent space of $\R^d$, consider a WideNorm with $c$ such learned norms with $W$ matrices of shape $\R_{\geq>0}^{(2d)\times w}$. This WideNorm predictor $f_\mathsf{WideNorm}$ (on latent space), has \begin{align*}
        \text{\#parmaters of $f_\mathsf{WideNorm}$}
        & = \underbrace{c \times (2d \times w)}_{\text{$W$ matrices}} \\
        & + \underbrace{c \times (4 + 5)}_{\text{concave function (with $5$ components) parameters}}  \\
        & + \underbrace{1}_{\text{final MaxMean aggregation}},
    \end{align*}
    which is on the order of $\bigO(cdw)$. In the common case where both the number of components $c$ and the output size of each component (before applying the $l2$-norm) are  chosen to be on the same magnitude as $d$, this becomes $\bigO(d^3)$.

\end{itemize}
For both DeepNorm and WideNorm, their parameter counts are much larger than the number of effective parameters of PQEs ($d$ for \PQELH and $d+d^2$ for \PQEGG). For a $256$-dimensional latent space, this difference can be on the order of $10^6\sim10^7$.

\paragraph{DeepNorm and WideNorm settings.}
Across all experiments of this section, we evaluate 2 DeepNorm settings and 3 WideNorm settings, all derived from the experiment setting of the original paper \citep{pitis2020inductive}.  For both DeepNorm and WideNorm, we use MaxReLU activations, MaxMean aggregation, and concave function of $5$ components. For DeepNorm, we use $3$-layer networks with 2 different hidden sizes: 48 and 128 for the $48$-dimensional latent space in random directed graphs experiments, 512 and 128 for the $512$-dimensional latent space in the large-scale social graph experiments, 128 and 64 for the $128$-dimensional latent space in offline Q-learning experiments. For WideNorm, we components of size $32$ and experiment with 3 different numbers of components: $32$, $48$, and $128$.

\paragraph{Error range.} Results are gathered across $5$ random seeds, showing both averages and population standard deviations.


\subsubsection{Random Directed Graphs \Qmet Learning}
\paragraph{Graph generation.}
The random graph generation is controlled by three parameters $d$, $\rho_\mathsf{un}$ and $\rho_\mathsf{di}$. $d$ is the dimension of the vertex features. $\rho_\mathsf{un}$ specifies the fraction of pairs that should have at least one (directed) edge between them. $\rho_\mathsf{di}$ specifies the fraction of \emph{such} pairs that should \emph{only} have one (directed) edge between them. Therefore, if $\rho_\mathsf{un}=1, \rho_\mathsf{di}=0$, we have a fully connected graph; if $\rho_\mathsf{un}=0.5, \rho_\mathsf{di}=1$, we have a graph where half of the vertex pairs have exactly one (directed) edge between them, and the other half are not connected. For completeness, the exact generation procedure for a graph of $n$ vertices is the following: \begin{enumerate}
    \item randomly add $\rho_\mathsf{un} \cdot n^2$ undirected edges, each represented as two opposite directed edges;
    \item optimize $\R^{n\times d}$ vertex feature matrix using Adam \citep{kingma2014adam} \wrt $\mathcal{L}_\mathsf{align}(\alpha=2) + 0.3 \cdot \mathcal{L}_\mathsf{uniform}(t=3)$ from \citep{wang2020hypersphere}, where each two node is considered a positive pair if they are connected;
    \item randomly initialize a network $f$ of architecture $d$-4096-4096-4096-4096-1 with $\tanh$ activations;
    \item for each connected vertex pair $(u,v)$, obtain $d_{u\rightarrow v} \trieq f(\mathsf{feature}(u)) - f(\mathsf{feature}(v))$ and $d_{v\rightarrow u} = -d_{u\rightarrow v}$;
    \item for each $(u, v)$ such that $d_{u\rightarrow v}$ is among the top $1 - \rho_\mathsf{di} / 2$ of such values (which is guaranteed to not include both directions of the same pair due to symmetry of $d_{u\rightarrow v}$), make $v \rightarrow u$ the only directed edge between $u$ and $v$.
\end{enumerate}

We experiment with three graphs of $300$ vertices and $64$-dimensional vertex features: \begin{itemize}
    \item \textbf{\Cref{fig:random-graph-dense}: }A graph generated with $\rho_\mathsf{un}=0.15, \rho_\mathsf{di}=0.85$;
    \item \textbf{\Cref{fig:random-graph-sparse}: }A sparser graph generated with $\rho_\mathsf{un}=0.05, \rho_\mathsf{di}=0.85$;
    \item \textbf{\Cref{fig:random-graph-blocksparse}: }A sparse graph with block structure by \begin{enumerate}
        \item generating $10$ small dense graphs of $30$ vertices and $32$-dimensional vertex features, using $\rho_\mathsf{un}=0.18, \rho_\mathsf{di}=0.15$,
        \item generating a sparse $10$-vertex ``supergraph'' with $32$-dimensional vertex features, using  $\rho_\mathsf{un}=0.22, \rho_\mathsf{di}=0.925$,
        \item for each supergraph vertex \begin{enumerate}
            \item associating it with a different small graph,
            \item for all vertices of the small graph, concatenate the supergraph vertex's feature to the existing feature, forming $64$-dimensional vertex features for the small graph vertices,
            \item picking a random representative vertex from the small graph,
        \end{enumerate}
        \item connecting all $10$ representative vertices in the same way as their respective supergraph vertices are connected in the supergraph.
    \end{enumerate}
\end{itemize}

\paragraph{Architecture.} All encoder based methods (PQEs, metric embeddings, dot products) use 64-128-128-128-48 network with ReLU activations, mapping $64$-dimensional inputs to a $48$-dimensional latent space. \Uncon networks use a similar 128-128-128-128-48-1 network, mapping concatenated the $128$-dimensional input to a scalar output.

\paragraph{Data.} For each graph, we solve the groundtruth distance matrix and obtain $300^2$ pairs, from which we randomly sample the training set, and use the rest as the test set. We run on $5$ training fractions evenly spaced on the logarithm scale, from $0.01$ to $0.7$.

\paragraph{Training.} We use $2048$ batch size with the Adam optimizer \citep{kingma2014adam}, with learning rate decaying according to the cosine schedule without restarting \citep{loshchilov2016sgdr} starting from $10^{-4}$ to $0$ over $3000$ epochs. All models are optimized \wrt MSE on the $\gamma$-discounted distances, with $\gamma=0.9$. When running with the triangle inequality regularizer, $683 \approx 2048/3$ triplets are uniformly sampled at each iteration.

\paragraph{Full results and ablation studies.}
\Cref{fig:random-graph-dense,fig:random-graph-sparse,fig:random-graph-blocksparse} show full results of all methods running on all three graphs. In \Cref{fig:random-graph-ablation}, we perform ablation studies on the implementation techniques for PQEs mentioned in \Cref{sec:pqe-impl-full}: outputting discounted distance and deep linear networks. On the simple directed graphs such as the dense graph, the basic \PQELH without theses techniques works really well, even surpassing the results with both techniques. However, on graphs with more complex structures (\eg, the sparse graph and the sparse graph with block structure), basic versions of \PQELH and \PQEGG starts to perform badly and show large variance, while the versions with both techniques stably trains to the best results. Therefore, for robustness, we use both techniques in other experiments.

\begin{figure}
    \centering
    \includegraphics[scale=0.4, trim=60 10 0 60]{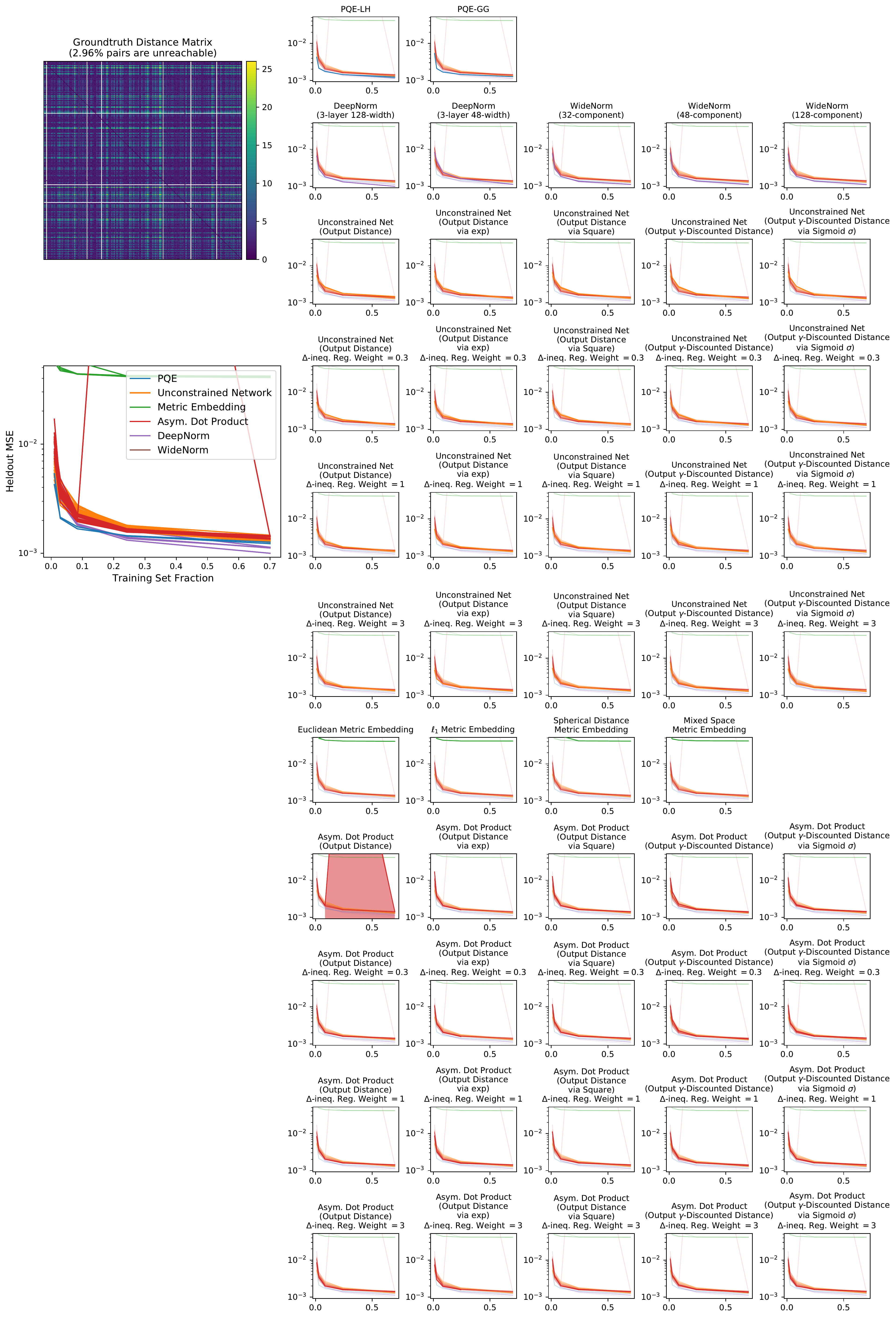}
    \caption{A dense graph. Individual plots on the right show standard deviations.}\label{fig:random-graph-dense}
\end{figure}
\begin{figure}
    \includegraphics[scale=0.4, trim=60 10 0 60]{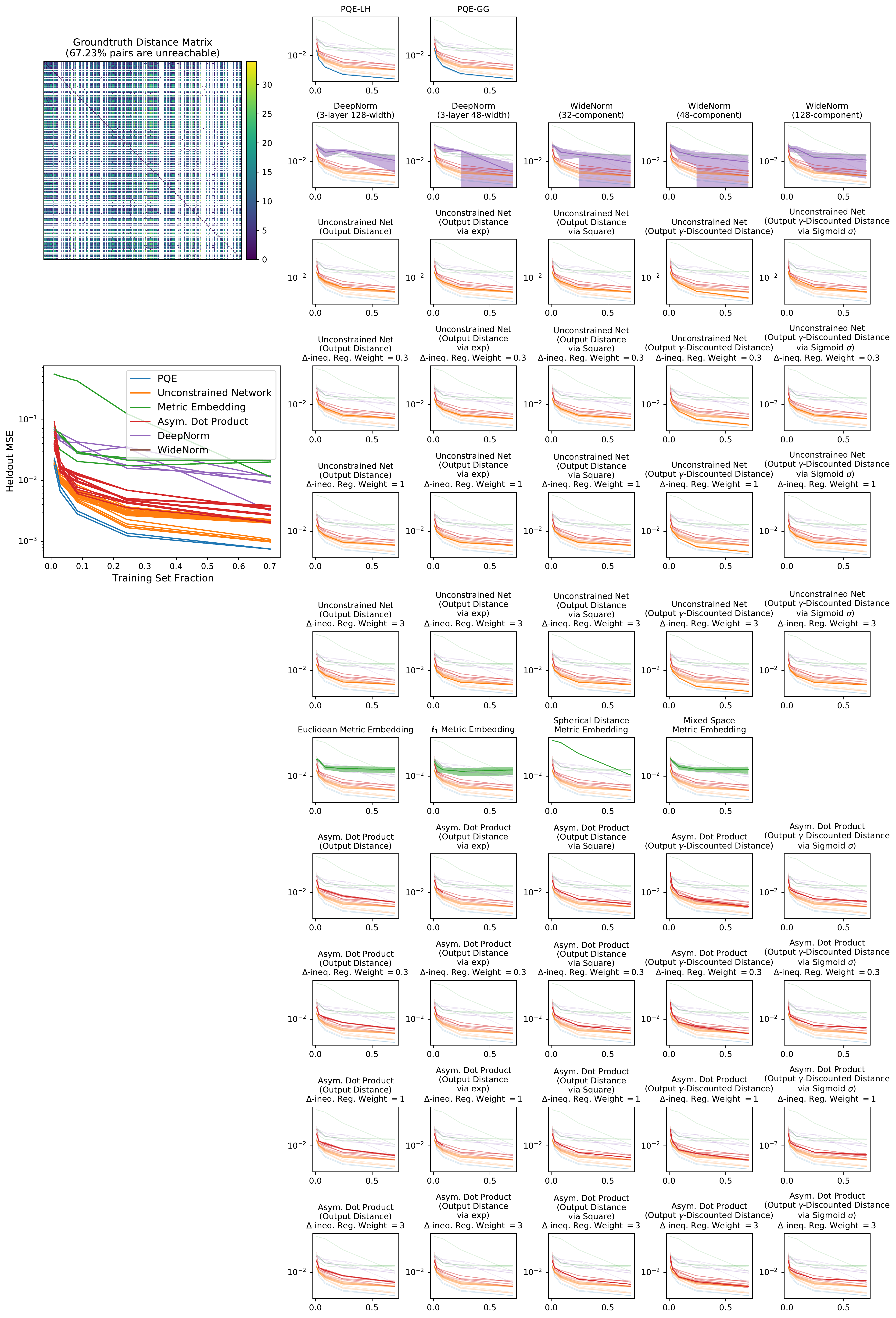}
    \caption{A sparse graph. Individual plots on the right show standard deviations.}\label{fig:random-graph-sparse}
\end{figure}
\begin{figure}
    \includegraphics[scale=0.4, trim=60 10 0 60]{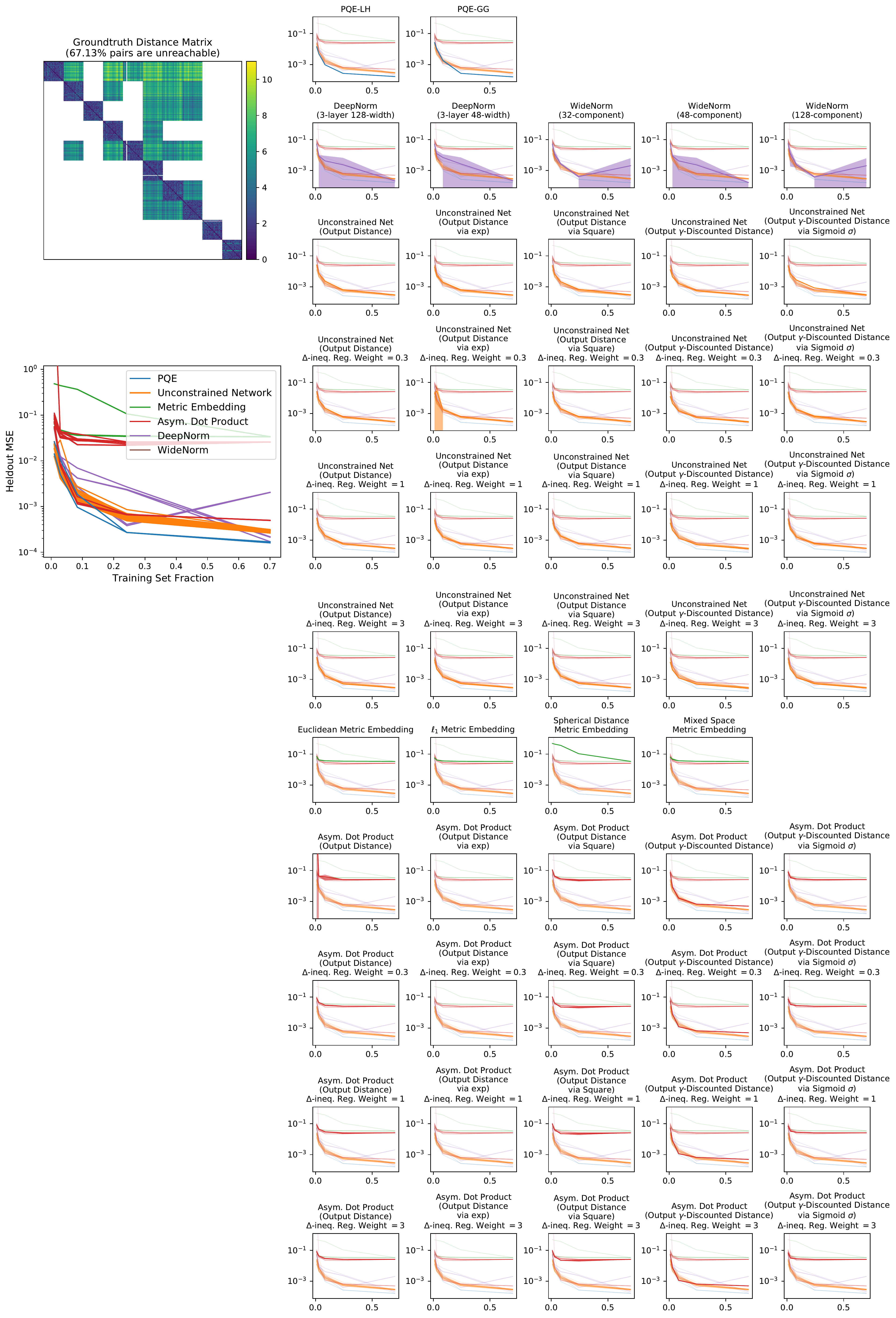}
    \caption{A sparse graph with block structure. Individual plots on the right show standard deviations.}\label{fig:random-graph-blocksparse}
\end{figure}

\begin{figure}
    \centering
    \includegraphics[scale=0.35, trim=70 5 0 0]{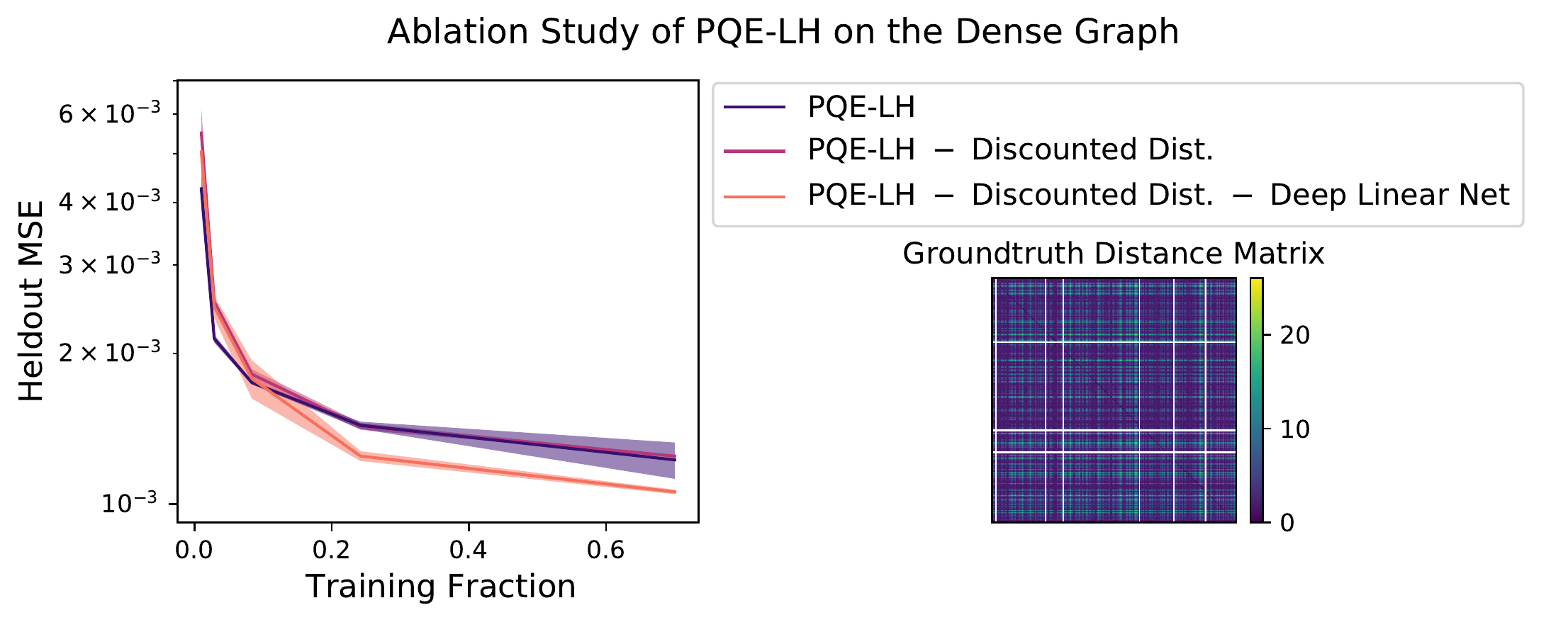}\hfill%
    \includegraphics[scale=0.35, trim=30 5 70 0]{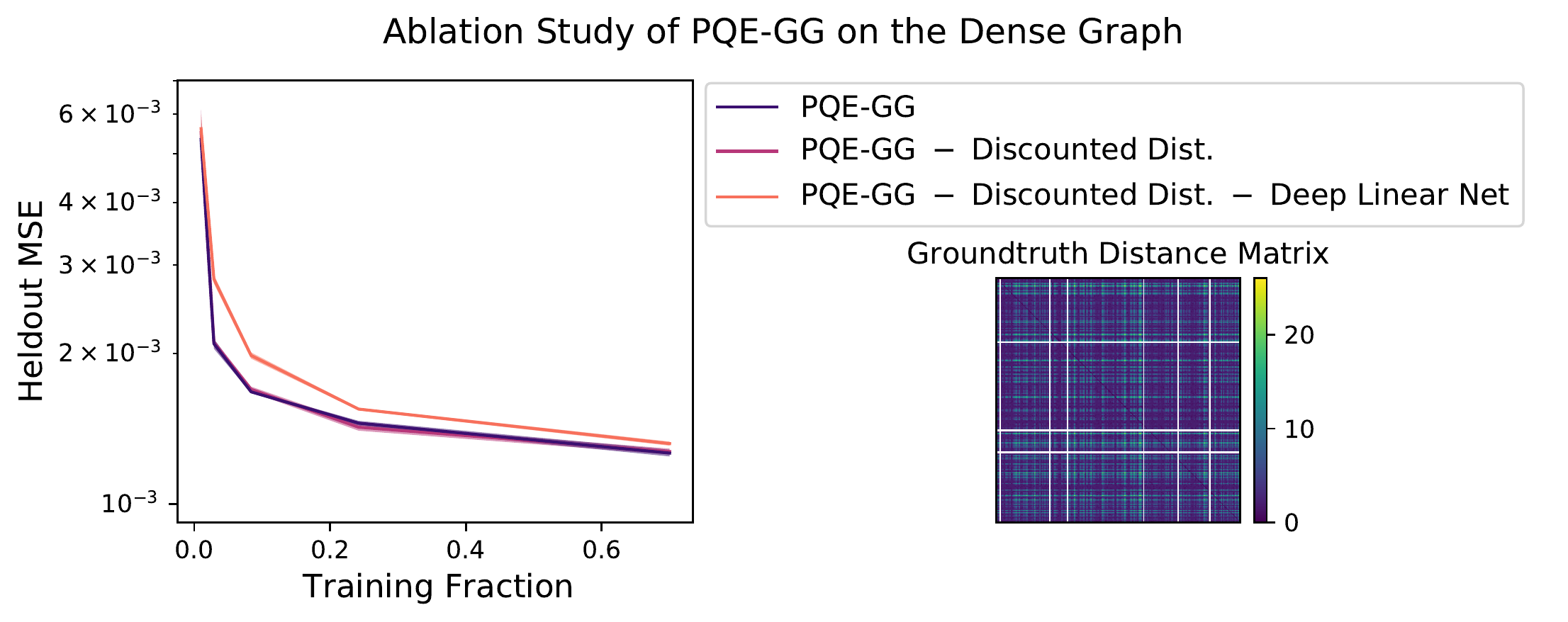}\\[2ex]
    \includegraphics[scale=0.35, trim=70 5 0 0]{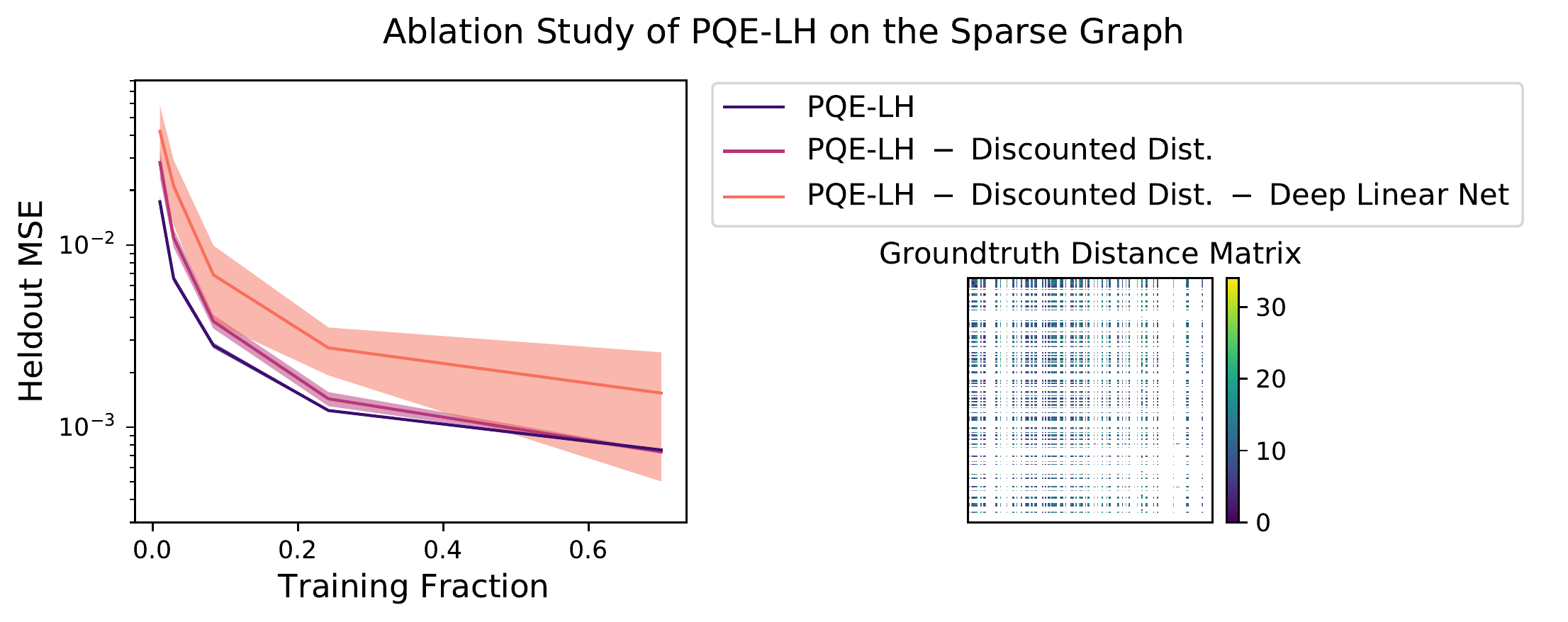}\hfill%
    \includegraphics[scale=0.35, trim=30 5 70 0]{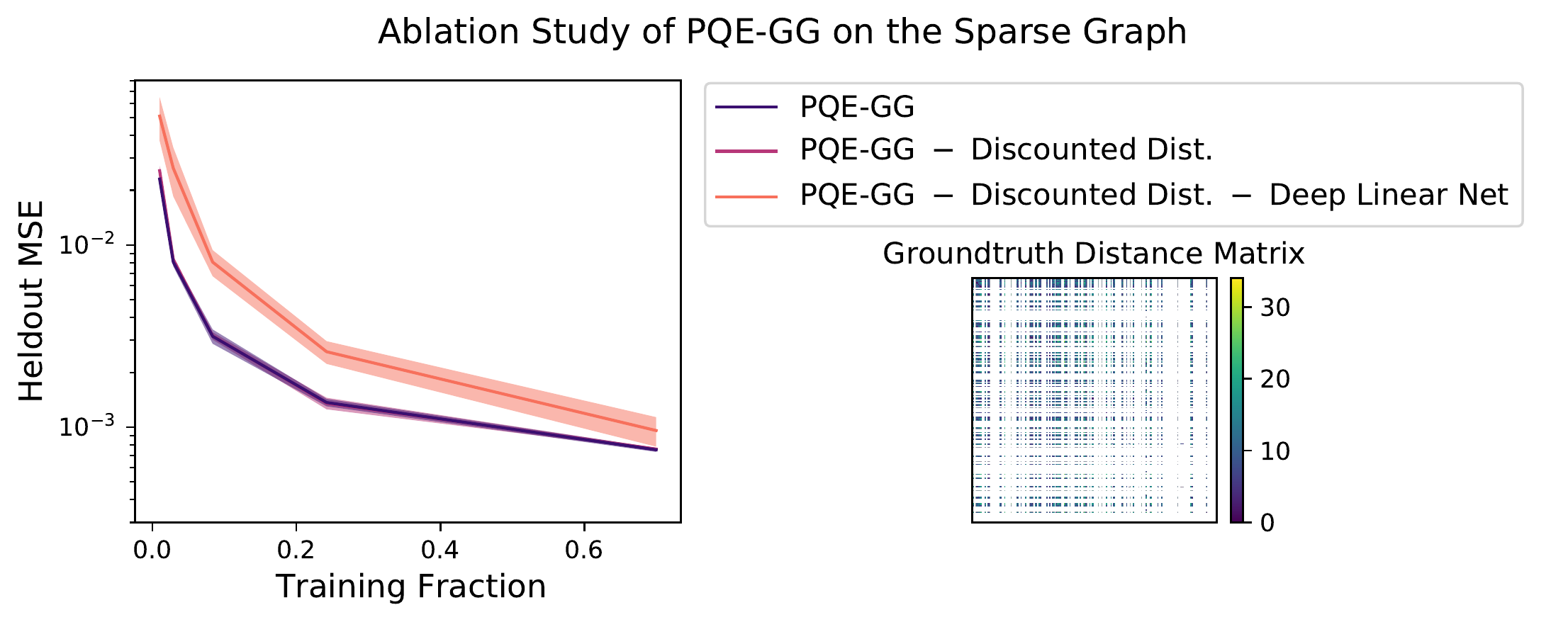}\\[2ex]
    \includegraphics[scale=0.35, trim=70 5 0 0]{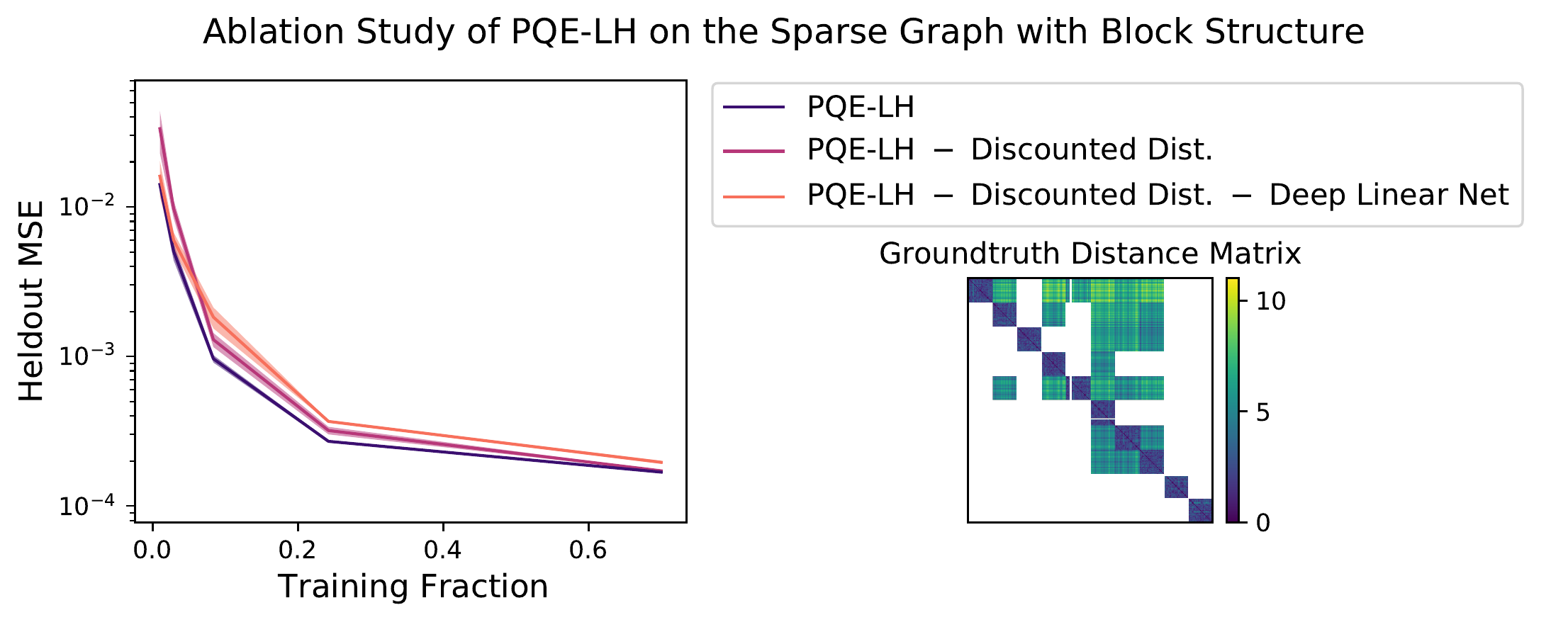}\hfill%
    \includegraphics[scale=0.35, trim=30 5 70 0]{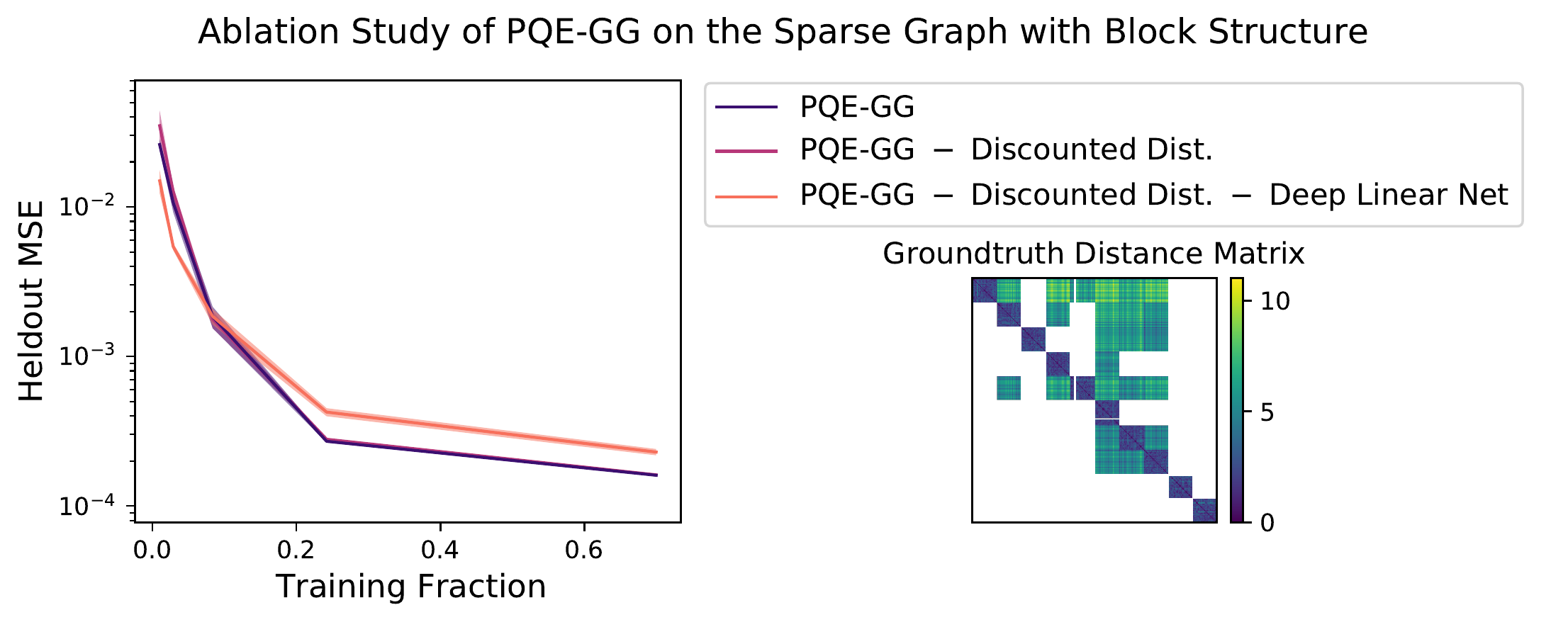}
    \caption{Ablation studies of \PQELH and \PQEGG on three random graphs.}\label{fig:random-graph-ablation}
\end{figure}

\subsubsection{Large-Scale Social Graphs \Qmet Learning}

\paragraph{Data source.} We choose the $\mathsf{Berkeley\hbox{-}Stanford Web Graph}$ \citep{snapnets} as the large-scale \emph{directed} social graph, which consists of $685{,}230$ pages as nodes, and $7{,}600{,}595$ hyperlinks as directed edges. Additionally, we also use the $\mathsf{Youtube}$ social network \citep{snapnets,mislove-2007-socialnetworks} as a \emph{undirected} social graph, which consists of $1{,}134{,}890$ users as nodes, and $2{,}987{,}624$ friendship relations as undirected edges. Both datasets are available from the SNAP website \citep{snapnets} under the BSD license.

\paragraph{Data processing.} For each graph, we use \texttt{node2vec} to obtain $128$-dimensional node features \citep{grover2016node2vec}. Since the graph is large, we use the landmark method \citep{rizi2018shortest} to construct training and test sets. Specifically, we randomly choose $150$ nodes, called landmarks, and compute the distances between these landmarks and all nodes. For directed graph, this means computing distances of both directions. From the obtained pairs and distances, we randomly sample $2{,}500{,}000$ pairs to form the training set. Similarly, we form a test set of $150{,}000$ from a disjoint set of $50$ landmarks. For the undirected graph, we double the size of each set by reversing the pairs, since the distance is symmetrical.

\paragraph{Architecture.} All encoder based methods (PQEs, metric embeddings, dot products) use 128-2048-2048-2048-512 network with ReLU activations and Batch Normalization \citep{ioffe2015batch} after each activation, mapping $128$-dimensional inputs to a $512$-dimensional latent space. \Uncon networks use a similar 256-2048-2048-2048-512-1 network, mapping concatenated the $256$-dimensional input to a scalar output.

\paragraph{Training.} We use $1024$ batch size with the Adam optimizer \citep{kingma2014adam}, with learning rate decaying according to the cosine schedule without restarting \citep{loshchilov2016sgdr} starting from $10^{-4}$ to $0$ over $80$ epochs. All models are optimized \wrt MSE on the $\gamma$-discounted distances, with $\gamma=0.9$. When running with the triangle inequality regularizer, $342 \approx 1024/3$ triplets are uniformly sampled at each iteration.

\begin{table}
    \centering
    \input{figtext/appendix/expriments/social_graph_berstan_full}
    \caption{
    \Qmet learning on the large-scale \emph{directed} $\mathsf{Berkeley\hbox{-}Stanford Web Graph}$.}
    \label{tbl:expr-social-graph-berstan-full}
\end{table}

\begin{table}
    \centering
    \input{figtext/appendix/expriments/social_graph_youtube_full}
    \caption{
    Metric learning on the large-scale \emph{undirected} $\mathsf{Youtube}$ graph. This graph does not have unreachable pairs so the last column is always \texttt{NaN}.}
    \label{tbl:expr-social-graph-youtube-full}
\end{table}

\paragraph{Full results.} \Cref{tbl:expr-social-graph-berstan-full,tbl:expr-social-graph-youtube-full} show full results of distance learning on these two graphs. On the \emph{directed} $\mathsf{Berkeley\hbox{-}Stanford Web Graph}$, \PQELH performs the best (\wrt discounted distance MSE). While \PQEGG has larger discounted distance MSE than some other baselines, it accurately predicts finite distances and outputs large values for unreachable pairs. On the \emph{undirected}  $\mathsf{Youtube}$ graph, perhaps as expected, metric embedding methods have an upper hand, with the best performing method being an Euclidean space embedding. Notably, DeepNorms and WideNorms do much worse than PQEs on this symmetric graph.

\subsubsection{Offline Q-Learning}

As shown in \Cref{prop:opt-goal-reaching-plan-costs-qmet} and \Cref{rmk:mdp-qmet}, we know that a \qmet is formed with the optimal goal-reaching plan costs in a MDP $\mathcal{M} = (\mathcal{S}, \mathcal{A}, \mathcal{R}, \mathcal{P}, \gamma)$ where each action has \emph{unit cost} (\ie, negated reward). The \qmet is defined on $\mathcal{X} \trieq \mathcal{S} \cup (\mathcal{S} \times \mathcal{A})$.

Similarly, \citet{tian2020model} also make this observation and propose to optimize a distance function by Q-learning on a collected set of trajectories. The optimized distance function (\ie, Q-function) is then used with standard planning algorithms such as the Cross Entropy Method (CEM) \citep{de2005tutorial}. The specific model they used is an \uncon network $f \colon (s, a, s') \rightarrow \R$, outputting discounted distances (Q-values).

Due to the existing \qmet structure, we explore using PQEs as the distance function formulation. We mostly follow the algorithm in \citet{tian2020model} except for the following minor differences: \begin{itemize}
    \item \citet{tian2020model} propose to sample half of the goal from future steps of the same trajectory, and half of the goal from similar states across the entire dataset, defined by a nearest neighbor search. For simplicity, in the latter case, we instead sample a random state across the entire dataset.
    \item In \citet{tian2020model}, target goals are defined as single states, and the Q-learning formulation only uses quantities distances from state-action pairs $(s, a) \in \mathcal{S} \times \mathcal{A}$ to states $s'$: $\hat{d}((s, a), s')$.

    However, if we only train on $\hat{d}((s, a), s')$, \qmet embeddings might not learn much about the distance to state-action pairs, or from states, because it may simply only assign finite distances to  $\hat{d}((s, a), s')$, and set everything else to infinite. To prevent such issues, we choose to use state-action pairs as target goals, by adding a random action. Then, the embedding methods only need to embed state-action pairs.

    In planning when the target is actually a single goal $s' \in \mathcal{S}$, we use the following distance/Q-function \begin{equation}
        \hat{d}((s, a), s') \trieq -1 + \frac{1}{\size{\mathcal{A}}} \sum_{a' \in \mathcal{A}} \hat{d}((s, a), (s', a')).
    \end{equation}

    Such a modification is used for all embedding methods (PQEs, metric embeddings, asymmetrical dot products). For \uncon networks, we test both the original formulation (of using single state as goals) and this modification.
\end{itemize}

\begin{figure}[ht]
    \centering
    \includegraphics[scale=0.55, trim=40 10 0 0]{figures/expr/q_learning/5runs/noInf+Asym3R3C9SZ+Init+200+NoSimpleGoal+DoorExt+Seed333_fix_average_wndn.pdf}\hfill%
    \includegraphics[scale=0.4475, trim=0 -46 0 0]{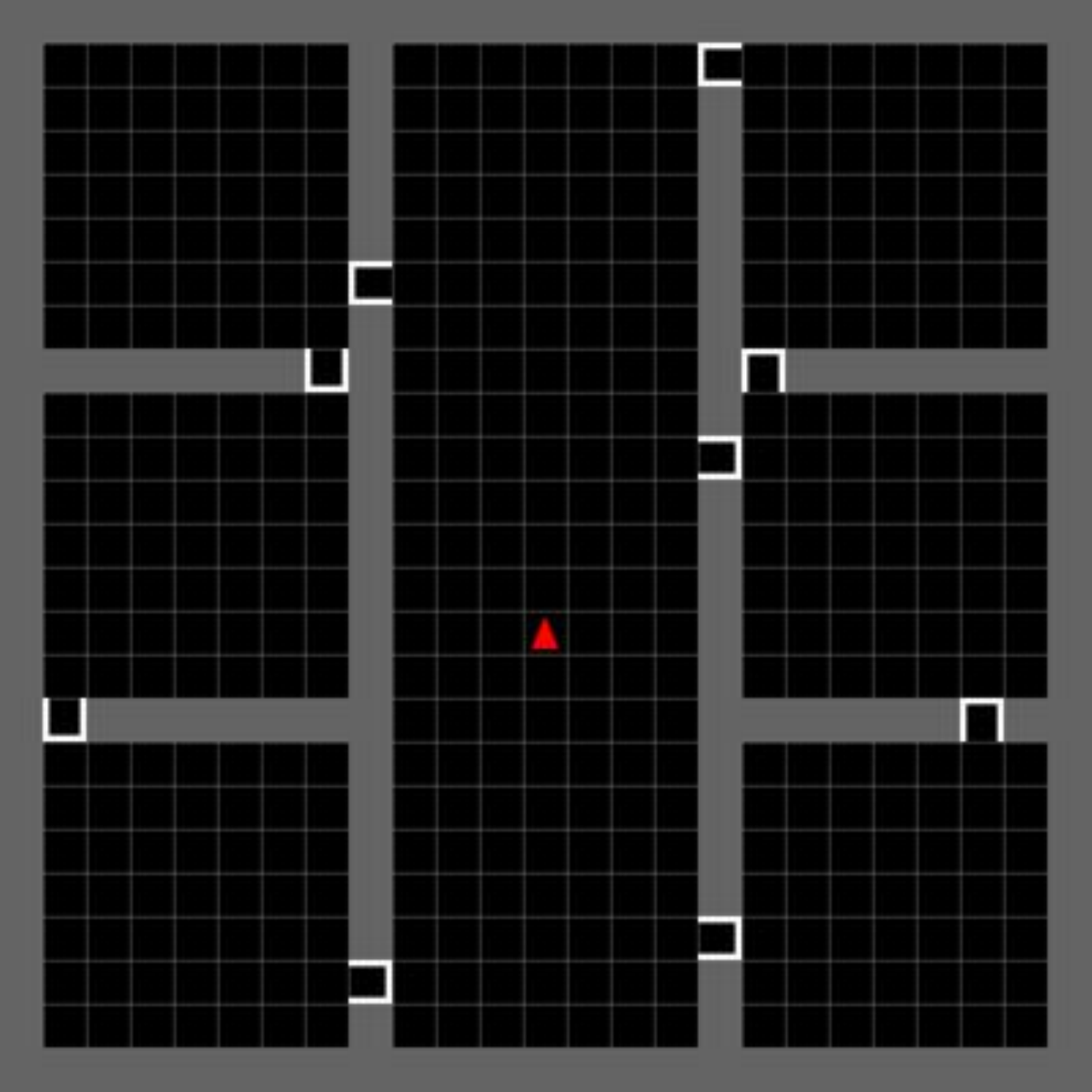}\\[0.3ex]
    \caption{Grid-world offline Q-learning average planning success rates. Right shows the environment.}
    \label{fig:q-learning-avg-env}
\end{figure}

\paragraph{Environment.} The environment is a grid-world with one-way doors, as shown in of \Cref{fig:q-learning-avg-env}, which is built upon \texttt{gym-minigrid} \citep{gym_minigrid} (a project under Apache 2.0 License). The agent has $4$ actions corresponding to moving towards $4$ directions. When it moves toward a direction that is blocked by a wall or an one-way door, it does not move. States are represented as $18$-dimensional vectors, containing the 2D location of the agent (normalized to be within $[-1, 1]^2$). The other dimensions are always constant in our enviroment as they refer to information that can not be changed in this particular environment (\eg, the state of the doors). The agent always starts at a random location in the center room (\eg, the initial position of the red triangle in \Cref{fig:q-learning-avg-env}). The environment also defines a goal sampling distribution as a random location in one of the rooms on the left or right side. Note that this goal distribution is only used for data collection and evaluation. In training, we train goal-conditional policies using the goal sampling mechanism adapted from \citet{tian2020model}, as described above.

\paragraph{Training trajectories.} To collect the training trajectories, we use an $\eps$-greedy planner with groundtruth distance toward the environment goal, with a large $\eps = 0.6$. Each trajectory is capped to have at most $200$ steps.

\paragraph{Architecture.} All encoder based methods (PQEs, metric embeddings, dot products) use 18-2048-2048-2048-1024 network with ReLU activations and Batch Normalization \citep{ioffe2015batch} after each activation, mapping a $18$-dimensional state to four $256$-dimensional latent vectors, corresponding to the embeddings for all four state-action pairs. \Uncon networks use a similar architecture and take in concatenated $36$-dimensional inputs. With the original formulation with states as goals, we use a 36-2048-2048-2048-256-4 network to obtain a $\R^{\size{\mathcal{A}}}$ output, representing the distance/Q-values from each state-action pair to the goal; with the modified formulation with state-action pairs as goals, we use a 36-2048-2048-2048-256-16 network to obtain a $\R^{\size{\mathcal{A}}\times \size{\mathcal{A}}}$ output.

\paragraph{Training.} We use $1024$ batch size with the Adam optimizer \citep{kingma2014adam}, with learning rate decaying according to the cosine schedule without restarting \citep{loshchilov2016sgdr} starting from $10^{-4}$ to $0$ over $1000$ epochs. Since we are running Q-learning, all models are optimized \wrt MSE on the $\gamma$-discounted distances, with $\gamma=0.95$.  When running with the triangle inequality regularizer, $341 \approx 1024/3$ triplets are uniformly sampled at each iteration.

\paragraph{Planning details.} To use the learned distance/Q-function for planning towards a given goal, we perform greedy 1-step planning, where we always select the best action in $\mathcal{A}$ according to the learned model, without any lookahead. In each of $50$ runs, the planner is asked to reach a goal given by the environment within $300$ steps. The set of $50$ initial location and goal states is entirely decided by the seed used, regardless of the model. We run each method $5$ times using the same set of $5$ seeds.

\paragraph{Full results.} Average results across $5$ runs are shown in \Cref{fig:q-learning-avg-env}, with full results (with standard deviations) shown in  \Cref{fig:q-learning-full}. Planning performance across the formulations vary a lot, with PQEs and the Euclidean metric embedding being the best and most data-efficient ones. Using either formulation (states vs.~state-action pairs as goals) does not seem to affect the performance of \uncon networks. We note that the the asymmetrical dot product formulation outputting discounted distance is similar to Universal Value Function Approximators (UVFA) formulation \citep{schaul2015universal}; the \uncon network outputting discounted distance with states as goals is the same formulation as the method from \citet{tian2020model}.

\afterpage{%
\begin{figure}
    \centering
    \includegraphics[scale=0.43, trim=40 0 0 0]{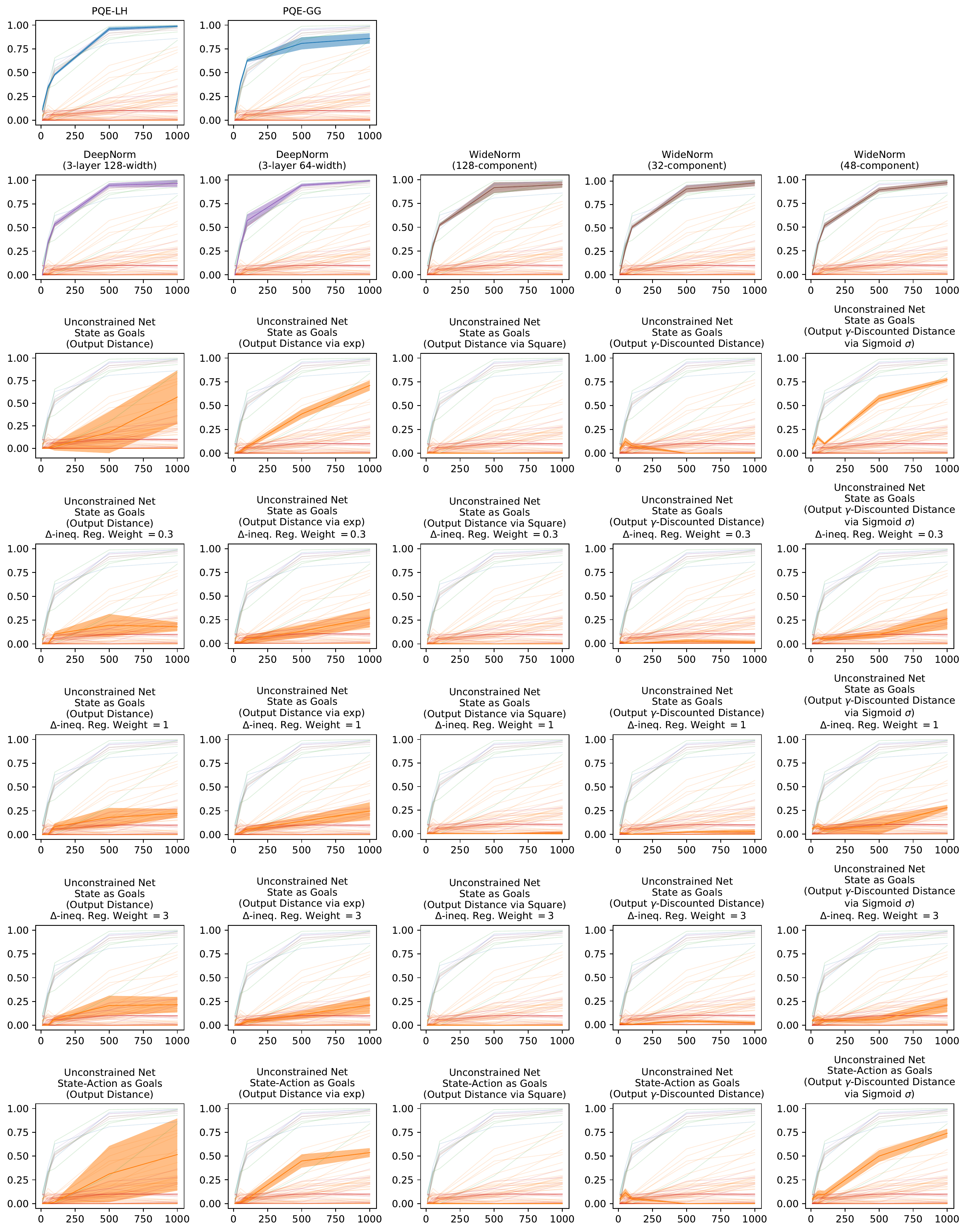}
    \caption{Grid-world offline Q-learning full results. Individual plots on show standard deviations.}
    \label{fig:q-learning-full}
\end{figure}%
\clearpage
}
\afterpage{%
\begin{figure}\ContinuedFloat
    \centering
    \includegraphics[scale=0.43, trim=40 0 0 0]{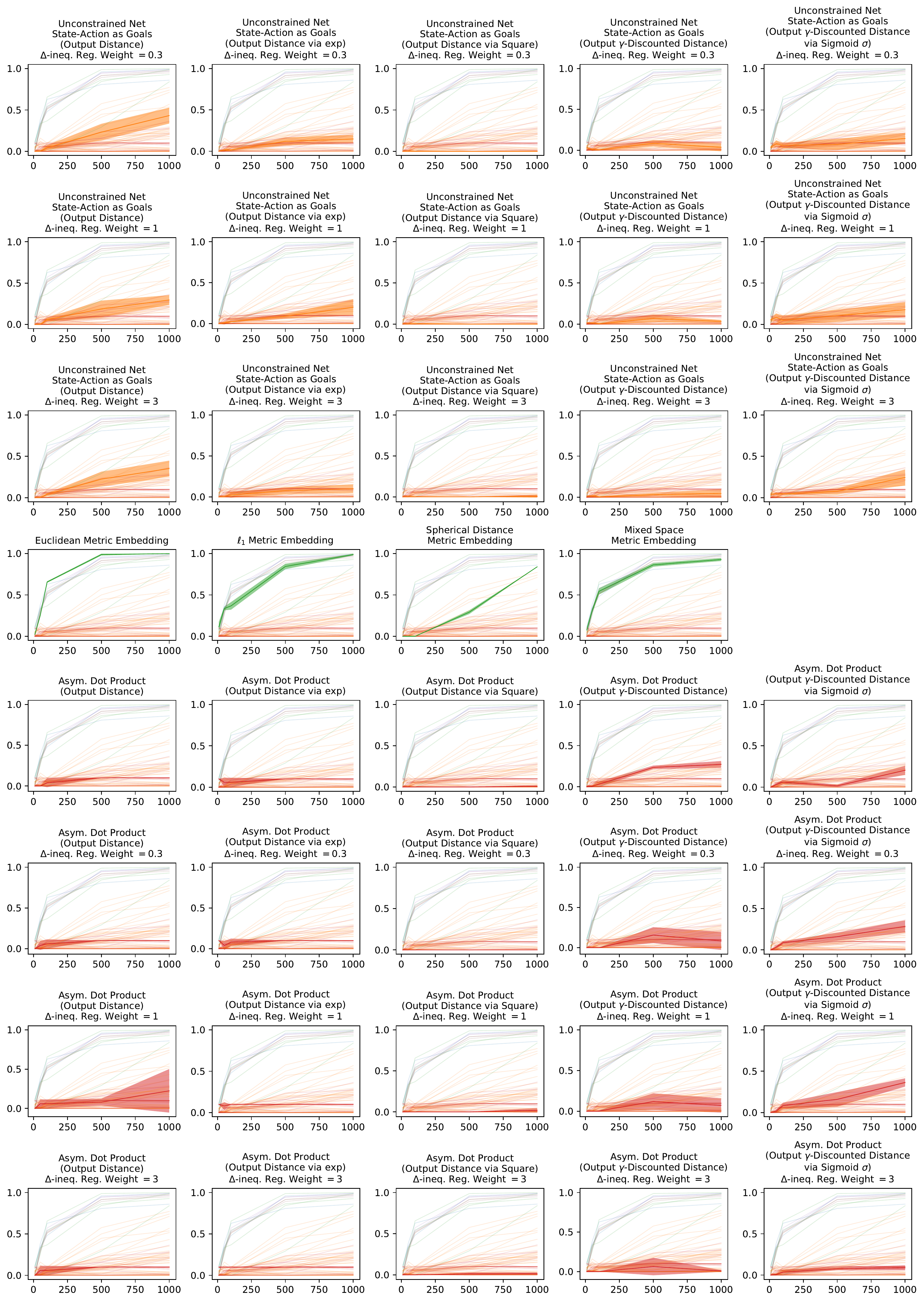}
    \caption[]{Grid-world offline Q-learning full results (cont.). Individual plots on show standard deviations.}
\end{figure}%
\clearpage
}

\endgroup

\end{document}